\documentclass[10pt]{article} 
\usepackage[preprint]{tmlr}

\usepackage{amsmath,amsfonts,bm}

\def\eqref#1{equation~\ref{#1}}

\def\1{\bm{1}}

\DeclareMathAlphabet{\mathsfit}{\encodingdefault}{\sfdefault}{m}{sl}
\SetMathAlphabet{\mathsfit}{bold}{\encodingdefault}{\sfdefault}{bx}{n}

\usepackage{hyperref}
\usepackage{url}
\usepackage{booktabs}
\usepackage{tcolorbox}
\usepackage{float}
\usepackage{tikz} 
\usepackage{xcolor} 

\usepackage{graphicx}
\usepackage{enumitem}
\usepackage{multirow}
\usepackage{booktabs}
\usepackage[table]{xcolor}
\usepackage{wrapfig}
\usepackage{wrapfig}
\usepackage{siunitx}
\usepackage{amsmath}
\usepackage{amsmath,amssymb}
\usepackage{algorithm}
\usepackage{algorithmic}

\usepackage{float}
\usepackage{subcaption} 
\usepackage{wrapfig}
\usepackage{caption}        

\usepackage{titletoc} 

\newcommand{\ourmethod}{{{PRISMA }}}

\usepackage{pifont}
\newcommand{\cmark}{\ding{51}}%
\newcommand{\xmark}{\normalfont\ding{55}}%

\newcommand{\bu}{\mathbf{u}}
\newcommand{\ba}{\mathbf{a}}
\newcommand{\br}{\mathbf{r}}
\newcommand{\bM}{\mathbf{M}}
\newcommand{\bx}{\mathbf{x}}
\newcommand{\by}{\mathbf{y}}

\newcommand{\sra}{\textcolor{blue}{\text{SRA}}} 
\renewcommand{\vec}[1]{\mathbf{#1}}

\title{PRISMA: Improving the Accuracy–Latency Frontier of Diffusion-based PDE Solvers Using Physics-Informed Spectral Attention}

\author{Medha Sawhney, Abhilash Neog*, Mridul Khurana* \& Anuj Karpatne  \\
Department of Computer Science\\
Virginia Tech\\
\texttt{\{medha,abhilash22,mridul,karpatne\}@vt.edu} 
}

\begin{document}

\maketitle

\begin{abstract}
Diffusion-based solvers for partial differential equations (PDEs) are often bottle-necked by slow gradient-based test-time optimization routines that use PDE residuals for loss guidance. They additionally suffer from optimization instabilities and are unable to dynamically adapt their inference scheme in the presence of noisy PDE residuals. 
To address these limitations, we introduce PRISMA (PDE Residual Informed Spectral Modulation with Attention), a conditional diffusion neural operator that embeds PDE residuals directly into the model's architecture via an attention-inspired modulation mechanism in the spectral domain, enabling gradient-descent free inference. In contrast to previous methods that use PDE loss solely as external optimization targets, PRISMA integrates PDE residuals as integral architectural features, making it inherently fast, robust, accurate, and free from sensitive hyperparameter tuning. We show that {PRISMA has competitive accuracy, at substantially lower inference costs}, compared to previous methods across five benchmark PDEs especially with noisy observations, while using 10x to 100x fewer denoising steps, leading to 15x to 250x faster inference.
\end{abstract}

\section{Introduction}
\label{sec:intro}

Learning to solve partial differential equations (PDEs) is a rapidly growing area of research in machine learningn where given a spatial domain $\Omega \subset \mathbb{R}^d$, we are interested in solving parametric PDEs of the form:
\begin{align}
    \mathcal{{P}}(\bu(c); \ba(c)) = 0, \quad c \in \Omega, \label{eq:pde}
\end{align}
subject to some initial and boundary conditions where $\ba \in \mathcal{A}$ represents the parameter field of the PDE (e.g., material coefficients or source terms), $\bu \in \mathcal{U}$ denotes the solution field, and $\mathcal{{P}}$ is a non-linear differential operator \citep{pinnraissi2019physics, lu2019deeponet}. There are two main classes of problems when solving PDEs: (i) the \textit{forward problem}, where we infer $\bu$ given observations of the input parameters, $\ba_\text{obs}$, and the \textit{inverse problem}, where we infer $\ba$ given observations of $\bu_\text{obs}$. 
Note that in real-world applications, 
both $\ba_{\text{obs}}$ and/or $\bu_{\text{obs}}$ may be corrupted with sparsity, observation noise, or other forms of  degradation, and test-time conditions may deviate from the training distribution.

A prominent approach to solve a family of PDEs is to use \textit{operator learning}  methods such as the Fourier Neural Operator (FNO)~\citep{fno}, which learns resolution-independent mappings between function spaces of $\ba$ and $\bu$. Once trained, neural operators act as fast surrogates to numerical PDE solvers delivering orders-of-magnitude faster inference. However, their effectiveness typically relies on dense, clean observations and can degrade substantially under sparse or noisy observations, especially for ill-posed inverse problems.\looseness=-1

The rise of \textit{generative models} has inspired another class of methods for solving PDEs using diffusion-based backbones \citep{huang2024diffusionpde, fundps, jacobsen2025cocogen, jiang2024ode}. These methods typically train an unconditional model to learn the joint distribution of $\ba$ and $\bu$, and introduce PDE residuals only at inference to optimize using gradient steps via diffusion posterior sampling (DPS). They offer two key advantages over operator learning methods: (i) they generate full posterior distributions of $\ba$ and/or $\bu$, enabling principled uncertainty quantification, and (ii) they are flexible and can be adapted to novel task settings (e.g., with sparse observations) at test-time without full retraining.

Despite these advantages, DPS-based diffusion PDE solvers suffer from three major limitations. \textit{First}, they are significantly slow during inference because unconditional generation typically requires several denoising steps starting off from pure noise. This cost is further amplified when PDE residuals are used in the gradient updates of DPS, requiring expensive autograd-based computations. \textit{Second}, their performance is highly sensitive to guidance hyperparameters (or weights) used in DPS, often requiring task-dependent tuning or calibration. In realistic scenarios, the form of input data corruption in $\ba$ and $\bu$ may not be fully known (e.g., the type and magnitude of sensor noise), making it challenging to reliably tune guidance weights for unseen tasks. 
\textit{Third}, optimizing PDE residuals as external loss is known to suffer from training instabilities as PDE residual gradients vary sharply across noise levels spanning multiple frequency modes ~\citep{piddmzhang2025physics,cheng2024gradient,utkarsh2025physics}.

An alternative to DPS-based methods is to train \emph{conditional} diffusion models that directly generate $(\ba,\bu)$ given observations $(\ba_\text{obs},\bu_\text{obs})$, avoiding per-instance gradient-based optimization at inference \citep{shysheya2024conditional, kohl2026benchmarking, dasgupta2025conditional}. However, these methods require task-specific training of the conditional model to fixed observation regimes (e.g., full vs.\ sparse vs.\ noisy) and problem directions (forward vs.\ inverse), making it challenging to deploy them in novel task settings without  retraining.

Given the complementary strengths of DPS-based and conditional diffusion models, we ask:  \textbf{\emph{can we use PDE knowledge during inference in diffusion models without invoking expensive gradient descent and without performing any retraining or hyperparameter fine-tuning on unseen tasks?}} In other words, can we bridge the best of both worlds: conditional models and DPS paradigms, achieving fast inference as well as zero-shot adaptability to novel task settings?

\begin{figure*}[t]
    \centering
     \includegraphics[width=0.95\textwidth]{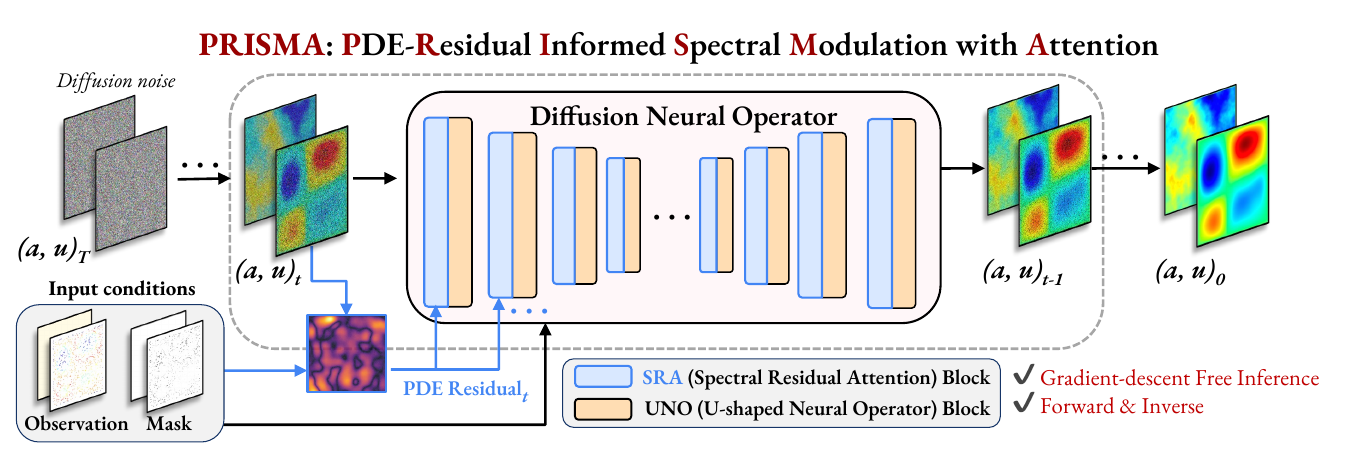}
    \small \caption{\textbf{PRISMA inference pipeline}. We use a U-shaped Diffusion Neural Operator (UNO) to iteratively refine estimates of $\ba$ and $\bu$ starting from diffusion noise $(\mathbf{a}, \mathbf{u})_T$ to a clean solution $(\mathbf{a}, \mathbf{u})_0$. At each denoising step, PDE residuals 
    are architecturally injected via a novel \sra{} block at every layer of the UNO, enabling fast, {gradient-descent free} inference for both forward and inverse problems.}
    \label{fig:main_architecture}
    
    \vspace{-2ex}
\end{figure*}

We introduce \textbf{\ourmethod} (PDE Residual Informed Spectral Modulation with Attention), a  \emph{conditional} 
diffusion neural operator that integrates PDE residual structure \emph{inside} the denoiser architecture rather than enforcing it through test-time optimization in DPS. Specifically, we propose a \textbf{Spectral Residual Attention} (\textbf{SRA}) block (Fig.~\ref{fig:main_architecture}) inserted throughout a U-shaped denoiser, which attends to PDE residual fields in the \emph{spectral domain} to modulate the role of PDE residuals across frequency modes.
This is important because different residual frequencies contribute unevenly and can be differentially corrupted under noise. \ourmethod uses physics during both training and inference, yet enables \textbf{gradient-descent-free} sampling. In addition, \ourmethod uses observation masks to encode arbitrary conditioning patterns involving observation regimes and problem directions, allowing a \emph{single} model to solve forward and inverse problems under full, sparse, or noisy observations without task-specific training or inference-time hyperparameter adaptation.

Our work makes five key contributions. 
\textbf{\textit{(1)}}  We introduce a novel strategy for incorporating physical knowledge in diffusion models as  \textit{\textbf{ architectural guidance}}, embedding PDE residual information as an \textit{internal} feature in the model rather than enforcing physics through an external, test-time guidance loss. 
\textbf{\textit{(2)}} This design enables \textit{\textbf{gradient-descent-free inference}} which fundamentally removes the need for costly and sensitive optimization of likelihood and PDE loss terms during inference, improving speed and robustness. \textbf{\textit{(3)}} We use a \textit{\textbf{mask-conditioned unified training}} scheme that can be zero-shot applied to any target task (sparse/noisy/full \& forward/inverse) without retraining or hyperparameter tuning. 
\textbf{\textit{(4)}} Across diverse PDE benchmarks, we demonstrate a favorable \textit{\textbf{accuracy--latency trade-off}}, achieving \textbf{15$\times$--250$\times$ faster} inference than diffusion baselines while maintaining competitive accuracy.
\textbf{\textit{(5)}} We further demonstrate the generalizability and robustness of \ourmethod under varying distribution shifts, including noisy observations and extrapolation in physical parameters.

\section{Related Works}

\textbf{Neural PDE solvers:} Among the many deep learning approaches for solving PDEs, two widely used families are physics-informed methods and operator-learning methods. While Physics-Informed Neural Networks (PINNs)~\citep{pinnraissi2019physics} enforce the governing equations by minimizing PDE residual losses during training, they suffer from optimization and scaling challenges ~\citep{zhongkai2024pinnacle, liu2024preconditioning}. These challenges motivate operator-learning methods such as FNO and DeepONet, which learn resolution-invariant mappings between parameter fields and solution fields for fast amortized inference~\citep{fno, unorahman2022u, kovachki2023neural, lu2019deeponet}. Hybrid variants such as PINO inject PDE residual terms into neural operators to further improve physical consistency~\citep{pino}, but still inherit sensitivity and tuning issues associated with PDE-loss optimization. Importantly, PINNs and neural operators typically produce deterministic predictions and are most effective under clean, dense observations, offering limited native uncertainty quantification, particularly problematic for ill-posed inverse problems.

\textbf{Generative models for PDEs:} 
Diffusion models enable sampling from distributions over fields, providing uncertainty quantification and flexible conditioning. For example, DiffusionPDE ~\cite{huang2024diffusionpde} learns the joint probability distribution of $(\ba, \bu)$ using a diffusion model during training, and employs PDE-residual guidance as a loss term during inference to produce physically consistent solutions. By minimizing PDE residuals during inference,  DiffusionPDE can work with arbitrary forms of sparsity in $\ba$ and/or $\bu$ that has not been seen during training. However, since DiffusionPDE operates directly in the native spatial domain, it remains tied to a fixed spatial resolution of $\ba$ and $\bu$. Newer methods expand the design space by incorporating score-based or diffusion backbones for physics  under different training and conditioning paradigms ~\citep{shu2023physics,bastek2024physicsdiff,jacobsen2025cocogen,li2025videopde}. Because PDE states are functions, not images, several works lift generative modeling to function spaces for resolution independence: Denoising Diffusion Operators (DDOs) perturb Gaussian random fields and provide discretization robustness~\citep{ddolim2023score, fundps}; and infinite-dimensional diffusion frameworks formalize well-posedness and dimension-free properties~\citep{pidstrigach2023infinite}.
Building on this, FunDPS~\citep{fundps} uses U-shaped neural operators (UNO)~\citep{unorahman2022u} as the denoising backbone, enabling diffusion neural operators that support posterior inference over $\ba$ and/or $\bu$ while remaining resolution-agnostic.

\begin{table}[t]
\centering
\caption{
Comparing \ourmethod with prior PDE solvers based on where PDE knowledge is used and whether inference is gradient-free, with inference time and forward error on Darcy.
}
\label{tab:paradigm_comparison_final}

\renewcommand{\arraystretch}{0.7}
\fontsize{8pt}{8pt}\selectfont
\setlength{\tabcolsep}{3pt}

\resizebox{0.75\linewidth}{!}{%
\begin{tabular}{@{}lccccc@{}}
\toprule
\bfseries Method
& \bfseries Grad.-Free
& \bfseries PDE Train
& \bfseries PDE Inf.
& \bfseries Time (s) $\downarrow$
& \bfseries Error $\downarrow$ \\
\midrule
PINN         & \xmark & \cmark & \cmark & 3.3  & 15.40\% \\
FNO          & \cmark & \xmark & \xmark & 0.10 & 5.3\%  \\
DeepONet     & \cmark & \xmark & \xmark & 0.09 & 12.3\% \\
PINO         & \cmark & \cmark & \xmark & 0.11 & 4.0\%  \\
DiffusionPDE & \xmark & \xmark & \cmark & 213  & 2.9\%  \\
FunDPS       & \xmark & \xmark & \cmark & 11.8 & 1.4\%  \\
\midrule
\bfseries PRISMA
& \bfseries \cmark
& \bfseries \cmark
& \bfseries \cmark
& 0.18
& \bfseries 1.1\% \\
\bottomrule
\end{tabular}%
}
\end{table}
\textbf{Limitations of diffusion-based PDE solvers:}
Diffusion PDE solvers follow two common paradigms. \emph{Unconditional} approaches learn a diffusion prior over PDE fields (often jointly over $(\ba,\bu)$) and enforce observations/physics at inference via DPS-style guidance, which typically starts from noise and requires many denoising steps, with additional gradient-descent-like updates that are expensive and sensitive to guidance weights~\citep{chung2022diffusion,huang2024diffusionpde,piddmzhang2025physics}. \emph{Conditional} diffusion models can reduce reliance on per-sample optimization by conditioning directly on measurements during training~\citep{shysheya2024conditional, li2025videopde}, but require task specific finetuning. Moreover, physics is still commonly imposed as an \emph{external objective} rather than an internal architectural signal, making residual-based optimization prone to instability, tuning overhead, and lack of sensitivity to varying frequency modes of PDE residuals~\citep{piddmzhang2025physics,cheng2024gradient,utkarsh2025physics}.

Table~\ref{tab:paradigm_comparison_final} summarizes key differences between \ourmethod and prior PDE solvers. While previous works use PDE knowledge either only during training (PINNs/PINO) or only during inference (DiffusionPDE/FunDPS), we use PDE knowledge during \textit{both} training and inference. Further, by incorporating PDE residuals through attention-inspired spectral modulation
inside the model architecture, \ourmethod is free from gradient-descent during inference, and also does not require any task-specific retraining or hyper-parameter tuning. This helps us improve on both accuracy and latency frontiers, as further demonstrated in our experiments.

\section{PRISMA: Proposed Unified Framework for Solving PDEs}

\subsection{Problem Statement for Unified Generative Modeling of PDEs}

We first present a unified problem statement for solving PDEs using conditional modeling paradigm.
We consider a common framework for solving both forward and inverse problems by training a single model to learn the joint probability of
$(\ba, \bu)$ conditioned on any partial set of observations, $(\ba_{\text{obs}}, \bu_{\text{obs}})$. Concretely, we model $(\ba,\bu)\sim p_\theta(\ba,\bu\mid \ba_{\text{obs}},\bu_{\text{obs}},\bM_a,\bM_u)$, i.e., $(\ba_{\text{obs}},\bu_{\text{obs}},\bM_a,\bM_u)\rightarrow(\ba,\bu)$. Mask-based conditioning has also been explored in prior diffusion models for arbitrarily conditioned physical-field diffusion~\citep{long2025arbitrarily}; here, masks serve as a unified task interface. 
To encode varying configurations of input observations in the conditioning of the joint model, we introduce binary masks $(\bM_a, \bM_u)$ that serve two uses: (i) they indicate the sparsity patterns in $\ba_{\text{obs}}$ and $\bu_{\text{obs}}$, and (ii) they specify whether we are solving the forward or inverse problem. A value of 1 in these masks indicates that an observation is present, while 0 indicates no observation. Table \ref{tab:prisma_configs_mask} summarizes the different configurations of input conditions that result in different problem settings, unifying forward and inverse problems as well as full and sparse observations.


\subsection{Backbone: Conditional U-Shaped Diffusion Neural Operators} 

We model the joint distribution of the concatenated continuous field, $\bx = [\ba, \bu]$ using a conditional Denoising Diffusion Operator (DDO) \citep{ddolim2023score}, a recent class of score-based generative models designed for function spaces \cite{functionspacekerrigan2022diffusion, pidstrigach2023infinite}. The forward process in DDO perturbs a clean function $\bx_0$ with progressively larger noise, yielding a noisy function $\bx_\sigma = \bx_0 + \sigma \boldsymbol{\epsilon}$ at noise level $\sigma$. 
To preserve spatial coherence and functional structure, the noise
$\boldsymbol{\epsilon}$ is sampled from a Gaussian Random Field (GRF)
with zero mean and a kernel-defined covariance,
$\boldsymbol{\epsilon} \sim \mathcal{N}(0, \mathbf{C})$ (Appendix \ref{appendix:implementation_details}).

The corresponding reverse diffusion process is learned by a denoiser network, $D_\theta$, parameterized as a neural operator to predict $\bx_0$ from $\bx_\sigma$. 
We specifically instantiate $D_\theta$ as a U-shaped Neural Operator (UNO) \citep{unorahman2022u} comprising of a multi-scale hierarchy of $L$ layers, where the transformation of features from layer $l$ to layer $l+1$ is defined as:
{\small
\begin{align}
\label{eq:uno}
    \bx^{l+1} = \underbrace{\mathcal{F}^{-1}\!\Big(W^{l}\odot \mathcal{F}(\bx^l)\Big)}_{\text{Global Spectral Path}} + \underbrace{\psi^{\,l}(\bx^{l})}_{\text{Local Spatial Path}},
\end{align}
}
where the global spectral path applies learnable weights $W^l$ in Fourier space using the Fast Fourier Transform ($\mathcal{F}$) and its inverse ($\mathcal{F}^{-1}$), while the local spatial path is a local residual block $\psi^l$.
  
We train $D_\theta$ using the Elucidated Diffusion Model (EDM) \citep{karras2022elucidating} objective, defined as: 
{\small
\begin{align}
\label{eq:edm_loss} 
    \mathcal{L}_{\text{EDM}} = \mathbb{E}_{\bx_0, \by, \sigma, \boldsymbol{\epsilon}} \left[\lambda(\sigma)\, \| D_\theta(\bx_{\sigma,\epsilon}, \sigma, \by) - \bx_0 \|_2^2 \right].
\end{align}
}
\vspace{-1em}

\vspace{-1em}

\begin{table}[h]
\centering
\caption{ \small Input configurations for different PDE-solving settings.}
\label{tab:prisma_configs_mask}

\renewcommand{\arraystretch}{0.9}
\setlength{\tabcolsep}{3pt}

\resizebox{0.58\linewidth}{!}{%
\begin{tabular}{@{}lcccc@{}}
\toprule
\textbf{Task}
& \textbf{Obs. ($\ba_{\mathrm{obs}}$)}
& \textbf{Obs. ($\bu_{\mathrm{obs}}$)}
& \textbf{Mask ($\bM_a$)}
& \textbf{Mask ($\bM_u$)} \\
\midrule
Full (For)   & Full   & --     & True          & False \\
Full (Inv)   & --     & Full   & False         & True  \\
\midrule
Sparse (For)  & Sparse & --     & Sparse mask & False \\
Sparse (Inv)  & --     & Sparse & False       & Sparse mask \\
Sparse (Both) & Sparse & Sparse & Sparse mask & Sparse mask \\
\bottomrule
\end{tabular}%
}
\end{table}
\vspace{-1em}
\subsection{PRISMA Model Architecture}
\label{sec:prisma_method}

There are two key innovations that we introduce in our UNO backbone to arrive at {PRISMA}. First, we compute PDE residuals of $\bx$ at every denoising step informed by available observations $\bx_\text{obs}$, and the current denoiser state $\bx_\sigma$. Second, we inject the residuals in the attention mechanism of a novel \textbf{Spectral Residual Attention} (\sra) block applied at every UNO layer (Figure \ref{fig:sra_block}). We describe both these innovations below:

\paragraph{Computing Observation-Informed PDE Residuals:} Given some observations $\bx_\text{obs}$ and inputs to the denoiser network $\bx_\sigma = (\ba_\sigma, \bu_\sigma)$, we want to compute the physical consistency of $\bx$ at unobserved locations (i.e., locations that are not part of $\bx_\text{obs}$. To accomplish this, we first copy the information from $\bx_\text{obs}$ to $\bx$ using masks $\bM = (\bM_a, \bM_u)$, obtaining \textit{mixed fields}: $\bx_\text{mix} = (\ba_\text{mix}, \bu_\text{mix})$ as follows: 
{\small
\begin{equation}
  {\ba}_\text{mix} = \bM_a \odot \ba_{\text{obs}} + (1-\bM_a) \odot {\ba_\sigma},
\quad
  {\bu}_\text{mix} = \bM_u \odot \bu_{\text{obs}} + (1-\bM_u) \odot {\bu_\sigma},
\end{equation}
}
where $\bx_\sigma = (\ba_\sigma, \bu_\sigma)$ is the input to the denoiser during training, which can be replaced with $\bx_t = (\ba_t, \bu_t)$ during inference. We then compute PDE residuals, $\br = \mathcal{R}({\ba}_\text{mix}, {\bu}_\text{mix})$ (Figure \ref{fig:sra_block} (left)).
Note that $\br$ is computed once at every denoising step at the model's native resolution and then progressively downsampled to provide a {multi-resolution} guidance signal at every layer of UNO.

\paragraph{Injecting PDE residuals via SRA:}

We adapt the UNO architecture by introducing a novel SRA block inside the global spectral path of every UNO layer, enabling frequency-selective corrections guided by the spectral content of the PDE residual.
The SRA block at layer $l$ first modulates the input feature maps $\bx^l$ using the PDE residual $\br$ to produce an intermediate physics-informed state, $\bx^{l}_{\text{SRA}}$. Intuitively, $\bx^{l}_{\text{SRA}}$ reweights the spectrum of $\bx^l$ by attending to frequency modes where the residual indicates larger physical inconsistency. This state is then passed through the global spectral path of the UNO block at layer $l$, while the original feature map $\bx^l$ is passed into the local spatial path as:

{\small
\begin{equation}
\label{eq:uno_phys_step}
\textcolor{blue}{\bx^{l}_{\text{SRA}}}
= \sra(\bx^l, \br),
\quad
\bx^{l+1}
= \mathcal{F}^{-1}\Big(
W^{l}\odot\mathcal{F}(\textcolor{blue}{\bx^{l}_{\text{SRA}}})
\Big)+\psi^{l}(\bx^{l}).
\end{equation}
}

\subsection{Spectral Residual Attention (SRA) Block}
\label{sec:sra}

\begin{figure*}[t]
    \centering
     \includegraphics[width=0.8\textwidth]{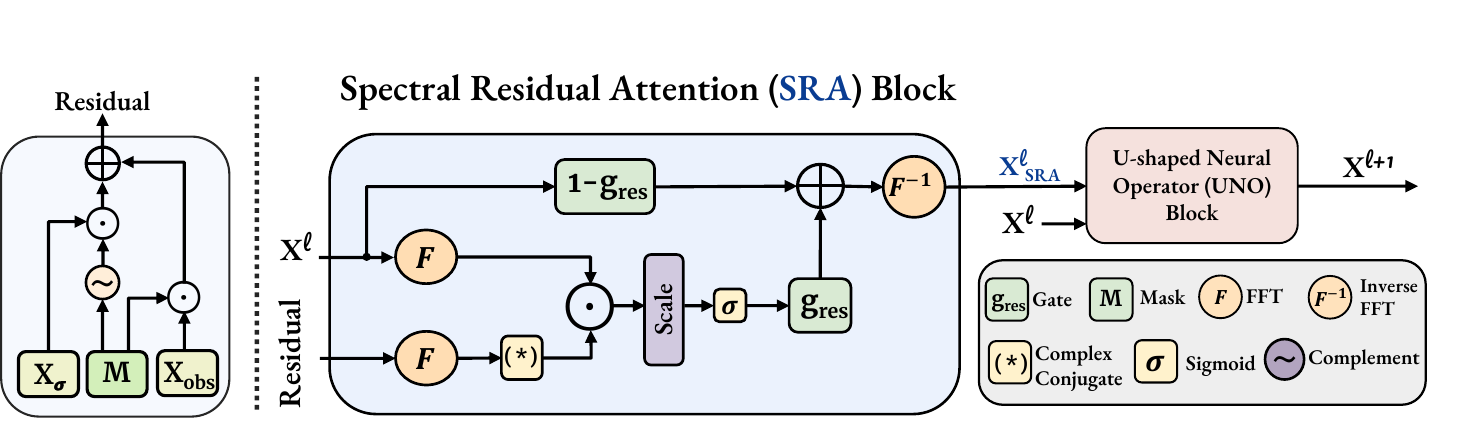}
     \small \caption{Overview of the PRISMA model architecture. \textbf{(Left)} Computation of the \textbf{\textit{observation-informed PDE residual}}: The model's current estimate ($\mathbf{x}_\sigma$) is mixed with the known observations ($\mathbf{x}_{\text{obs}}$) using the mask ($\mathbf{M_x}$) to produce a residual field. \textbf{(Right)} \textbf{\textit{Spectral Residual Attention (SRA) block}}: The SRA block operates in the Fourier domain to compute a physics-informed attention mask between the network state ($\mathbf{x}^l_t$) and the residual. This mask is applied in a gated skip-connection, controlled by a learned guidance strength ($g_{\text{res}}$), to produce a modulated state ($\mathbf{x}^{t}_{\text{SRA}}$) that is then passed to the UNO block.}
    \label{fig:sra_block}
\end{figure*}

Figure \ref{fig:sra_block} shows the architecture of the SRA block, the core mechanism for injecting physics-based architectural guidance in PRISMA.  SRA is inspired by cross-attention, and is implemented as a diagonal, frequency-wise spectral modulation rather than a full token-mixing attention layer. In particular, the feature spectrum acts as the query/value and the residual spectrum acts as the key, producing a per-frequency compatibility mask that modulates feature modes without performing dense mixing across all frequencies. This design preserves the key idea of attention-based query--key compatibility while remaining computationally efficient and well aligned with UNO layers.

Let $\widetilde{\bx}^{\,l}=\mathcal{F}(\bx^{l})$ and $\widetilde{\br}^{\,}=\mathcal{}(\br)$ be the 2D Fourier transforms of the layer's input feature maps and the corresponding PDE residuals, respectively. 
First, a compatibility score $S^l$ measures the phase alignment between them at each 2D frequency mode $\vec{k}=(k_1, k_2)$,  via their complex inner product. This is calculated as the magnitude of their complex inner product across all channels $C$. This score is then used to create a physics-informed \textit{attention mask} $A^l$ via learnable spectral gain weights $\vec{w}_{\text{gain}}^{\,l}$, and passing it through a sigmoid activation as follows:
\begin{align}
    S^{l}(\vec{k}) = \frac{1}{\sqrt{C}} \left| \sum\nolimits_{c=1}^{C}\,\widetilde{\bx}^{\,l}_{c}(\vec{k})\,\overline{\widetilde{\br}^{\,l}_{c}(\vec{k})} \right|,
A^{l}(\vec{k}) = \sigma\!\left(\vec{w}_{\text{gain}}^{\,l}(\vec{k}) \odot S^{l}(\vec{k})\right).
\end{align}

Unlike softmax, the sigmoid mask enables non-competitive activation across frequencies~\citep{ramapuram2025theory}. This allows the model to learn what frequency modes of the feature spectrum are most informative for error correction based on guidance from PDE residuals.  To dynamically control this process based on the relevance of PDE residuals in noisy settings, a scalar \textit{guidance strength} $g_{\text{res}}^l \in [0,1]$ is learned by an MLP from the spatially-averaged residual  $\br^l_{\text{avg}}$ and the diffusion noise embedding $c_\sigma$.  Finally, $g_{\text{res}}$ is used to modulate $\widetilde{\bx}^{\,}$ via a skip-connection with the attention mask to produce SRA's output:
{\small
\begin{align}
    g_{\text{res}} = \sigma\left( \mathrm{MLP}^{\,l}\big([\br^{\,l}_{\text{avg}},\; c_{\sigma}] \big) \right),
    \quad
    \sra(\bx^l, \br^l) = \mathcal{F}^{-1} \left( \left((1 - g_{\text{res}}^{l}) + g_{\text{res}}^{l} A^{l}\right) \odot \widetilde{\bx}^{\,l} \right)
    \label{eq:skip_connection_gating}
\end{align}
}

In the earlier stages of diffusion where noise level is high, $\mathbf{r}^l_{avg}$ is typically unreliable and enforcing them using hard constraints may make the training unstable. This is addressed by our scalar gate $g_{\text{res}}^l$ that down-weights PDE residual correction in noisy residual regimes. As the denoising trajectory progresses and solution becomes cleaner, $g_{\text{res}}^l$ increases, allowing stronger PDE-driven modulation through $A^l(k)$. In this way, PRISMA decouples residual guidance into a point-wise map $A^l(\vec{k})$ and a scalar signal $g_{\text{res}}^l$.
During training, this entire process of forward propagation is incorporated into EDM loss computation, teaching the denoiser to produce physically-consistent states. At inference, the model engages in a closed-loop, self-correcting process: it predicts a state, computes the residual based on that state, and then uses that residual to refine its own prediction in the next step of reverse diffusion trajectory. 
The complete training loop is detailed in Algorithm~\ref{alg:prisma_training_revised}; inference details and theoretical motivation are provided in Appendices~\ref{appendix:inference_algo} and~\ref{app:theory_sra}.

\noindent \textbf{PRISMA does not require task-specific training:} During training, observation masks $\bM=(\bM_a,\bM_u)$ are sampled from a fixed distribution of tasks to simulate different regimes (Unconditional, Full \& Sparse) in both directions (forward/inverse). Note that we randomly sample from this distribution of tasks and do not bias the training to any particular (e.g., target) task. The denoiser takes $(\bx_\sigma,\bx_{\text{obs}},\bM)$ as input, where $\bx_{\text{obs}}$ provides the observed PDE input/output values and $\bM$ indicates their locations. For sparse tasks, we sample an observation rate $p_{\text{obs}}\in[0.01,0.5]$ to generate masks. We
do not train on noisy observations, making this setting strictly out-of-distribution at test time.
Details of $p(task)$ and $p_{\text{obs}}$ in Appendix~\ref{appendix:implementation_details}.

\let\AND\relax
\begin{algorithm}
\small
\caption{\ourmethod Model Training.}
\label{alg:prisma_training_revised}
\begin{algorithmic}[1]
\REQUIRE Data distribution $\mu$, GRF covariance $\mathbf{C}$, noise distribution $p(\sigma)$, PDE operator $\mathcal{R}$, task distribution $p(task)$ and and sparse-mask distribution $p(\bM \mid task)$.
\STATE Initialize model parameters $\theta$.
\REPEAT
    \STATE Draw clean state $\mathbf{x}_0=[\mathbf{a}, \mathbf{u}] \sim \mu$ and noise level $\sigma \sim p(\sigma)$.
    \STATE Sample task type $t \sim p(task)$ (uncond, full-fwd, full-inv, sparse-fwd, sparse-inv).
    \STATE Sample masks $(\bM_a,\bM_u)\sim p(\bM\mid t)$.
    \COMMENT{full: $\bM\!=\!\mathbf{1}$;\; sparse: sample $p_{\text{obs}}$}
    \STATE Set observations $(\mathbf{a}_{\text{obs}}, \mathbf{u}_{\text{obs}}) \leftarrow (\mathbf{a}, \mathbf{u})$.
    \STATE Sample GRF noise $\mathbf{\eta} \sim \mathcal{N}(0, \sigma^2\mathbf{C})$.
    \STATE Construct noisy sample $\mathbf{x}_\sigma \leftarrow \mathbf{x}_0 + \mathbf{\eta}$. Let $\mathbf{x}_\sigma = [\mathbf{a}_\sigma, \mathbf{u}_\sigma]$.
    \STATE Compute guided residual $\mathbf{r} \leftarrow \mathcal{R}(\mathbf{M}_a \odot \mathbf{a}_{\text{obs}} + (1-\mathbf{M}_a) \odot \mathbf{a}_\sigma, \mathbf{M}_u \odot \mathbf{u}_{\text{obs}} + (1-\mathbf{M}_u) \odot \mathbf{u}_\sigma)$
    \STATE $\hat{\mathbf{x}}_0 \leftarrow D_\theta(\mathbf{x}_\sigma, \sigma, \mathbf{a}_{\text{obs}}, \mathbf{u}_{\text{obs}}, \mathbf{M}_a, \mathbf{M}_u, \mathbf{r})$. 
    \COMMENT{Compute denoised prediction}
    \STATE $L \leftarrow \lambda(\sigma)\|\hat{\mathbf{x}}_0 - \mathbf{x}_0\|^2_H$. \COMMENT{Compute training loss}
    \STATE Update parameters $\theta$ by minimizing $L$.
\UNTIL{converged}
\ENSURE Trained Denoiser $D_\theta$
\end{algorithmic}
\end{algorithm}

\section{Results}

\subsection{Experimental Setup}
\label{sec:experimental_setup}

\textbf{Dataset \& Tasks:} We validate our approach on five PDE problems from the dataset of \citet{huang2024diffusionpde}
We evaluate both forward and inverse problems in three regimes: 
(1) \textit{Full Observation}, with complete data; 
(2) \textit{Sparse Observation}, where only 3\% of the observed field is provided via a mask; and 
(3) an OOD \textit{Noisy} setting, where the conditioning observations are corrupted by additive Gaussian measurement noise (applied to input observations). Note that this noise is directly added to the observations and is different from the intrinsic noise used in the forward or reverse diffusion processes. We additionally evaluate generalization to unseen viscosities on Navier--Stokes. For Navier--Stokes, we formulate the task as \emph{one-step} vorticity prediction ($\omega_t \!\to\! \omega_{t+\Delta t}$) instead of final-time prediction ($\omega_0 \!\to\! \omega_T$), which enables a physically meaningful vorticity-transport residual for guidance (Appendix~\ref{appendix:detailed_dataset_desc}). These noisy-observation and viscosity-shift evaluations serve as deployment-motivated stress tests beyond standard full/sparse settings.

\textbf{Baselines \& Evaluation Metrics:} We compare against deterministic solvers including FNO \citep{fno}, PINO \citep{pino}, 
and diffusion-based solvers DiffusionPDE \citep{huang2024diffusionpde} and FunDPS \citep{fundps}, which perform diffusion posterior sampling via inference-time loss guidance. Performance is measured using relative $L_2$ error for all tasks (and classification error rate for Darcy inverse). Since DPS-based methods can be sensitive to guidance hyperparameters, we use the authors' released implementations and their recommended guidance settings \emph{without per-task retuning} to ensure a fair and reproducible comparison. Likewise, \ourmethod is trained once using masks sampled from a fixed, diverse distribution over observation patterns and problem directions, and is \emph{not} tuned to the specific patterns used at test time. We treat the \textit{Noisy} regime and viscosity shifts as deployment-motivated tests where the corruption structure and physical parameters may differ from training and may not be known in advance, making additional tuning impractical. Due to reported variability in reproduced results for some diffusion baselines, we include numbers reproduced from the authors’ codebases and checkpoints where applicable. Additional details in Appendix~\ref{appendix:implementation_details}.

\subsection{Results on Noisy Observations}

We first evaluate \ourmethod's robustness in a challenging noisy observation setting that mimics
real-world measurement corruption. Importantly, \emph{no method is trained on noisy observations}:
all models are trained on clean data, and noise is introduced only at test time, making this a strict
out-of-distribution (OOD) evaluation. Table~\ref{tab:noisy_main} shows that \ourmethod\ achieves
state-of-the-art performance across most equations; for Darcy Flow forward, \ourmethod\ with
50 steps obtains 12.28\% relative $L_2$ error, compared to 49.18\% for DiffusionPDE and 55.09\%
for FunDPS. We observe similar trends across other equations. Appendix~\ref{app::spectral_power} further shows that \ourmethod\ better preserves ground-truth spectral power under noise, while Appendix~\ref{appendix:noise_percent} and ~\ref{appendix:sparsity_robutness}  evaluate robustness across varying noise-corruption percentages and random sparsity levels, respectively.

\begin{table*}[t]
    \centering
    \caption{Comparison of different models on PDE problems with noise applied to all observations (100\% coverage), using unit-variance Gaussian corruption to simulate measurement noise. Results are reported as relative $L_2$ error for all tasks and error rate for Darcy inverse.}
    \setlength{\tabcolsep}{0.25em}
    \resizebox{\textwidth}{!}{%
    \renewcommand{\arraystretch}{1.2}
    \begin{tabular}{l c c c c c c c c c c c c c c}
    \toprule
      & \multirow{2}{*}{\textbf{Steps} $(N)$} 
      & \multirow{2}{*}{\shortstack{\textbf{Inference}\\\textbf{Time (s)}}}
      & \multicolumn{2}{c}{\textbf{Darcy Flow}} 
      & \multicolumn{2}{c}{\textbf{Poisson}} 
      & \multicolumn{2}{c}{\textbf{Helmholtz}}
      & \multicolumn{2}{c}{\textbf{Navier--Stokes}}
      & \multicolumn{2}{c}{\textbf{Kolmogorov Flow}}
      & \multirow{2}{*}{\textbf{Avg. Rank} $\downarrow$} \\
    \cmidrule(lr){4-5}
    \cmidrule(lr){6-7}
    \cmidrule(lr){8-9}
    \cmidrule(lr){10-11}
    \cmidrule(lr){12-13}
      & & 
      & \textbf{Forward} & \textbf{Inverse} 
      & \textbf{Forward} & \textbf{Inverse} 
      & \textbf{Forward} & \textbf{Inverse} 
      & \textbf{Forward} & \textbf{Inverse}
      & \textbf{Forward} & \textbf{Inverse}
      & \\
    \midrule
    
    \textbf{FNO}  & $-$ & $0.1$ &
    15.70\% & 52.3\% 
    & 25.05\% & 3.0e7\% 
    & 264.8\% & 6.1e5\% 
    & 140.9\% & 772.03\%
    & 210.8\% & 220.59\% 
    & 5.7 \\
    
    \textbf{PINO} & $-$ & $0.11$ &
    190.40\% & 52.3\% 
    & 26.64\% & 2.3e6\% 
    & 449.1\% & 1.7e6\% 
    & 84.61\% & 359.05\%
    & 173.81\% & 243.84\%
    & 6 \\
    
    \midrule

    \textbf{DiffusionPDE} & 2000 & 213.0 &
    49.18\% & 70.08\% 
    & 44.44\% & 130.01\% 
    & 30.97\% & 119.52\% 
    & 109.00\% & 46.00\%
    & 158.95\% & 170.53\%
    & 4.6 \\

    \textbf{FunDPS} & 200 & 4.72 &
    26.9\% & 49.39\% 
    & 78.36\% & 1491.36\% 
    & 54.99\% & 629.43\% 
    & \textbf{31.55\%} & 38.44\%
    & 122.37\% & 125.86\%
    & 3.6  \\
    
    \textbf{FunDPS} & 500 & 11.8 &
    55.09\% & 49.62\% 
    & 120.0\% & 1772.23\% 
    & 40.31\% & 695.49\% 
    & 37.96\% & 46.25\%
    & 128.65\% & 127.38\%
    & 4.8 \\

    \midrule
    
    \textbf{PRISMA (ours)} & 20 & \textbf{0.18} &
    \underline{12.29\%} & \underline{25.3\%} 
    & \underline{16.55\%} & \underline{41.86\%} 
    & \underline{14.19\%} & \underline{49.23\%} 
    & 36.80\% & \underline{34.44\%}
    & \underline{95.35\%} & \underline{91.47\%}
    & \underline{2.00} \\
    
    \textbf{PRISMA (ours)} & 50 & \underline{0.8} &
    \textbf{12.28\%} & \textbf{25.3\%} 
    & \textbf{15.84\%} & \textbf{41.38\%} 
    & \textbf{13.49\%} & \textbf{48.53\%} 
    & \underline{36.34\%} & \textbf{34.13\%}
    & \textbf{93.96\%} & \textbf{90.39\%}
    & \textbf{1.1} \\
    
    \bottomrule
    \end{tabular}
    }
    \label{tab:noisy_main}
\end{table*}

\begin{figure*}[t]
    \centering
    \begin{subfigure}[t]{0.49\linewidth}
        \centering
        \includegraphics[width=\linewidth]{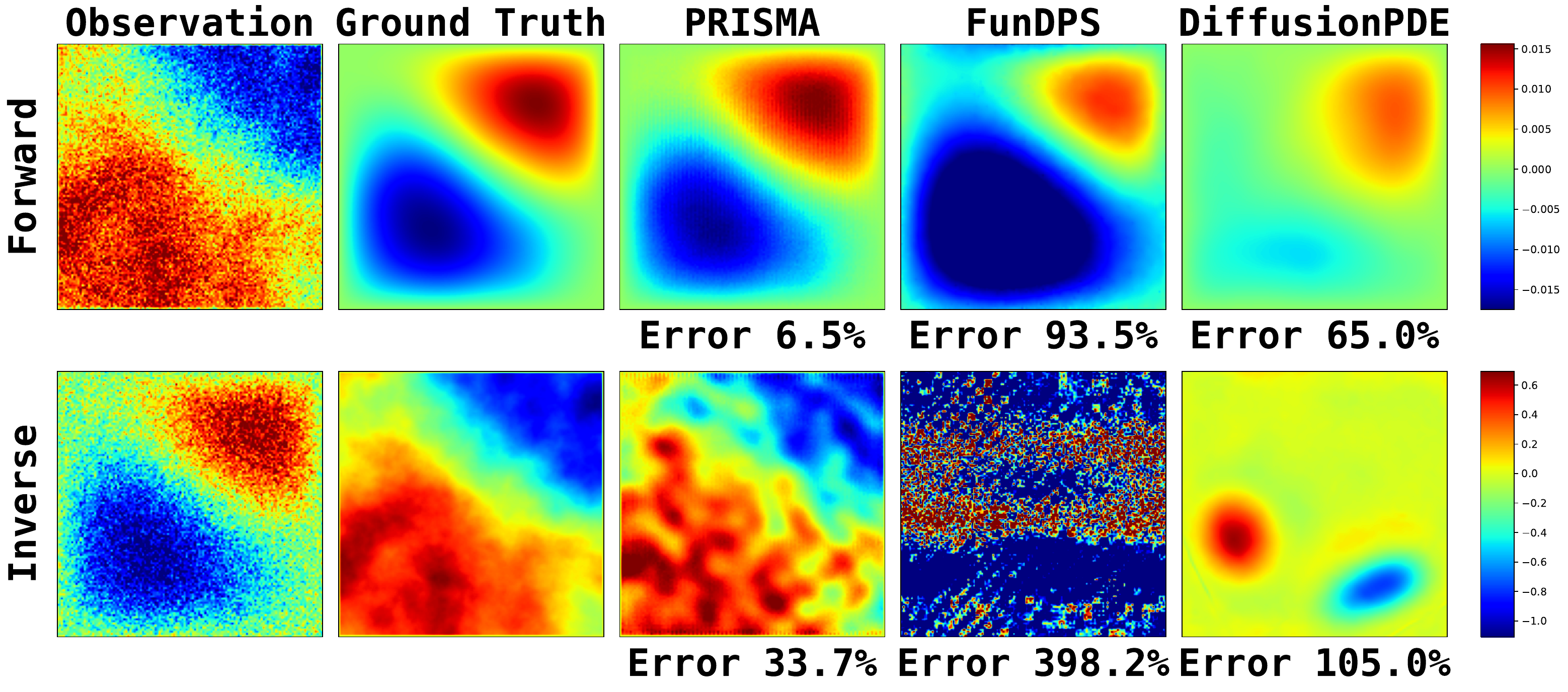}
        \vspace{-1ex}
        \caption{Helmholtz. \textit{Noisy Observation}}
        \label{fig:helm_noisy}
    \end{subfigure}
    \hfill
    \begin{subfigure}[t]{0.49\linewidth}
        \centering
        \includegraphics[width=\linewidth]{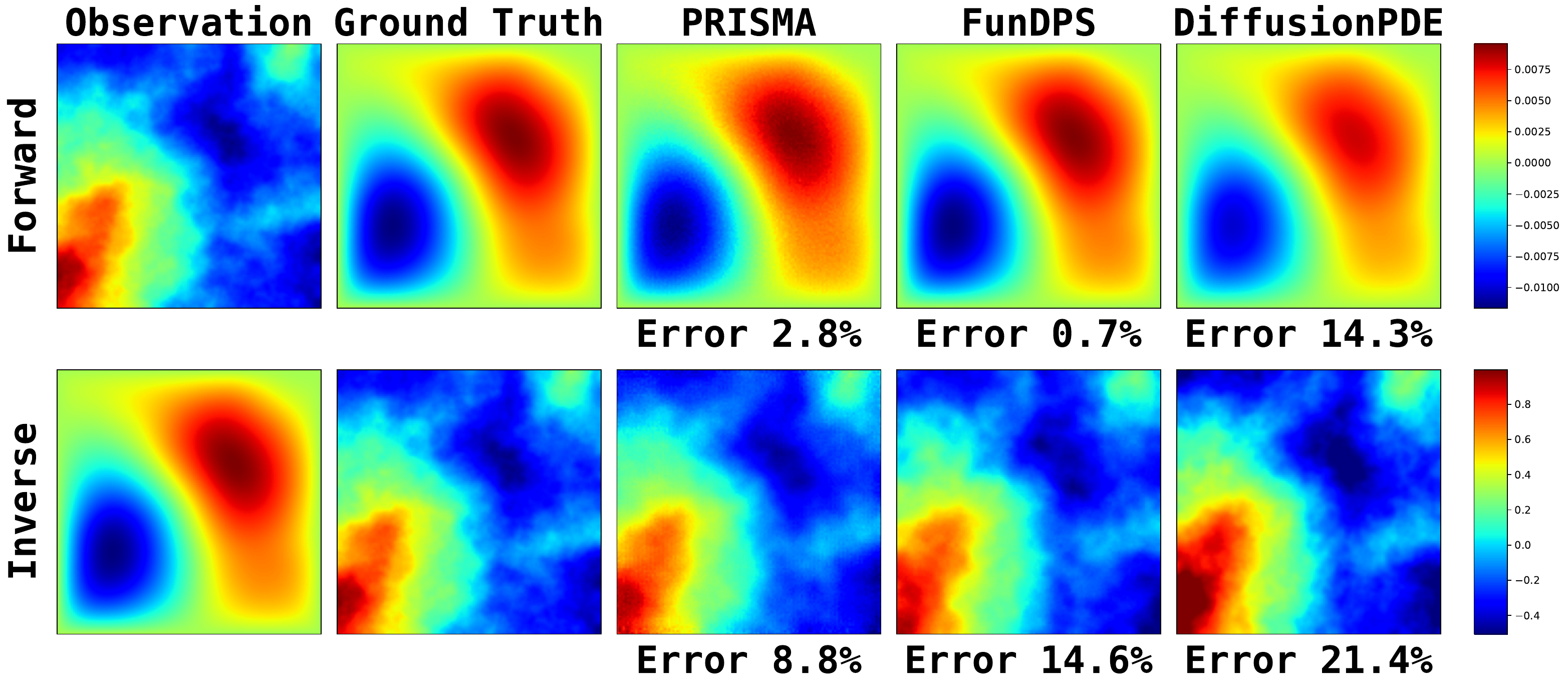}
        \vspace{-1ex}
        \caption{Poisson. \textit{Full Observation}}
        \label{fig:nsnb_full}
    \end{subfigure}

    \vspace{1ex}
    \begin{subfigure}[t]{0.49\linewidth}
        \centering
        \includegraphics[width=\linewidth]{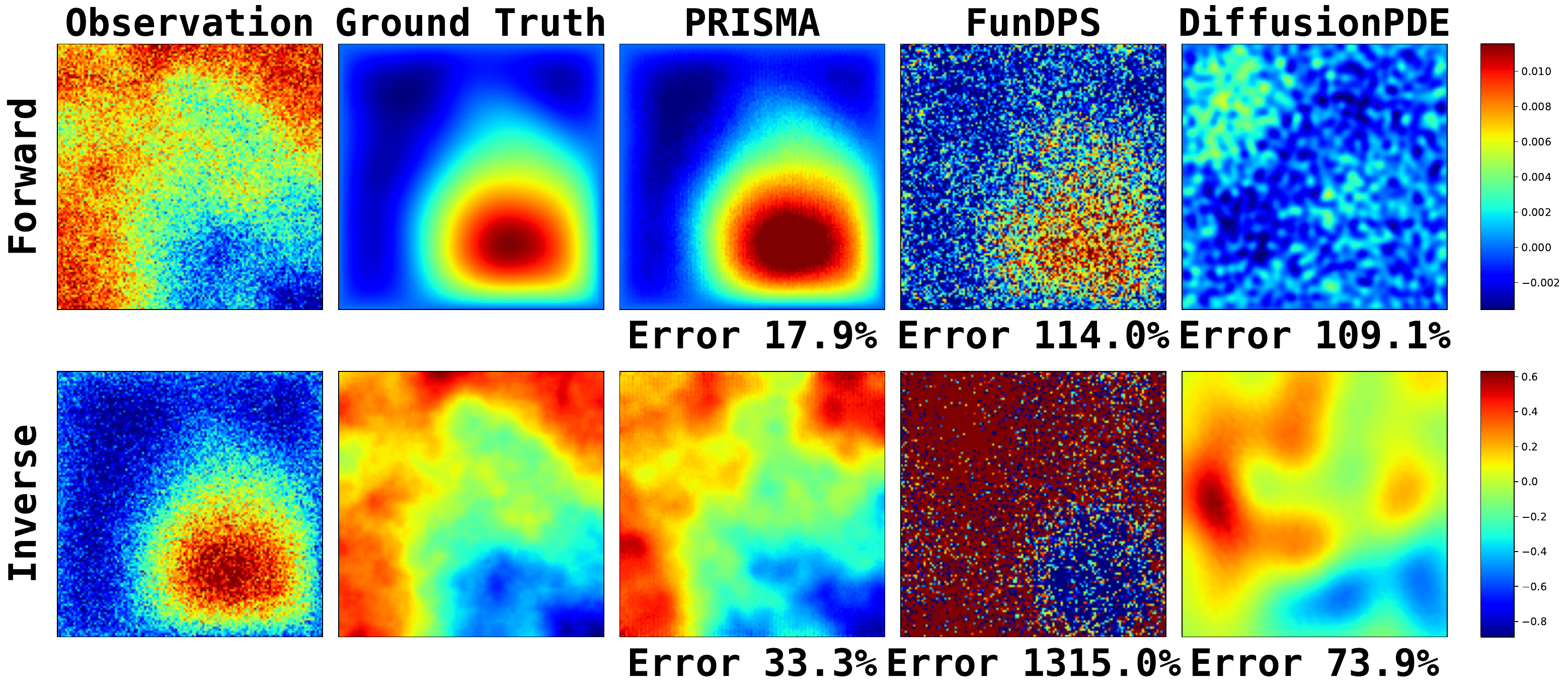}
        \vspace{-1ex}
        \caption{Poisson. \textit{Noisy Observation}}
        \label{fig:poisson_noisy}
    \end{subfigure}
    \hfill
    \begin{subfigure}[t]{0.49\linewidth}
        \centering
        \includegraphics[width=\linewidth]{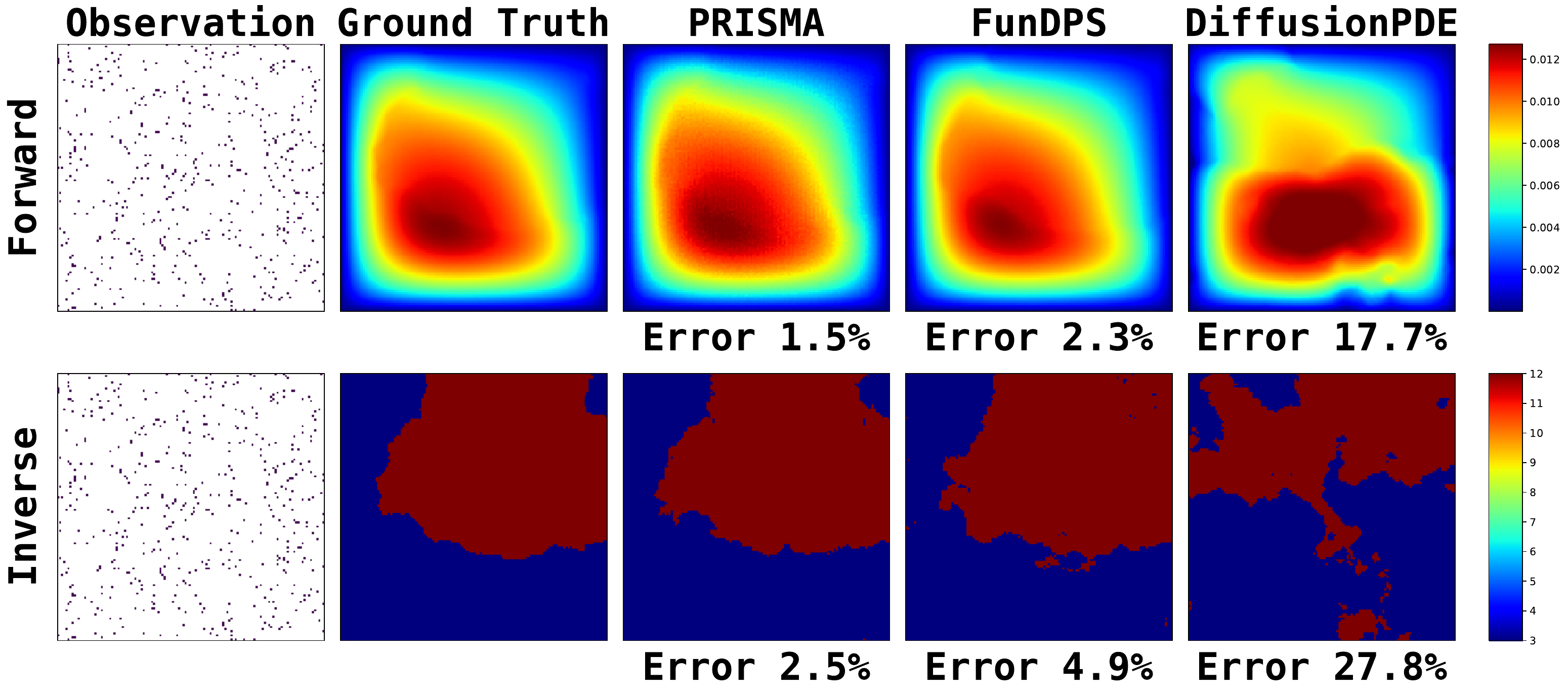}
        \vspace{-1ex}
        \caption{Darcy Flow. \textit{Sparse Observations}}
        \label{fig:darcy_sparse}
    \end{subfigure}

    \vspace{-1ex}
    \caption{\small Qualitative results for PRISMA and baseline models on three PDE benchmarks. We evaluate performance under three distinct conditions: (a) noisy observations (corrupted by $\mathcal{N}(0,1)$ Gaussian noise), (b) full, clean observations, (c) noisy observations, (d) sparse observations (3\% of data known). Relative $\ell_2$ error is reported below each prediction (pixel-wise error rate for the Darcy inverse case).}
    \label{fig:viz_comparison}
    \vspace{-2ex}
\end{figure*}

\subsection{Inference Efficiency}

\begin{figure}[h]
    \centering
    \includegraphics[width=0.55\textwidth]{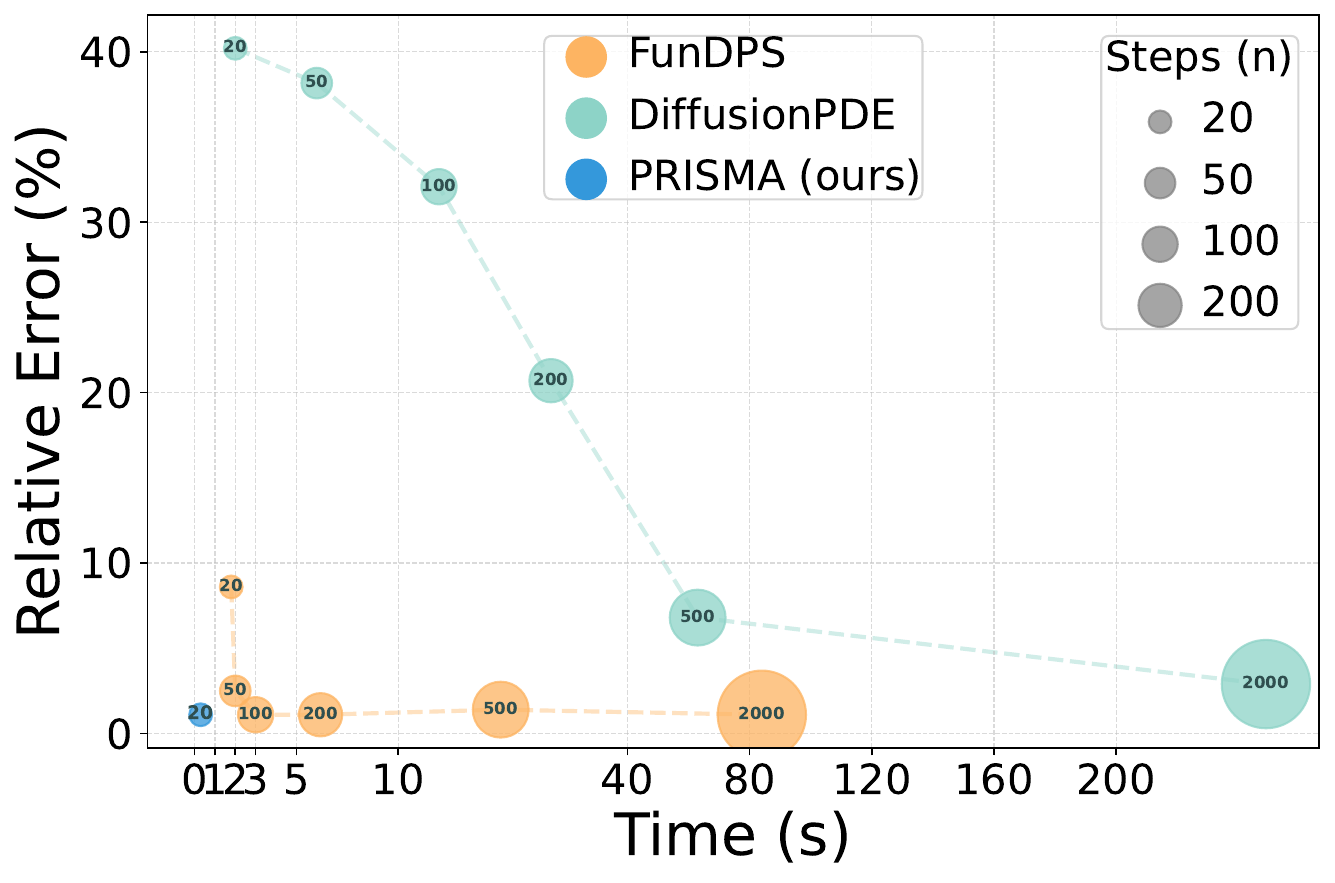}
    \caption{Inference time versus accuracy trade-off on the Darcy flow forward problem under full observations. Point size corresponds to the number of denoising steps.}
    \label{fig:time_error}
\end{figure}


A key advantage of \ourmethod is its significant inference efficiency while   
having competitive accuracy, compared to the current state-of-the-art. As shown in Table~\ref{tab:noisy_main} and visualized in Figure~\ref{fig:time_error}, \ourmethod occupies the optimal low-error, low-time quadrant. We attribute this inference speedup as a direct result of our unified conditioning framework and the strong physical guidance from the SRA block. This enables a {gradient-descent} free sampling process during inference, whereas DiffusionPDE and FunDPS rely on computationally expensive PDE-based inference sampling. \ourmethod converges in just \textbf{20-50 steps}, while DiffusionPDE and FunDPS require 200 to 2000 steps. When they are also restricted to 20 steps, their performance degrades significantly
as shown in Table~\ref{tab:abla_20steps_combined}. \ourmethod has an overall inference speed-up of \textbf{15x to 250x} (sec/sample) compared to competing diffusion-based PDE solvers. We also show in Appendix \ref{app:compute_cost} that PDE-residual computation in \ourmethod is a small fraction (1.16$\%$) of the overall per-step sampling wall-clock time.

\begin{table*}[!t]
    \centering
    \fontsize{8pt}{8pt}\selectfont
    \renewcommand{\arraystretch}{1.2} 
    \caption{\small Comparison under varying viscosity with Noisy / Sparse / Full observations (in $L_2$ relative error). \textbf{Best} is boldened, \underline{second-best} underlined (computed within each viscosity block).  }
    \setlength{\tabcolsep}{0.5em}
    \begin{tabular}{l c c c c c c c c}
    \toprule
      & \multirow{2}{*}{\textbf{Steps} $(N)$} & \multicolumn{2}{c}{\textbf{Noisy}}
      & \multicolumn{2}{c}{\textbf{Sparse}}
      & \multicolumn{2}{c}{\textbf{Full}}
      & \multirow{2}{*}{\shortstack{\textbf{Viscosity}\\$\boldsymbol{\nu}$}} \\
    \cmidrule(lr){3-4}
    \cmidrule(lr){5-6}
    \cmidrule(lr){7-8}
      & & \textbf{Forward} & \textbf{Inverse} & \textbf{Forward} & \textbf{Inverse} & \textbf{Forward} & \textbf{Inverse} & \\
    \midrule
   
    \textbf{DiffusionPDE} & 2000 &
    114.00\% & 51.12\% & \underline{8.53\%} & \underline{10.54\%} & \textbf{3.16\%} & \underline{5.99\%} & $1\times 10^{-4}$ \\

    \textbf{FunDPS} & 200 &
    \underline{46.00\%} & \underline{44.60\%} & 223.57\% & 102.39\% & 29.74\% & 27.69\% & $1\times 10^{-4}$ \\

    \textbf{PRISMA (ours)} & 20 &
    \textbf{41.73\%} & \textbf{34.17\%} & \textbf{8.38\%} & \textbf{9.56\%} & \underline{3.89\%} & \textbf{5.69\%} & $1\times 10^{-4}$ \\

    \midrule

    \textbf{DiffusionPDE} & 2000 &
    94.70\% & 51.30\% & \underline{8.60\%} & \underline{10.65\%} & \textbf{3.60\%} & \textbf{5.42\%} & $1\times 10^{-5}$ \\

    \textbf{FunDPS} & 200 &
    \underline{44.90\%} & \underline{44.11\%} & 244.44\% & 119.68\% & 29.12\% & 27.00\% & $1\times 10^{-5}$ \\

    \textbf{PRISMA (ours)} & 20 &
    \textbf{42.21\%} & \textbf{34.45\%} & \textbf{8.12\%} & \textbf{9.72\%} & \underline{4.11\%} & \underline{6.20\%} & $1\times 10^{-5}$ \\

    \bottomrule
    \end{tabular}
    \label{tab:vis_sweep_full_sparse_noisy}
\end{table*}

\vspace{-1ex}

\subsection{Results on Physics Extrapolation}

We study generalization to unseen viscosities on Navier--Stokes by evaluating models on datasets generated with viscosities different from the training viscosity ($\nu{=}10^{-3}$; $\mathrm{Re}{=}1000$). Specifically, all methods are trained only at $\nu{=}10^{-3}$ and evaluated using the authors’ released checkpoints and default inference settings, without additional tuning on lower viscosities ($\nu\in{10^{-4},10^{-5}}$, corresponding to $\mathrm{Re}\in{10^{4},10^{5}}$). Table~\ref{tab:vis_sweep_full_sparse_noisy} summarizes performance across full, sparse, and noisy observation regimes. Overall, \ourmethod remains competitive under viscosity shifts while requiring substantially fewer sampling steps. We further evaluate Helmholtz wavenumber extrapolation from the training setting $k{=}1$ to
$k\in\{2,3,4\}$ in Appendix~\ref{appendix:phys_helm_wave}.

\begin{figure}[t]
    \centering

    \begin{subfigure}[t]{0.48\linewidth}
        \centering
        \includegraphics[width=\linewidth]
        {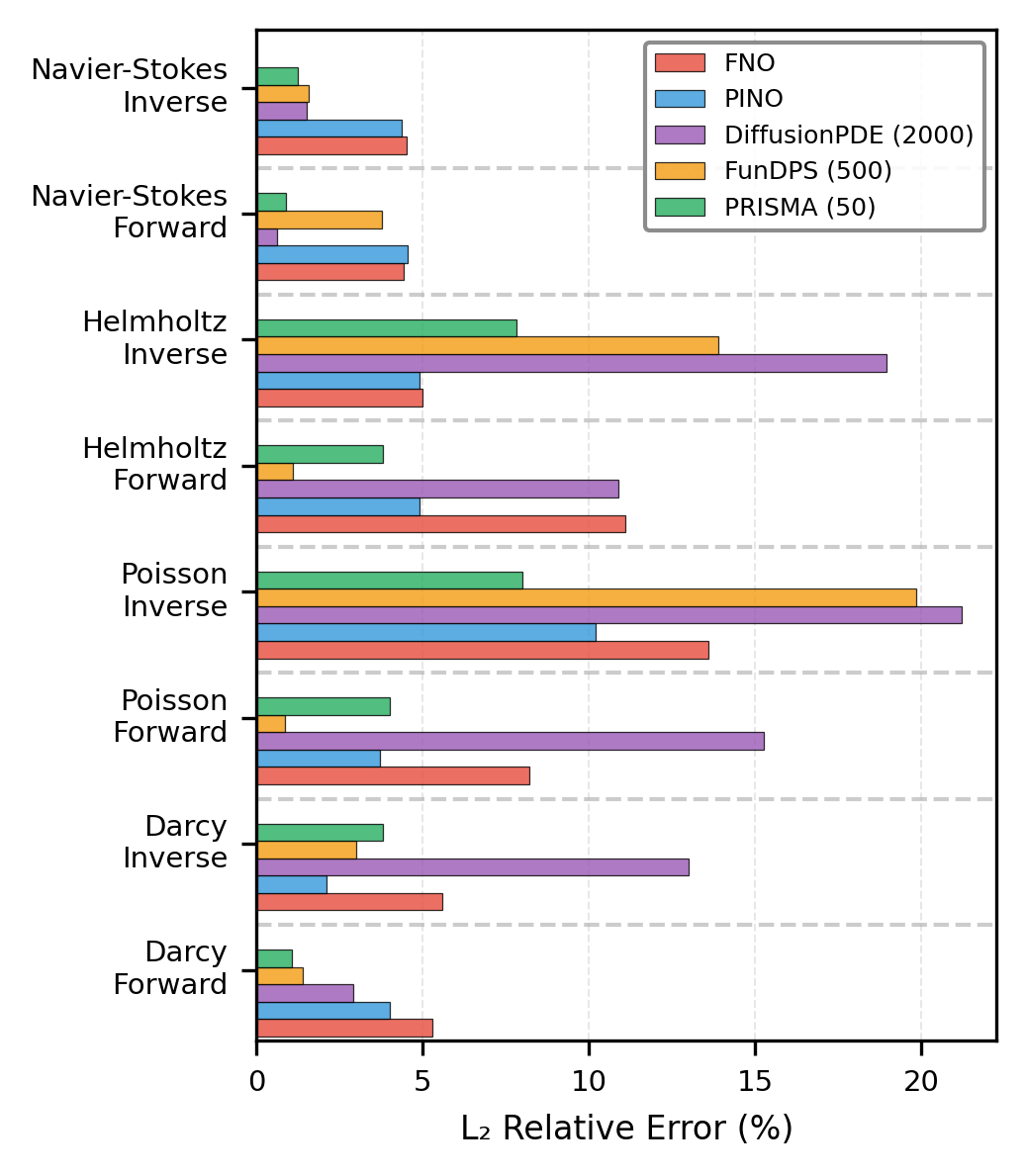}
        \caption{Full observations.}
        \label{fig:full_obs_bar_plot}
    \end{subfigure}
    \hfill
    \begin{subfigure}[t]{0.48\linewidth}
        \centering
        \includegraphics[width=\linewidth]
        {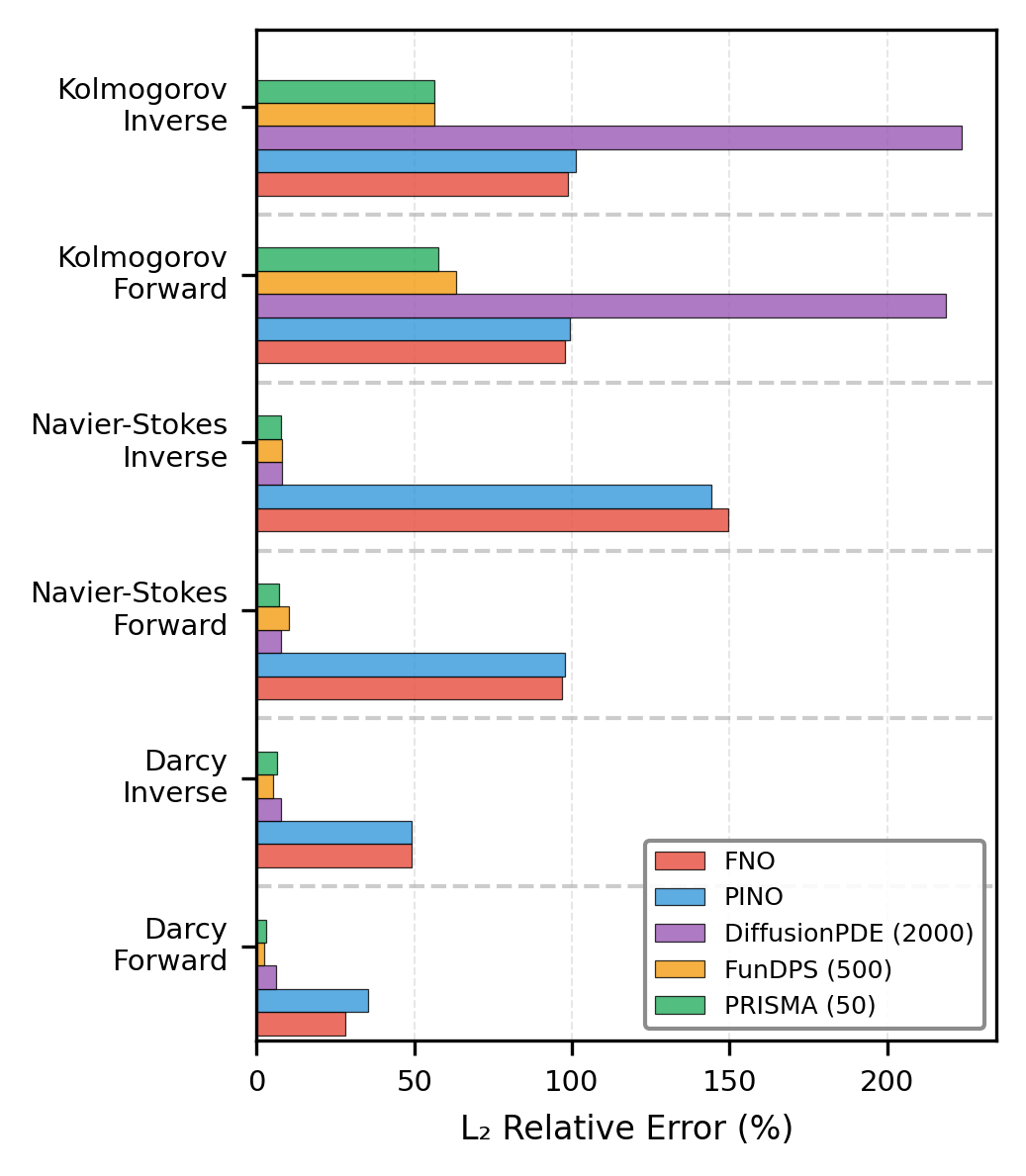}
        \caption{Sparse observations.}
        \label{fig:sparse_obs_bar_plot}
    \end{subfigure}

    \caption{Comparison of model performance under different observation settings: 
    \subref{fig:full_obs_bar_plot} full observations and 
    \subref{fig:sparse_obs_bar_plot} sparse observations.}
    \label{fig:observation_comparison}
\end{figure}
\subsection{Results on Full and Sparse Observations}

\ourmethod\ also demonstrates competitive performance under full and sparse observation settings (Appendix Tables~\ref{tab:full_main}, \ref{tab:sparse_main}), while requiring far fewer sampling steps. 
Representative comparisons in Figures ~\ref{fig:full_obs_bar_plot} and \ref{fig:sparse_obs_bar_plot} show that deterministic operators are competitive under {full} observations but degrade sharply under {sparse} conditioning, whereas diffusion-based baselines are more robust.  Across both settings, PRISMA achieves the best average rank while using only 20--50 steps.
For qualitative comparison, we provide key visualizations in Figure~\ref{fig:viz_comparison}, with extensive results available in Appendix~\ref{appendix:qualitative}. Additional evaluations on Burgers, 2D+time Navier--Stokes
rollout, and super-resolution are provided in Appendices~\ref{appendix:spatio-temporal} \& \ref{appendix:super-res}.

\begin{table*}[h]
    \centering
    \fontsize{8pt}{8pt}\selectfont
    \caption{{\small Ablation of residual strategies across Helmholtz and Kolmogorov (noisy, sparse, full) for both forward/inverse problems ($L_2$ relative error, shown as percentages). All results are on 64$\times$64 resolution with 20 steps.}}
    \setlength{\tabcolsep}{0.35em}
    \renewcommand{\arraystretch}{0.8}
    \resizebox{\textwidth}{!}{%
    {\color{black}
    \begin{tabular}{l l c c c c c c}
    \toprule
    \multicolumn{2}{c}{\textbf{PRISMA}} & \multicolumn{2}{c}{\textbf{Noisy}}
      & \multicolumn{2}{c}{\textbf{Sparse}}
      & \multicolumn{2}{c}{\textbf{Full}} \\
    \cmidrule(lr){1-2}
    \cmidrule(lr){3-4}
    \cmidrule(lr){5-6}
    \cmidrule(lr){7-8}
    & & \textbf{Forward} & \textbf{Inverse}
      & \textbf{Forward} & \textbf{Inverse}
      & \textbf{Forward} & \textbf{Inverse} \\
    \midrule
    \multirow{3}{*}{\textbf{Helmholtz}}
    & \textit{(w/o PDE res)} &
    $\textbf{29.0}_{\textbf{0.20}}\textbf{\%}$ & $111.87_{2.12}\%$ &
    $36.2_{0.15}\%$ & $66.75_{0.19}\%$ &
    $\textbf{3.15}_{\textbf{0.05}}\textbf{\%}$ & $15.67_{0.34}\%$ \\
    & \textit{(PDE res with concat)} &
    $34.5_{1.9}\%$ & $97.09_{1.32}\%$ &
    $94.2_{0.5}\%$ & $124.7_{0.24}\%$ &
    $28.7_{0.9}\%$ & $51.55_{5.75}\%$ \\
    & \textit{(PDE res with SRA) \textbf{(ours)}} &
    $30.35_{0.45}\%$ & $\textbf{91.85}_{\textbf{1.35}}\textbf{\%}$ &
    $\textbf{30.33}_{\textbf{0.12}}\textbf{\%}$ & $\textbf{60.96}_{\textbf{0.09}}\textbf{\%}$ &
    $3.34_{0.14}\%$ & $\textbf{12.47}_{\textbf{0.11}}\textbf{\%}$ \\
    \midrule
    \multirow{3}{*}{\textbf{Kolmogorov}}
    & \textit{(w/o PDE res)} &
    $125.67\%$ & $126.05\%$ &
    $67.47\%$ & $66.28\%$ &
    $11.12\%$ & $10.98\%$ \\
    & \textit{(PDE res with concat)} &
    $120.42\%$ & $118.16\%$ &
    $64.06\%$ & $63.08\%$ &
    $11.76\%$ & $12.10\%$ \\
    & \textit{(PDE res with SRA) \textbf{(ours)}} &
    $\textbf{118.47}\textbf{\%}$ & $\textbf{105.73}\textbf{\%}$ &
    $\textbf{63.10}\textbf{\%}$ & $\textbf{62.33}\textbf{\%}$ &
    $\textbf{10.85}\textbf{\%}$ & $\textbf{10.91}\textbf{\%}$ \\
    \bottomrule
    \end{tabular}
    }}
    \label{tab:main_ablation}
    \vspace{-1ex}
\end{table*}

\subsection{Analysis of Model Robustness and Residuals}
\label{subsec:analysis}

\textbf{Impact of PDE Residual Guidance: } A key design choice in \ourmethod is the architectural integration of PDE residuals via the SRA block. To justify this choice, we compare three variants: (1)~\textbf{No PDE Residual}, a baseline without physics guidance; (2)~\textbf{Concatenation}, a simple channel-wise PDE residual concatenation; and (3)~\textbf{PDE Residual with SRA}, where the PDE residual is input to the SRA block. Table~\ref{tab:main_ablation} highlights that the PDE residual with SRA block outperforms both simple channel-wise concatenation of the PDE residual and an unguided baseline. We further ablate the internal design of the SRA block itself in Appendix~\ref{sec:app:sra_ablatioons}.

\begin{figure}[t]
    \centering
    \begin{subfigure}[t]{0.48\linewidth}
        \centering
        \includegraphics[height=4.2cm]{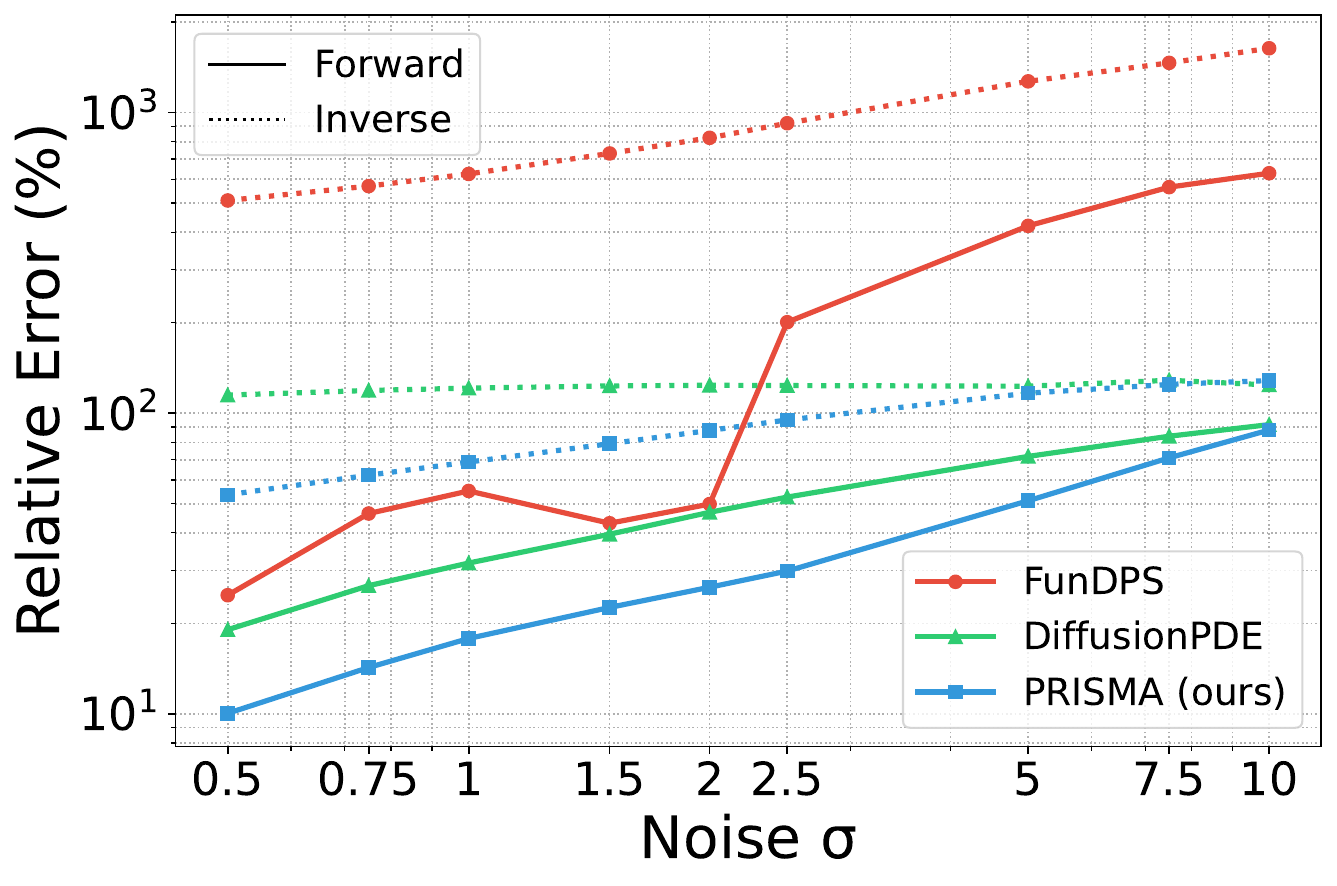}
        \caption{\small Relative error vs. input noise level ($\sigma$) for the Helmholtz fwd (solid lines) and inv (dotted lines).}
        \label{fig:noise_sigma_vary}
    \end{subfigure}
    \hfill
    \begin{subfigure}[t]{0.48\linewidth}
        \centering
        \includegraphics[height=4.2cm]{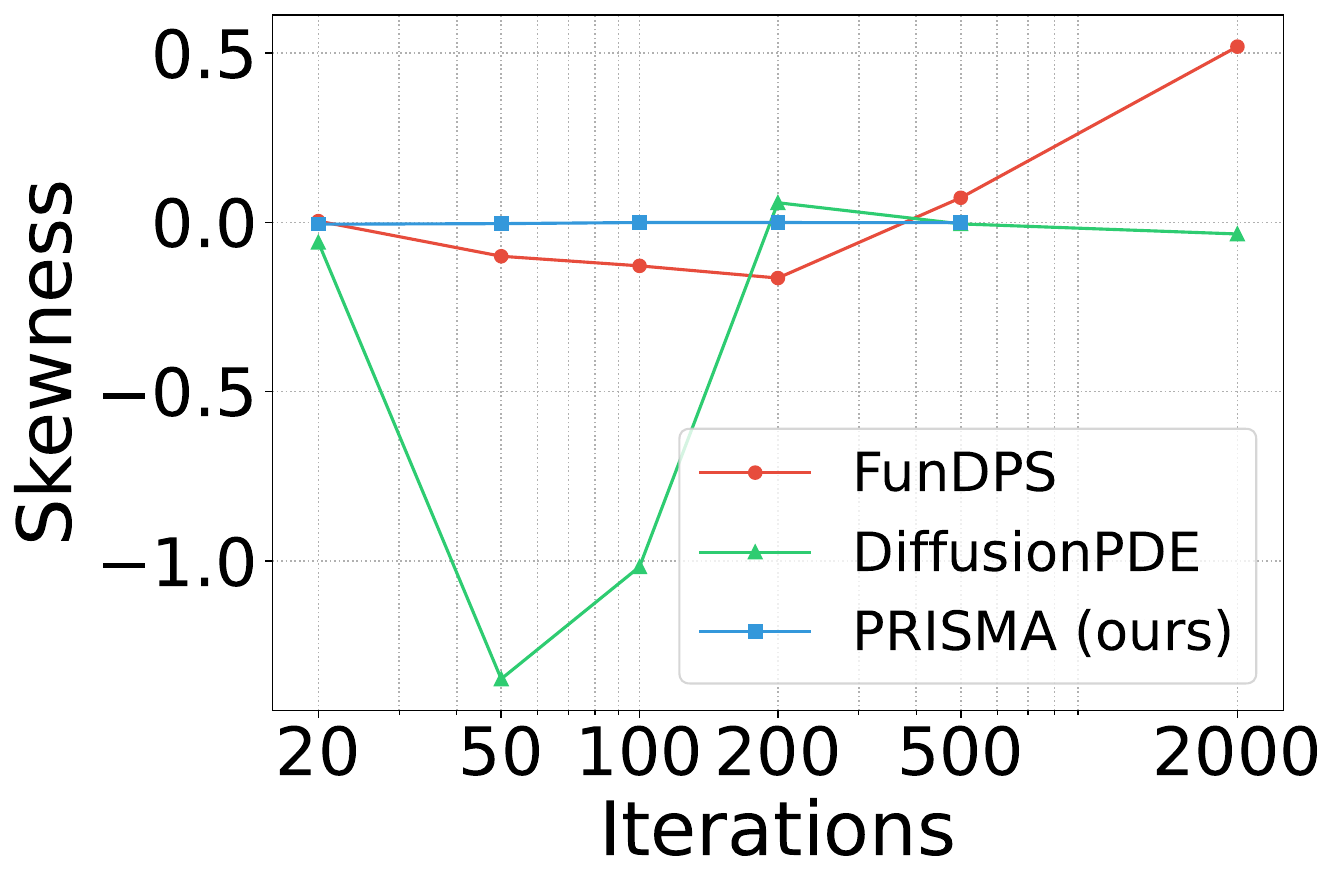}
        \caption{\small Skewness of the PDE residual field against inference iterations for Poisson (fwd).}
        \label{fig:noise_skewness}
    \end{subfigure}
    \caption{ Analysis of PRISMA's physical fidelity and robustness. \subref{fig:noise_sigma_vary} Error degrades gracefully with increasing observation noise, unlike baselines. \subref{fig:noise_skewness} PDE residual skewness stays near zero across inference iterations, indicating unbiased errors.}
    \label{fig:robustness_analysis}
\end{figure}

\textbf{Robustness to Varying Noise Levels}: We tested \ourmethod on the Helmholtz equation with varying intensities of Gaussian noise ($\sigma$). As plotted in Figure~\ref{fig:noise_sigma_vary}, \ourmethod consistently maintains a lower error rate than FunDPS and DiffusionPDE across all noise levels for both forward and inverse tasks. Notably, while the error for baseline models increases sharply with noise, PRISMA's performance degrades gracefully, highlighting its enhanced stability in high-noise regimes.

\textbf{Fidelity to Physical Constraints:} We further analyze skewness and kurtosis of residual fields to assess whether residual errors are biased or heavy-tailed. Figure~\ref{fig:noise_skewness} shows the skewness residual statistics for the Poisson forward problem. PRISMA's residuals (blue) exhibit near-zero skewness across different inference sampling iterations. We observe a similar trend for kurtosis (Appendix Figure \ref{fig:noise_kurtosis}).

\noindent\textbf{Posterior Diversity and Uncertainty Calibration:} To verify PRISMA is not collapsing to a deterministic point predictor, we sample the Darcy inverse problem 16 times from independent latents under identical conditioning, across full, sparse, and noisy observations (50 held-out test samples per mode). Ensemble spread (per-pixel standard deviation across members,
averaged per sample) increases with observation difficulty (0.32 $\rightarrow$ 0.52 $\rightarrow$ 1.45, full to noisy) and correlates with per-sample error ($r=0.85$/$0.81$/$0.46$), confirming meaningful, well-calibrated posterior diversity rather than a disguised deterministicmap. Full details are in Appendix~\ref{app:uncertainty}.

\subsection{Generalization to Geometric Boundary Recovery: Eikonal SDF}
\label{subsec:eikonal_main}
Beyond scalar physical fields defined on regular domains, we test whether \ourmethod\ generalizes to recovering geometric boundary structure, using an Eikonal signed distance function (SDF) benchmark with irregular boundaries~\citep{daw2023mitigating}. Full problem definition and evaluation protocol are provided in Appendix~\ref{app:eikonal}.

\begin{figure}[h]
    \centering
    \includegraphics[width=0.8\linewidth]{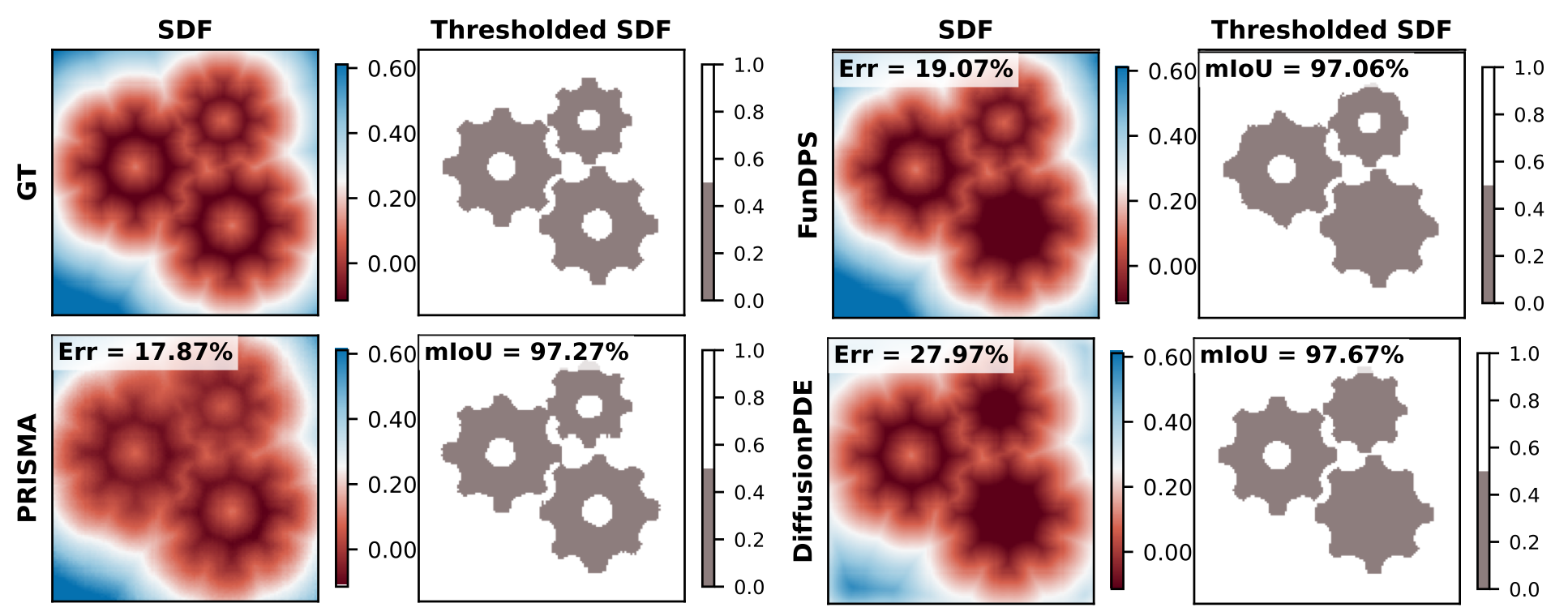}
    \vspace{-1ex}
    \caption{ Qualitative comparison on an Eikonal test sample. Left: predicted
    signed distance field (SDF); right: binary reconstruction obtained by thresholding
    the SDF. We compare ground truth against PRISMA, FunDPS, and DiffusionPDE. For the
    SDF panels, we report relative $L_2$ error; for the thresholded reconstructions, we
    report mIoU. A second qualitative test sample is shown in
    Appendix~\ref{app:eikonal}, Figure~\ref{fig:eikonal_91}.}
    \label{fig:eikonal_main}
\end{figure}

\ourmethod\ attains 6.87\% relative $L_2$ error with only 50 steps, and 6.97\% with 20 steps, outperforming FunDPS (7.77\%, 200 steps) and DiffusionPDE (35.83\%, 2000 steps), while requiring an order of magnitude fewer sampling steps than both baselines (Table~\ref{tab:eikonal_full_forward}). Figure~\ref{fig:eikonal_main} shows a representative qualitative comparison: \ourmethod\ recovers sharper boundary detail than DiffusionPDE and achieves comparable thresholded-boundary mIoU to both baselines, despite
using far fewer inference steps.

\vspace{-1ex}
\section{Conclusion \& Future Work}
\label{sec:conclusion}

We introduced PRISMA, a novel conditional diffusion neural operator that fundamentally challenges the prevailing paradigm of using PDE residuals as external loss terms for guidance. Our approach centers on architectural guidance, embedding physical constraints directly into the model as learnable features via a novel Spectral Residual Attention (SRA) block. 
This design enables two critical advantages: first, it facilitates an entirely {gradient-descent free} sampling process during inference, eliminating slow and unstable test-time optimization. Second, it allows the creation of a single, unified model capable of seamlessly solving both forward and inverse problems across the full spectrum of full, sparse, and noisy observation regimes. Our experiments validate the effectiveness of our approach, demonstrating inference speedups of 15x to 250x (seconds per sample) relative to state-of-the-art diffusion-based methods.

\paragraph{Limitations and Future Directions.}
Like most existing diffusion-based PDE solvers, PRISMA is currently implemented for PDE fields sampled on regular grids in 2D space.  While the core idea is not restricted to regular 2D domains, training and evaluation on 3D domains or fully unstructured meshes are out of
scope of this work because of the non-trivial architectural changes involved. As an initial step, Section~\ref{subsec:eikonal_main} evaluates PRISMA on an Eikonal benchmark where the solution encodes irregular geometric boundaries, which is still sampled on a regular grid. Future work can further build on this to handle irregular boundaries, 3D domains, and unstructured meshes directly, either by mapping fields to latent regular grids through encoder--decoder architectures or by replacing the Fourier backbone with geometry-aware operators such as graph, mesh, or geometric
neural operators.
PRISMA is also evaluated on
time-dependent systems through one-step prediction with fixed time intervals, Appendix~\ref{appendix:spatio-temporal} provides a 2D+time experiment illustrating an extension to multiple timesteps, while scaling to
long-horizon temporal generation remains an important future direction.
Future work can further explore unified residual-aware diffusion models across multiple PDE families and specialized inverse problems such as full-waveform inversion, biomedical reconstruction, and data assimilation from sparse sensors.

\section*{Impact Statement}
\label{app:impact-statement}
This work develops fast and robust diffusion-based PDE solvers for scientific problems with noisy or sparse observations. PRISMA's main impact is to accelerate scientific research by enabling more efficient physics-guided inference in applications such as fluid dynamics, subsurface modeling, and biomedical imaging. The models are specialized for scientific PDE tasks and trained on simulated, non-sensitive data, so they carry a low risk of direct societal misuse.

\bibliography{main}

@article{pinnraissi2019physics,
  title={Physics-informed neural networks: A deep learning framework for solving forward and inverse problems involving nonlinear partial differential equations},
  author={Raissi, Maziar and Perdikaris, Paris and Karniadakis, George E},
  journal={Journal of Computational physics},
  volume={378},
  pages={686--707},
  year={2019},
  publisher={Elsevier}
}

@article{fno,
  title={Fourier neural operator for parametric partial differential equations},
  author={Li, Zongyi and Kovachki, Nikola and Azizzadenesheli, Kamyar and Liu, Burigede and Bhattacharya, Kaushik and Stuart, Andrew and Anandkumar, Anima},
  journal={arXiv preprint arXiv:2010.08895},
  year={2020}
}

@article{lu2019deeponet,
  title={Deeponet: Learning nonlinear operators for identifying differential equations based on the universal approximation theorem of operators},
  author={Lu, Lu and Jin, Pengzhan and Karniadakis, George Em},
  journal={arXiv preprint arXiv:1910.03193},
  year={2019}
}

@article{huang2024diffusionpde,
  title={DiffusionPDE: Generative PDE-solving under partial observation},
  author={Huang, Jiahe and Yang, Guandao and Wang, Zichen and Park, Jeong Joon},
  journal={Advances in Neural Information Processing Systems},
  volume={37},
  pages={130291--130323},
  year={2024}
}

@article{li2025videopde,
  title={VideoPDE: Unified Generative PDE Solving via Video Inpainting Diffusion Models},
  author={Li, Edward and Wang, Zichen and Huang, Jiahe and Park, Jeong Joon},
  journal={arXiv preprint arXiv:2506.13754},
  year={2025}
}

@article{fundps,
  title={Guided Diffusion Sampling on Function Spaces with Applications to PDEs},
  author={Yao, Jiachen and Mammadov, Abbas and Berner, Julius and Kerrigan, Gavin and Ye, Jong Chul and Azizzadenesheli, Kamyar and Anandkumar, Anima},
  journal={arXiv preprint arXiv:2505.17004},
  year={2025}
}

@article{ddolim2023score,
  title={Score-based diffusion models in function space},
  author={Lim, Jae Hyun and Kovachki, Nikola B and Baptista, Ricardo and Beckham, Christopher and Azizzadenesheli, Kamyar and Kossaifi, Jean and Voleti, Vikram and Song, Jiaming and Kreis, Karsten and Kautz, Jan and others},
  journal={arXiv preprint arXiv:2302.07400},
  year={2023}
}

@article{shu2023physics,
  title={A physics-informed diffusion model for high-fidelity flow field reconstruction},
  author={Shu, Dule and Li, Zijie and Farimani, Amir Barati},
  journal={Journal of Computational Physics},
  volume={478},
  pages={111972},
  year={2023},
  publisher={Elsevier}
}

@article{chung2022diffusion,
  title={Diffusion posterior sampling for general noisy inverse problems},
  author={Chung, Hyungjin and Kim, Jeongsol and Mccann, Michael T and Klasky, Marc L and Ye, Jong Chul},
  journal={arXiv preprint arXiv:2209.14687},
  year={2022}
}

@article{unorahman2022u,
  title={U-no: U-shaped neural operators},
  author={Rahman, Md Ashiqur and Ross, Zachary E and Azizzadenesheli, Kamyar},
  journal={arXiv preprint arXiv:2204.11127},
  year={2022}
}

@article{shysheya2024conditional,
  title={On conditional diffusion models for PDE simulations},
  author={Shysheya, Aliaksandra and Diaconu, Cristiana and Bergamin, Federico and Perdikaris, Paris and Hern{\'a}ndez-Lobato, Jos{\'e} Miguel and Turner, Richard and Mathieu, Emile},
  journal={Advances in Neural Information Processing Systems},
  volume={37},
  pages={23246--23300},
  year={2024}
}

@article{jacobsen2025cocogen,
  title={Cocogen: Physically consistent and conditioned score-based generative models for forward and inverse problems},
  author={Jacobsen, Christian and Zhuang, Yilin and Duraisamy, Karthik},
  journal={SIAM Journal on Scientific Computing},
  volume={47},
  number={2},
  pages={C399--C425},
  year={2025},
  publisher={SIAM}
}

@article{karras2022elucidating,
  title={Elucidating the design space of diffusion-based generative models},
  author={Karras, Tero and Aittala, Miika and Aila, Timo and Laine, Samuli},
  journal={Advances in neural information processing systems},
  volume={35},
  pages={26565--26577},
  year={2022}
}

@article{pino,
  title={Physics-informed neural operator for learning partial differential equations},
  author={Li, Zongyi and Zheng, Hongkai and Kovachki, Nikola and Jin, David and Chen, Haoxuan and Liu, Burigede and Azizzadenesheli, Kamyar and Anandkumar, Anima},
  journal={ACM/IMS Journal of Data Science},
  volume={1},
  number={3},
  pages={1--27},
  year={2024},
  publisher={ACM New York, NY}
}

@article{functionspacekerrigan2022diffusion,
  title={Diffusion generative models in infinite dimensions},
  author={Kerrigan, Gavin and Ley, Justin and Smyth, Padhraic},
  journal={arXiv preprint arXiv:2212.00886},
  year={2022}
}

@article{pidstrigach2023infinite,
  title={Infinite-dimensional diffusion models},
  author={Pidstrigach, Jakiw and Marzouk, Youssef and Reich, Sebastian and Wang, Sven},
  journal={arXiv preprint arXiv:2302.10130},
  year={2023}
}

@misc{Christopher2024neuraloperator,
	title = {neuraloperator/cond-diffusion-operators-edm},
	url = {https://github.com/neuraloperator/cond-diffusion-operators-edm},
	author = {Christopher Beckham},
	date = {2024-04-05},
	year = {2024},
	month = {4},
	day = {5},
}

@article{bastek2024physicsdiff,
  title={Physics-informed diffusion models},
  author={Bastek, Jan-Hendrik and Sun, WaiChing and Kochmann, Dennis M},
  journal={arXiv preprint arXiv:2403.14404},
  year={2024}
}

@article{piddmzhang2025physics,
  title={Physics-Informed Distillation of Diffusion Models for PDE-Constrained Generation},
  author={Zhang, Yi and Zou, Difan},
  journal={arXiv preprint arXiv:2505.22391},
  year={2025}
}

@article{cheng2024gradient,
  title={Gradient-free generation for hard-constrained systems},
  author={Cheng, Chaoran and Han, Boran and Maddix, Danielle C and Ansari, Abdul Fatir and Stuart, Andrew and Mahoney, Michael W and Wang, Yuyang},
  journal={arXiv preprint arXiv:2412.01786},
  year={2024}
}

@article{utkarsh2025physics,
  title={Physics-Constrained Flow Matching: Sampling Generative Models with Hard Constraints},
  author={Utkarsh, Utkarsh and Cai, Pengfei and Edelman, Alan and Gomez-Bombarelli, Rafael and Rackauckas, Christopher Vincent},
  journal={arXiv preprint arXiv:2506.04171},
  year={2025}
}

@article{zhongkai2024pinnacle,
  title={Pinnacle: A comprehensive benchmark of physics-informed neural networks for solving pdes},
  author={Zhongkai, Hao and Yao, Jiachen and Su, Chang and Su, Hang and Wang, Ziao and Lu, Fanzhi and Xia, Zeyu and Zhang, Yichi and Liu, Songming and Lu, Lu and others},
  journal={Advances in Neural Information Processing Systems},
  volume={37},
  pages={76721--76774},
  year={2024}
}

@article{liu2024preconditioning,
  title={Preconditioning for physics-informed neural networks},
  author={Liu, Songming and Su, Chang and Yao, Jiachen and Hao, Zhongkai and Su, Hang and Wu, Youjia and Zhu, Jun},
  journal={arXiv preprint arXiv:2402.00531},
  year={2024}
}

@article{kovachki2023neural,
  title={Neural operator: Learning maps between function spaces with applications to pdes},
  author={Kovachki, Nikola and Li, Zongyi and Liu, Burigede and Azizzadenesheli, Kamyar and Bhattacharya, Kaushik and Stuart, Andrew and Anandkumar, Anima},
  journal={Journal of Machine Learning Research},
  volume={24},
  number={89},
  pages={1--97},
  year={2023}
}

@article{jiang2024ode,
  title={ODE-DPS: ODE-based Diffusion Posterior Sampling for Inverse Problems in Partial Differential Equation},
  author={Jiang, Enze and Peng, Jishen and Ma, Zheng and Yan, Xiong-Bin},
  journal={arXiv preprint arXiv:2404.13496},
  year={2024}
}

@article{kohl2026benchmarking,
  title={Benchmarking autoregressive conditional diffusion models for turbulent flow simulation},
  author={Kohl, Georg and Chen, Li-Wei and Thuerey, Nils},
  journal={Neural Networks},
  pages={108641},
  year={2026},
  publisher={Elsevier}
}

@article{dasgupta2025conditional,
  title={Conditional score-based diffusion models for solving inverse elasticity problems},
  author={Dasgupta, Agnimitra and Ramaswamy, Harisankar and Murgoitio-Esandi, Javier and Foo, Ken Y and Li, Runze and Zhou, Qifa and Kennedy, Brendan F and Oberai, Assad A},
  journal={Computer Methods in Applied Mechanics and Engineering},
  volume={433},
  pages={117425},
  year={2025},
  publisher={Elsevier}
}

@article{li2025efficient,
  title={Efficient diffusion posterior sampling for noisy inverse problems},
  author={Li, Ji and Wang, Chao},
  journal={SIAM Journal on Imaging Sciences},
  volume={18},
  number={2},
  pages={1468--1492},
  year={2025},
  publisher={SIAM}
}

@article{chung2023decomposed,
  title={Decomposed diffusion sampler for accelerating large-scale inverse problems},
  author={Chung, Hyungjin and Lee, Suhyeon and Ye, Jong Chul},
  journal={arXiv preprint arXiv:2303.05754},
  year={2023}
}

@inproceedings{dou2026constrained,
  title={Constrained Particle Seeking: Solving Diffusion Inverse Problems with Just Forward Passes},
  author={Dou, Hongkun and Chen, Zike and Li, Zeyu and Li, Hongjue and Yang, Lijun and Deng, Yue},
  booktitle={Proceedings of the AAAI Conference on Artificial Intelligence},
  volume={40},
  number={25},
  pages={20870--20878},
  year={2026}
}

@inproceedings{daw2023mitigating,
  title={Mitigating Propagation Failures in Physics-informed Neural Networks using Retain-Resample-Release (R3) Sampling},
  author={Daw, Arka and Bu, Jie and Wang, Sifan and Perdikaris, Paris and Karpatne, Anuj},
  booktitle={International Conference on Machine Learning},
  pages={7264--7302},
  year={2023},
  organization={PMLR}
}

@inproceedings{
ramapuram2025theory,
title={Theory, Analysis, and Best Practices for Sigmoid Self-Attention},
author={Jason Ramapuram and Federico Danieli and Eeshan Gunesh Dhekane and Floris Weers and Dan Busbridge and Pierre Ablin and Tatiana Likhomanenko and Jagrit Digani and Zijin Gu and Amitis Shidani and Russell Webb},
booktitle={The Thirteenth International Conference on Learning Representations},
year={2025},
url={https://openreview.net/forum?id=Zhdhg6n2OG}
}

@inproceedings{kwon2021diagonal,
  title={Diagonal attention and style-based gan for content-style disentanglement in image generation and translation},
  author={Kwon, Gihyun and Ye, Jong Chul},
  booktitle={Proceedings of the IEEE/CVF International Conference on Computer Vision},
  pages={13980--13989},
  year={2021}
}

@inproceedings{long2025arbitrarily,
  title={Arbitrarily-Conditioned Multi-Functional Diffusion for Multi-Physics Emulation},
  author={Long, Da and Xu, Zhitong and Yang, Guang and Narayan, Akil and Zhe, Shandian},
  booktitle={International Conference on Machine Learning},
  pages={40270--40289},
  year={2025},
  organization={PMLR}
}
\bibliographystyle{tmlr}

\appendix

\appendix
\startcontents[app]  

\section*{Appendices}
\printcontents[app]{l}{1}{} 

\section{Theoretical Motivation}
\label{app:theory_sra}
We provide theoretical insights to motivate PRISMA as a residual-informed, gradient descent-free alternative to diffusion posterior sampling (DPS). Our work falls within a broader class of methods that replace iterative gradient-based inference with forward-pass updates learned during training \citep{li2025efficient,chung2023decomposed, dou2026constrained}. In this context, PRISMA bridges the gap between DPS-style methods and optimization-based correction by learning residual-informed updates directly in feature space.

Specifically, given a log-likelihood function informed by PDE residuals such as $
\log p(y \mid x) \propto -|R(x)|^2,
$
DPS updates feature maps using the following gradients of the log-likelihood:
$$
\nabla_x \log p(y \mid x) \propto - J_R(x)^T R(x),
$$
where $J_R(x)$ is the Jacobian of the residual operator. While principled, DPS is computationally expensive due to repeated backpropagation through the residual operator. It also relies on local first-order approximations that may break down as we move far away from the actual solution or with improper guidance hyper-parameter weights.

As an alternative to DPS, we can employ Newton-style updates to directly compute corrections in $x_t$ for minimizing $R(x_t)$. In particular, given an ideal solution $x^\star$ that is consistent with the PDE (i.e., $R(x^\star) = 0$), we can perform a first-order Taylor series expansion by linearizing $R(x)$ around $x^\star$ as follows:
$$
R(x_t) \approx J_R(x^\star)(x_t - x^\star).
$$
If we further assume the Jacobian to be locally invertible, this yields an idealized correction:
$$
x^\star \approx x_t - J_R(x^\star)^{-1} R(x_t).
$$
While this provides an estimate of the idealized correction at $x_t$ to arrive at $x^\star$,  it is only valid locally when $x_t$ is close to $x^\star$ (e.g., in the final steps of diffusion). Further, computing Jacobian-inverse is computationally prohibitive in high-dimensional spaces and fraught with ill-conditioning challenges.

In contrast to DPS and Newton-style methods, PRISMA learns an amortized mapping:
$$
\Delta x = \tilde{x}_{t-1} - \tilde{x}_{t} \approx f_\theta(x_t, R(x_t)),
$$
which approximates residual-informed corrections in $x$ via a forward pass in the spectral domain, without performing any backpropagation of gradients or computing Jacobian. In particular, the feature update equation of PRISMA is given by
$$
\tilde{x}_{t-1} = (1 - g) \cdot \tilde{x}_t + g \cdot a(w, x_t, R(x_t)) \cdot \tilde{x}_t,
$$
where $g$ and $w$ are learnable components. When the residual $R(x_t)$ is noisy and uninformative (in the initial steps of diffusion), we can set $g=0$ to obtain a stable solution of $\Delta x = 0$. As we get closer to the ideal solution $x^\star$ in later steps, $g$ can be set to 1 to rely more on residual guidance. This provides robustness and adaptability to PRISMA while preserving fast inference.

We emphasize that this connection to Newton's method is approximate: PRISMA's learned update is a single-pass surrogate for the idealized correction above rather than an explicit iterative solver, so it is not expected to keep refining indefinitely as more steps are added. Appendix \ref{app:convergence_tracking} supports this empirically, the PDE-residual norm and prediction error decline together and plateau at the same point in the sampling  trajectory.

\paragraph{SRA as attention-inspired spectral modulation.}
\label{app:sra_attention_clarification}

SRA is inspired by cross-attention, but it is not a full token-mixing attention layer. Standard
cross-attention computes dense interactions between query and key tokens and mixes value tokens
through a softmax-normalized attention matrix. In contrast, SRA operates in the Fourier domain and
uses a diagonal, frequency-wise attention mask. At layer $l$ and frequency mode $k$, we interpret
the feature spectrum as query and value, and the residual spectrum as key:
\[
Q^l(k)=\tilde{x}^l(k), \qquad
K^l(k)=\tilde{r}^l(k), \qquad
V^l(k)=\tilde{x}^l(k).
\]
where $\tilde{x}^{l}$ and $\tilde{r}^{l}$ are the Fourier transforms of feature maps and residuals at frequency mode $k$. SRA computes a query--key attention score using the complex inner product between feature and
residual spectra,
\[
S^l(k)=\frac{1}{\sqrt{C}}\left|\sum_{c=1}^{C} Q_c^l(k)\overline{K_c^l(k)}\right|,
\]
and converts it into a frequency-wise mask,
\[
A^l(k)=\sigma(w^l_{\mathrm{gain}}(k)S^l(k)).
\]
where \(w^l_{gain}>0\) is a learnable scaling factor. We use sigmoid instead of the standard softmax function to allow every frequency mode to be activated rather than performing a selection operation over the frequencies, borrowing on recent literature on sigmoid self-attention \citep{ramapuram2025theory}. 

We then perform an element-wise multiplication of the attention mask with the value map $V^l(k)$ to obtain frequency-modulated features.
\[
\tilde{x}^l_{\mathrm{mod}}(k)=A^l(k)V^l(k),
\]


This is a major difference from standard cross-attention that performs dense matrix multiplication between attention and value pairs resulting in full token-mixing. Instead, we consider a computationally cheaper alternative that is similar to prior diagonal-style formulations of attention \cite{kwon2021diagonal}. The frequency-modulated features are then skip-connected with the original features using a learnable gate $g_{\text{res}}$ as follows 
$$\tilde{x}^l_{SRA}(k) = (1 - g_{\text{res}}^l) \tilde{x}^l(k) + g_{\text{res}}^l \tilde{x}^{l}_{modulated}(k).$$

Thus, SRA is an attention-inspired residual-conditioned spectral modulation
mechanism: it preserves the query--key--value compatibility structure of attention, but replaces
dense token mixing with efficient diagonal frequency-wise modulation.

\section{Inference sampling algorithm}
\label{appendix:inference_algo}

Algorithm~\ref{alg:prisma_inference_final} details the inference sampling procedure for {PRISMA}, which employs a 2nd-order solver to iteratively generate a physically consistent solution from a random field. A key innovation of this process is the use of an {observation-guided PDE residual}, which is re-computed and fed into the denoising model at every step of the sampling process.

The core of the algorithm lies in the guided self-correction step(Lines 4 and 9). At each iteration $i$, the PDE residual $r$ is not computed on the model's raw prediction alone. Instead, it is calculated on a composite field $(1 - M) \odot x_i + M \odot x_{\text{obs}}$, where $M$ is a binary mask. This formulation ensures that for the known parts of the domain (where $M=1$), the residual calculation is grounded by the true observations $x_{\text{obs}}$. For the unknown parts (where $M=0$), it uses the model's current estimate $x_i$.

For instance, consider a forward problem with full observation, where the state $x = [\mathbf{a}, \mathbf{u}]$ consists of the known input coefficients $\mathbf{a}$ and the unknown solution $\mathbf{u}$. In this case, the mask $M = [M_\mathbf{a}, M_\mathbf{u}]$ would have $M_\mathbf{a}$ as a matrix of ones (fully observed) and $M_\mathbf{u}$ as a matrix of zeros (fully unobserved). Consequently, the composite field used to calculate the residual becomes $[\mathbf{a}_{\text{obs}}, \mathbf{u}_i]$. This means the PDE residual operator $\mathcal{R}$ evaluates physical consistency based on the {true input $\mathbf{a}$} and the {model's current prediction for the solution ${u}_i$}. 

This {guided residual $r$} acts as an explicit, spatially-varying map of physical inconsistency, which is then passed as a direct input to the denoising operator $D_\theta$ (Lines 5 and 10). By providing this physical guidance at every step of the predictor-corrector solver, the model is continuously steered toward solutions that are not only consistent with the initial observations but also compliant with the governing PDE, enabling fast, gradient-descent free convergence.

\begin{algorithm}
\caption{\ourmethod Inference with 2nd-Order Solver and Guided Residual}
\label{alg:prisma_inference_final}
\begin{algorithmic}[1]
\REQUIRE Observations $\mathbf{x}_{\text{obs}}=[\mathbf{a}_{\text{obs}}, \mathbf{u}_{\text{obs}}]$, masks $\mathbf{M}=[\mathbf{M}_a, \mathbf{M}_u]$, PDE residual operator $\mathcal{R}$, denoising diffusion operator $D_\theta$, variance schedule $\{\sigma_i\}_{i=0}^N$ with $\sigma_0=0$.
\STATE $\mathbf{x}_N \sim \mathcal{N}(0, \mathbf{C})$ \COMMENT{Initialize from GRF}
\FOR{$i=N, \dots, 1$}
    \STATE Let current state be $\mathbf{x}_i = [\mathbf{a}_i, \mathbf{u}_i]$.
    \COMMENT{\textit{Predictor Step}}
    \STATE Compute guided residual $\mathbf{r} \leftarrow \mathcal{R}((1-\mathbf{M}) \odot \mathbf{x}_i + \mathbf{M} \odot \mathbf{x}_{\text{obs}})$. \COMMENT{Guided self-correction}
    \STATE Predict clean state $\hat{\mathbf{x}}_0 \leftarrow D_\theta(\mathbf{x}_i, \sigma_i, \mathbf{x}_{\text{obs}}, \mathbf{M}, \mathbf{r})$.
    \STATE $\mathbf{d}_i \leftarrow (\mathbf{x}_i - \hat{\mathbf{x}}_0) / \sigma_i$. \COMMENT{Evaluate derivative $d\mathbf{x}/d\sigma$}
    \STATE $\mathbf{x}'_{i-1} \leftarrow \mathbf{x}_i + (\sigma_{i-1} - \sigma_i) \mathbf{d}_i$. \COMMENT{Take an Euler step}
    
    \IF[\textit{Corrector Step}]{$\sigma_{i-1} \neq 0$}
        \STATE Compute new guided residual $\mathbf{r}' \leftarrow \mathcal{R}((1-\mathbf{M}) \odot \mathbf{x}_{i-1}' + \mathbf{M} \odot \mathbf{x}_{\text{obs}})$  \COMMENT{Guided self-correction}
        \STATE Predict clean state $\hat{\mathbf{x}}'_0 \leftarrow D_\theta(\mathbf{x}'_{i-1}, \sigma_{i-1}, \mathbf{x}_{\text{obs}}, \mathbf{M}, \mathbf{r}')$.
        \STATE $\mathbf{d}'_i \leftarrow (\mathbf{x}'_{i-1} - \hat{\mathbf{x}}'_0) / \sigma_{i-1}$.
        \STATE $\mathbf{x}_{i-1} \leftarrow \mathbf{x}_i + (\sigma_{i-1} - \sigma_i) \cdot (\frac{1}{2} \mathbf{d}_i + \frac{1}{2} \mathbf{d}'_i)$. \COMMENT{Apply 2nd-order correction}
    \ELSE
        \STATE $\mathbf{x}_{i-1} \leftarrow \mathbf{x}'_{i-1}$.
    \ENDIF
\ENDFOR
\ENSURE Output sample $\mathbf{x}_0$
\end{algorithmic}
\end{algorithm}

\section{Implementation Details}
\label{appendix:implementation_details}
We adopt a 4-level U-shaped neural operator architecture \citep{unorahman2022u} as the denoiser $D_\theta$, which has 64M parameters, similar to DiffusionPDE’s and FunDPS's network size. The network is trained using 50,000 training samples for 200 epochs on 2 NVIDIA A100 GPUs with a Batch size of 90 per GPU. The code-base is built upon \citet{Christopher2024neuraloperator}'s implementation of Denoising Diffusion Operators (DDO). The hyperparameters we used for training and inference are listed in Table \ref{tab:parameter} and were taken from \citet{Christopher2024neuraloperator}'s  DDO implementation. We source the quantitative results of deterministic baselines for the full and noisy cases from DiffusionPDE’s \citep{huang2024diffusionpde} table. We faced reproducibility issues for DiffusionPDE, also observed by \citet{fundps}. We had correspondence with DiffusionPDE’s authors about this and have rerun all their experiments for the Full and Sparse cases, and the tables \ref{tab:full_main} and  \ref{tab:sparse_main} show our reproduced results for their method.

For Darcy, Poisson, and Helmholtz, we follow the same problem setup as prior work and use the
released DiffusionPDE and FunDPS checkpoints with their recommended guidance settings. For
Navier--Stokes, the released DiffusionPDE/FunDPS setup predicts the terminal state from the initial
state; however, this does not provide a well-defined time-local residual. We therefore reformulate Navier--Stokes as a next-step prediction task and train all
methods under this setting, which also enables iterative rollout evaluation. We use the same next-step
formulation for Kolmogorov flow.

\begin{table}[ht]
\centering
\caption{Hyperparameters}
\label{tab:parameter}
\begin{tabular}{lr}
\hline
\rule{0pt}{2ex}\textbf{Hyperparameter} & \textbf{Value}      \\ \hline
\rule{0pt}{2ex}learning\_rate           & 0.0001              \\
learning\_rate\_warmup   & 50 epochs   \\
ema\_half\_life          & 5 epochs   \\
dropout                  & 0.13                \\
rbf\_scale               & 0.05                \\
sigma\_max               & 80                  \\
sigma\_min               & 0.002               \\
rho                      & 7                   \\ \hline
\end{tabular}
\end{table}

{
\textbf{Residual-aware Guidance Strength MLP:} Each SRA block uses a small two-layer MLP to predict the scalar guidance weight 
$g_{\text{res}} \in [0,1]$. 
The inputs to this MLP are:  
(i) the diffusion/timestep embedding $c_{\sigma}$ (of length $E$, e.g., $E{=}256$), and  
(ii) the spatial mean of the residual $r_{\text{avg}}$ (a scalar).  
We concatenate these to form a tensor of shape $B \times (E{+}1)$, where $B$ is the batch size, and pass it through
\[
\text{Linear}(E{+}1 \rightarrow E)
\;\rightarrow\; \text{ReLU}
\;\rightarrow\; \text{Linear}(E \rightarrow 1)
\;\rightarrow\; \text{Sigmoid},
\]
yielding a $B \times 1$ output used as the skip-connected gating weight within the SRA block.  
Each SRA block contains its own distinct MLP.
}

\textbf{Task and mask sampling:} PRISMA training is agnostic to the sparsity levels used at inference. Rather than training on a single missingness pattern, we sample tasks and masks on-the-fly to expose the model to a broad distribution of observation configurations. We sample the task type with fixed probabilities:
$p(t)=0.1$ for \texttt{uncond}, $0.25$ each for \texttt{full-forward} and \texttt{full-inverse}, and $0.2$ each for \texttt{sparse-forward} and \texttt{sparse-inverse}. For full-observation tasks, we set $\bM_a=\mathbf{1}$ and/or $\bM_u=\mathbf{1}$, for forward and inverse respectively. For sparse tasks, we generate binary masks by first sampling an observation rate $p_{\text{obs}}\in[0.01,0.5]$ using a mixture distribution that emphasizes highly sparse settings:
\[
p_{\text{obs}} \sim
\begin{cases}
\text{Uniform}\!\big[\text{min},\,\text{min} + 0.1(\text{max}-\text{min})\big], & \text{w.p. } 0.5,\\[4pt]
\text{min} + \left(1 - U^\alpha\right)(\text{max}-\text{min}),\;\; U\sim\text{Uniform}(0,1), & \text{w.p. } 0.5,
\end{cases}
\]
with $\alpha{=}3$, $\text{min}{=}0.01$, and $\text{max}{=}0.5$. Given $p_{\text{obs}}$, we sample each grid location i.i.d. from $\text{Bernoulli}(p_{\text{obs}})$ to form the mask. This mixture sampler covers diverse sparsity levels while emphasizing highly sparse masks to train for challenging partial-observation regimes. A single model is trained with this procedure (for 50\% to 99\% sparse settings) and evaluated at test time under full observations, fixed sparsity levels (e.g., 97\% missing), and noisy observations.

\textbf{Diffusion noise vs. measurement noise:} During training, we add noise only to the \emph{target fields} $(\ba,\bu)$ to obtain $\bx_\sigma$, and train the model to denoise. Importantly, the conditioning observations $\bx_{\text{obs}}$ are always \emph{clean} during training (i.e., the model is never trained on noisy observations).

\textbf{GRF Covariance Kernel:} In our implementation, we build a GRF covariance prior (as in EDM-FS \cite{Christopher2024neuraloperator} and also used by FunDPS) and sample noise with that covariance. 

\subsection{Compute Cost Comparison}
\label{app:compute_cost}
 
\paragraph{Training speed:} We compare per-iteration training time on a single
NVIDIA A100 GPU, for Darcy Flow at 128$\times$128 resolution, averaged across
batches.
 
\begin{table}[h]
\centering
\small
\caption{Training and inference speed, relative to PRISMA.}
\label{tab:train_speed}
\begin{tabular}{lrrr}
\toprule
Method & Train (s/iteration) & 20k-iter.\ wall-clock (h) & Inference (s/sample) \\
\midrule
PRISMA       & 26.0  & 144.4 & 0.18 \\
FunDPS       & 20.2  & 112.2 & 11.8 \\
DiffusionPDE & 58.73 & 326.3 & 213  \\
\bottomrule
\end{tabular}
\end{table}
 
PRISMA's residual computation and SRA block add about 5.8s/iteration over FunDPS's plain diffusion-operator backbone ($\sim$28\% slower to train), but PRISMA still trains roughly 2.3$\times$ faster than DiffusionPDE per iteration.

\paragraph{Inference speed:} All inference-time experiments are conducted on a single NVIDIA A100 GPU. To determine per-sample inference time, we average batch inference time over 10 runs and divide by the batch size. Under the standard sampling configurations, \ourmethod\ processes 1,000 samples in about 3 minutes at 20 steps, while FunDPS takes about 3.3 hours at 500 steps and DiffusionPDE takes about 59.2 hours at 2000 steps. These estimates come from the per-sample inference times in Table~\ref{tab:noisy_main}: $0.18$s/sample for \ourmethod, $11.8$s/sample for FunDPS, and $213$s/sample for DiffusionPDE.

\paragraph{PDE-residual computation vs.\ network forward pass:} We instrument the sampler with CUDA event timers around the PDE-residual computation and each network forward call, at every sampling step, aggregated over a batch of 20 samples at 50 steps.
 
\begin{table}[h]
\centering
\caption{Wall-clock breakdown of PDE-residual computation vs.\ network forward pass
during sampling (50 steps, batch of 20 samples). The residual computation is
negligible relative to the network's forward pass across every equation and
direction, consistently under 1.2\% of total per-step compute.}
\label{tab:timing_breakdown}
\begin{tabular}{llrrrr}
\toprule
Equation & Direction & Residual (ms) & Network (ms) & Residual \% & GPU Mem (MB) \\
\midrule
Helmholtz      & Forward & 22.2  & 24{,}504.9 & 0.09\% & 3282.3 \\
Helmholtz      & Inverse & 21.8  & 29{,}314.1 & 0.07\% & 3282.3 \\
Kolmogorov     & Forward & 120.1 & 26{,}396.2 & 0.45\% & 3280.5 \\
Kolmogorov     & Inverse & 129.8 & 25{,}609.6 & 0.50\% & 3280.5 \\
Navier--Stokes & Forward & 288.8 & 24{,}702.1 & 1.16\% & 3282.3 \\
Navier--Stokes & Inverse & 155.6 & 24{,}354.7 & 0.63\% & 3282.3 \\
\bottomrule
\end{tabular}
\end{table}
 
\begin{figure}[h]
\centering
\includegraphics[width=0.85\linewidth]{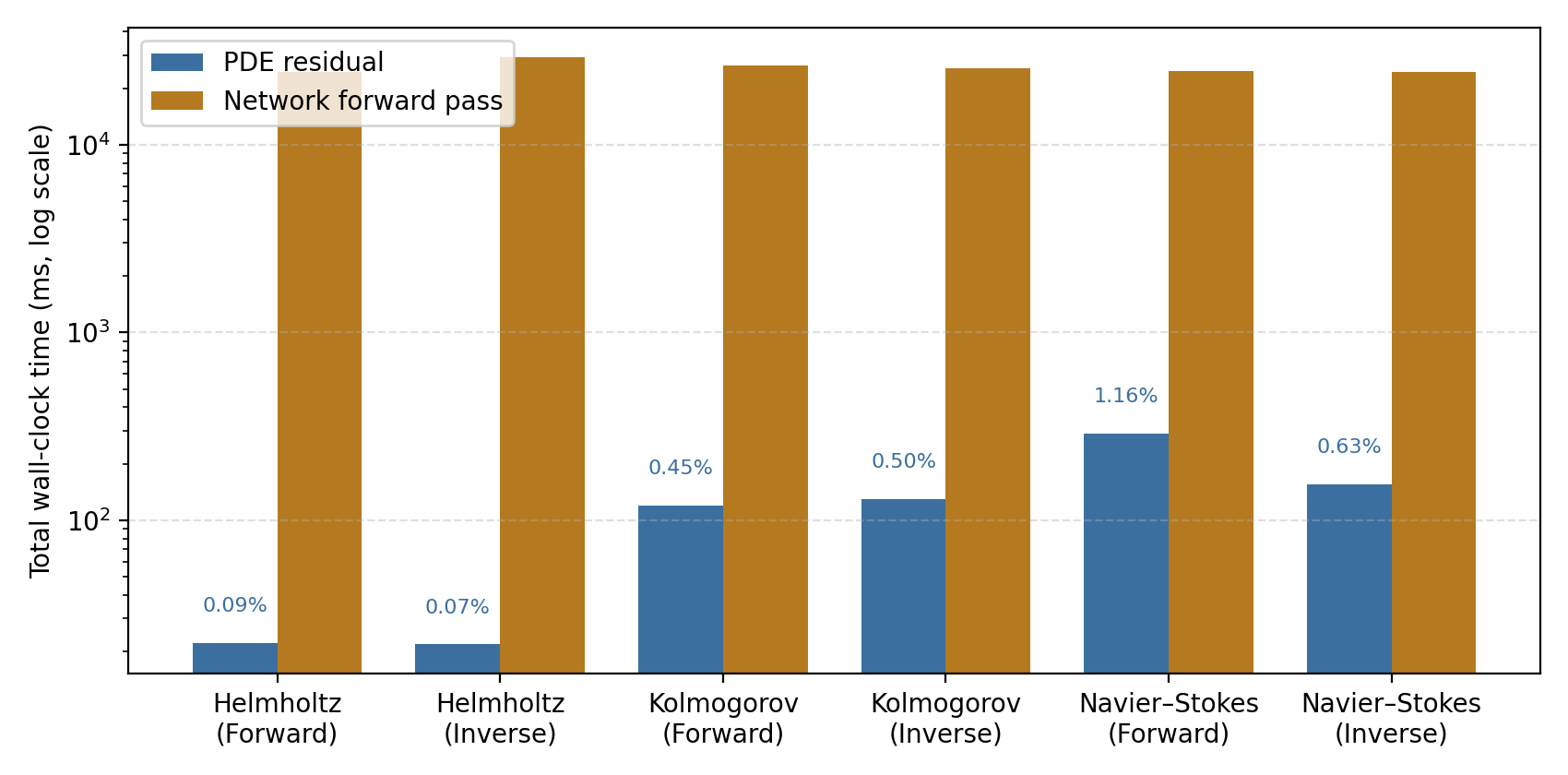}
\caption{Total wall-clock time (log scale) spent in PDE-residual computation vs.\
network forward pass, per equation and direction. The residual computation is
consistently negligible ($<1.2\%$) relative to the network's forward pass.}
\label{fig:timing_breakdown}
\end{figure}
 
The residual computation accounts for at most 1.16\% of per-step wall-clock time (Navier--Stokes, forward) and as little as 0.07\% (Helmholtz, inverse); the network forward pass dominates every case, and peak GPU memory ($\approx$3.28\,GB) is unaffected. The spread across equations follows from how each residual is computed: a finite-difference stencil for Helmholtz vs.\ an FFT-based spectral solve for Kolmogorov \& Navier--Stokes.

\section{Datasets}
\label{appendix:detailed_dataset_desc}


We evaluate our approach on five benchmark PDE problems: including Darcy Flow, Poisson, Helmholtz, Navier Stokes and Kolmogorov.
Each PDE is considered in both forward ($a\,\rightarrow\,u$) and Inverse ($u\,\rightarrow\,a$) settings. We use the same data as used in DiffusionPDE \citep{huang2024diffusionpde} which consists of $50{,}000$ training and $1000$ test samples for each PDE, at both $64{\times}64$ and $128{\times}128$ resolution. All models are trained and evaluated at $128{\times}128$, and some ablations when stated are on the $64{\times}64$ resolution. For quantitative evaluation, we report the mean relative $L^2$ error between the predicted and true solutions, except for the inverse Darcy Flow problem, where we use the binary error rate.

\paragraph{Darcy Flow Equation}

Darcy flow describes the movement of fluid through a porous medium. It is governed by the equation: 
\begin{equation}
-\nabla\!\cdot\!\big(a(x)\nabla u(x)\big)=g(x),\quad x\in \Omega,
\end{equation}
where the domain is $\Omega = (0,1)^2$, with $g(x) = 1$ (constant forcing) and zero Dirichlet boundary conditions. The coefficient function is sampled as $a \sim h_{\#}\mathcal{N}\!\big(0,(-\Delta+9\mathbf{I})^{-2}\big)$, following \citep{huang2024diffusionpde, fundps}. The mapping $h$ is defined piecewise, taking the value $12$ for positive inputs and $3$ otherwise.



\paragraph{Poisson Equation.}
We consider the Poisson equation on the unit square $\Omega = (0,1)^2$ which describes steady-state diffusion processes:
\begin{equation}
\nabla^2 \mathbf{u}(x) = \mathbf{a}(x), \quad x \in \Omega,
\end{equation}
subject to homogeneous Dirichlet boundary conditions:
\begin{equation}
\mathbf{u}(x) = 0, \quad x \in \partial \Omega.
\end{equation}
The source term $\mathbf{a}(x)$ is a binary field, obtained by thresholding a Gaussian random field at zero:
\begin{equation}
\mathbf{a}(x) = \mathbf{1}_{{Z(x) > 0}}, \quad Z \sim \mathcal{N}\big(0, (-\Delta + 9\mathbf{I})^{-2}\big),
\end{equation}
where $\mathbf{1}_{{\cdot}}$ is the indicator function.
The PDE residual, which serves as a physical constraint for the model, is defined as,
\begin{equation}
f(x) = \nabla^2 \mathbf{u}(x) - \mathbf{a}(x).
\end{equation}

\paragraph{Helmholtz Equation.}
We consider the static inhomogeneous Helmholtz equation on the domain $\Omega = (0,1)^2$:
\begin{equation}
\Delta u(x) + k^2 u(x) = a(x), \quad x \in \Omega,
\end{equation}
subject to homogeneous Dirichlet boundary conditions:
\begin{equation}
u(x) = 0, \quad x \in \partial \Omega,
\end{equation}
where $k = 1$. This equation describes wave propagation in heterogeneous media. The coefficient function $a(x)$ is generated as a piecewise-constant field sampled from a Gaussian random field. Unlike the Poisson case, where the source term is thresholded to a binary field, the Helmholtz coefficients can take a range of values across different regions, giving heterogeneous variation. The solution $u(x)$ is computed using second-order finite differences, with zero boundary enforced via the same mollifier as in \citep{huang2024diffusionpde}.

{
\paragraph{Navier--Stokes Equation (non-bounded).}
We consider the 2D incompressible Navier--Stokes equations in vorticity--streamfunction form:
\begin{align}
\partial_t \omega(x,t) + \mathbf{v}(x,t)\cdot\nabla \omega(x,t) &= \nu\,\Delta \omega(x,t) + f(x), \\
\Delta \psi(x,t) &= -\,\omega(x,t), \\
\mathbf{v}(x,t) &= \nabla^\perp \psi(x,t), \qquad
\nabla\!\cdot\!\mathbf{v}(x,t) = 0,
\end{align}
with periodic boundary conditions. Here $\omega=\partial_x v_y-\partial_y v_x$ is the vorticity, $\psi$ the streamfunction, $\mathbf{v}$ the velocity, and $\nu=10^{-3}$ ($\mathrm{Re}=1000$).
Initial vorticity $\omega_0$ is sampled from a Gaussian random field.}

{
For our experiments, we use the dataset provided by DiffusionPDE \citep{huang2024diffusionpde}, having pairs $(\omega_0(x),\omega_T(x))$ corresponding to the solution at final time $T=1$s. We \emph{deviate} from their final-frame prediction setup and instead predict the \emph{next} time step with a small interval $\Delta t=0.1$ to enable a physically meaningful residual. DiffusionPDE exploits the identity $\nabla\!\cdot(\nabla\times \mathbf{v})=0$, using a simplified residual written as:
\[
f(x)=\nabla\!\cdot \omega(x,t).
\]
However, this is not physically valid in 2D because $\omega$ is a scalar field and $\nabla\!\cdot \omega$ is not a well-defined divergence (nor a meaningful physics residual).}
{
We thus use the following {revised formulation of PDE residual in our experiments.}
Given $\omega_t$ and a candidate $\tilde{\omega}_{t+\Delta t}$, we first recover the velocity at $t{+}\Delta t$ via a streamfunction solve given by:
\[
\Delta \tilde{\psi}_{t+\Delta t} \;=\; -\,\tilde{\omega}_{t+\Delta t}
\quad\Rightarrow\quad
\tilde{\mathbf{v}}_{t+\Delta t} \;=\; \nabla^\perp \tilde{\psi}_{t+\Delta t}
= \big(\partial_y \tilde{\psi}_{t+\Delta t},\, -\,\partial_x \tilde{\psi}_{t+\Delta t}\big).
\]
The one-step vorticity-transport residual is then computed as:
\begin{equation}
\mathcal{R}(x)
= \frac{\tilde{\omega}_{t+\Delta t}(x)-\omega_t(x)}{\Delta t}
\;+\; \tilde{\mathbf{v}}_{t+\Delta t}(x)\cdot \nabla \tilde{\omega}_{t+\Delta t}(x)
\;-\; \nu\,\Delta \tilde{\omega}_{t+\Delta t}(x)
\;-\; f(x),
\end{equation}}

{
\paragraph{Kolmogorov Flow.}
We consider the two-dimensional Kolmogorov flow governed by the incompressible Navier--Stokes equations in vorticity form:
\begin{align}
\partial_t \omega(x,t) + \mathbf{u}(x,t)\cdot\nabla \omega(x,t)
&= \frac{1}{\mathrm{Re}}\Delta \omega(x,t) + f(x,t), \\
\nabla\!\cdot\!\mathbf{u}(x,t) &= 0,
\end{align}
on the periodic domain $x\in(0,2\pi)^2$. Here $\omega$ denotes the vorticity, $\mathbf{u}$ is the velocity field, and the Reynolds number is set to $\mathrm{Re}=1000$. Following prior work~\citep{shu2023physics}, the forcing term is defined as
\begin{equation}
f(x,t) = -4\cos(4x_2) - 0.1\,\omega(x,t),
\end{equation}
where the second term acts as a drag force to prevent energy accumulation at large scales. For our experiments, we use the Kolmogorov flow dataset from \citet{shu2023physics}. The original simulations are generated using a pseudo-spectral solver with periodic boundary conditions, where the initial vorticity field $\omega_0$ is sampled from a Gaussian random field. The data are downsampled to a $256\times256$ spatial grid. In our setup, we formulate the task as next-step prediction, similar to our Navier--Stokes experiments, so that the PDE residual is physically meaningful. This gives a training set of approximately $12$K samples and a test set of $600$ samples.
}

{
Given the current vorticity $\omega_t$ and a candidate next-step prediction $\tilde{\omega}_{t+\Delta t}$, we compute the Kolmogorov residual using a one-step finite-difference approximation in time. We first recover the velocity from the predicted vorticity through the streamfunction formulation:
\[
\Delta \tilde{\psi}_{t+\Delta t} = -\,\tilde{\omega}_{t+\Delta t},
\qquad
\tilde{\mathbf{u}}_{t+\Delta t}
= \nabla^\perp \tilde{\psi}_{t+\Delta t}
= \big(\partial_y \tilde{\psi}_{t+\Delta t},\, -\partial_x \tilde{\psi}_{t+\Delta t}\big).
\]
The one-step Kolmogorov residual is then computed as
\begin{equation}
\mathcal{R}(x)
=
\frac{\tilde{\omega}_{t+\Delta t}(x)-\omega_t(x)}{\Delta t}
+
\tilde{\mathbf{u}}_{t+\Delta t}(x)\cdot\nabla \tilde{\omega}_{t+\Delta t}(x)
-
\frac{1}{\mathrm{Re}}\Delta \tilde{\omega}_{t+\Delta t}(x)
-
f(x,t+\Delta t),
\end{equation}
where
\[
f(x,t+\Delta t) = -4\cos(4x_2) - 0.1\,\tilde{\omega}_{t+\Delta t}(x).
\]
Spatial derivatives are computed spectrally using Fourier transforms, and the temporal derivative is approximated using the one-step finite difference.
}
\section{Results}
\label{appendix:results}
\subsection{Full Observation}

\begin{table*}[!t]
    \centering
    \caption{Comparing different models on PDE problems under full observation (relative $L_2$ error). For baselines with known reproducibility variability reported in prior work, we report results reproduced using the authors' released code and checkpoints (marked with *). For Navier--Stokes and Kolmogorov, we use next-step prediction rather than last-step prediction to enable a well-defined physical residual. \textbf{Best} is bolded and \underline{second-best} is underlined.}
    \setlength{\tabcolsep}{0.22em}
    \renewcommand{\arraystretch}{1.1}
    \resizebox{\textwidth}{!}{%
    \begin{tabular}{l c c c c c c c c c c c c c}
    \toprule
      & \multirow{2}{*}{\textbf{Steps} $(N)$} 
      & \multirow{2}{*}{\shortstack{\textbf{Inference}\\\textbf{Time}}}
      & \multicolumn{2}{c}{\textbf{Darcy Flow}}
      & \multicolumn{2}{c}{\textbf{Poisson}}
      & \multicolumn{2}{c}{\textbf{Helmholtz}}
      & \multicolumn{2}{c}{\textbf{Navier--Stokes}}
      & \multicolumn{2}{c}{\textbf{Kolmogorov Flow}}
      & \multirow{2}{*}{\shortstack{\textbf{Avg.}\\\textbf{Rank} $\downarrow$}} \\
    \cmidrule(lr){4-5}
    \cmidrule(lr){6-7}
    \cmidrule(lr){8-9}
    \cmidrule(lr){10-11}
    \cmidrule(lr){12-13}
      & & 
      & \textbf{Forward} & \textbf{Inverse} 
      & \textbf{Forward} & \textbf{Inverse} 
      & \textbf{Forward} & \textbf{Inverse} 
      & \textbf{Forward} & \textbf{Inverse}
      & \textbf{Forward} & \textbf{Inverse}
      & \\
    \midrule

    \textbf{PINO} & $-$ & $0.11$ &
    4.00\% & \textbf{2.10\%} 
    & 3.70\% & 10.20\% 
    & 4.90\% & \textbf{4.90\%} 
    & 4.53\% & 4.38\%
    & \textbf{3.68\%} & \textbf{3.92\%}
    & \underline{3.4} \\

    \textbf{DeepONet} & $-$ & $-$ &
    12.30\% & 8.40\% 
    & 14.30\% & 29.00\% 
    & 17.80\% & 28.10\% 
    & -- & --
    & -- & --
    & 7.8 \\

    \textbf{PINNs} & $-$ & $3.3$ &
    15.40\% & 10.10\% 
    & 16.10\% & 28.50\% 
    & 18.10\% & 29.20\% 
    & -- & --
    & -- & --
    & 8.6 \\

    \textbf{FNO} & $-$ & $0.1$ &
    5.30\% & 5.60\% 
    & 8.20\% & 13.60\% 
    & 11.10\% & \underline{5.00\%} 
    & 4.43\% & 4.52\%
    & \underline{4.37\%} & \underline{6.93\%}
    & 4.9 \\

    \midrule

    \textbf{DiffusionPDE*} & 2000 & 213 &
    2.90\% & 13.00\% 
    & 15.27\% & 21.21\% 
    & 10.90\% & 18.97\% 
    & \textbf{0.60\%} & 1.50\%
    & 22.44\% & 24.63\%
    & 5.5 \\

    \textbf{FunDPS*} & 200 & 4.72 &
    \underline{1.10\%} & 4.20\% 
    & \textbf{0.70\%} & 23.32\% 
    & \textbf{1.08\%} & 18.48\% 
    & 4.00\% & 1.68\%
    & 31.92\% & 33.51\%
    & 4.40 \\

    \textbf{FunDPS*} & 500 & 11.78 &
    1.40\% & \underline{3.00\%} 
    & \underline{0.84\%} & 19.84\% 
    & \textbf{1.08\%} & 13.88\% 
    & 3.77\% & 1.58\%
    & 32.63\% & 41.12\%
    & 4.1 \\

    \midrule

    \textbf{PRISMA (ours)} & 20 & \textbf{0.18} &
    \underline{1.10\%} & 3.80\% 
    & 4.52\% & \underline{8.31\%} 
    & 4.29\% & 8.03\% 
    & 0.97\% & \underline{1.30\%}
    & 12.15\% & 12.52\%
    & \underline{3.4} \\

    \textbf{PRISMA (ours)} & 50 & \underline{0.8} &
    \textbf{1.05\%} & 3.79\% 
    & 4.00\% & \textbf{8.00\%} 
    & \underline{3.81\%} & 7.83\% 
    & \underline{0.88\%} & \textbf{1.25\%}
    & {12.02\%} & {12.40\%}
    & \textbf{2.4} \\
    
    \bottomrule
    \end{tabular}
    }
    \label{tab:full_main}
\end{table*}
Table \ref{tab:full_main} compares different models across five PDE problems for the full observation setting. Our method achieves competitive performance, achieving the best average rank of 2.4 across all tasks. In addition to accuracy, our approach demonstrates significant efficiency: compared to other diffusion-based models, it is 15x to 250x (seconds per sample) faster during inference. 
Note that DiffusionPDE's original reported numbers were not reproducible; after correspondence with the authors (Appendix~\ref{appendix:implementation_details}), we reran their released code and checkpoints ourselves, and Tables~\ref{tab:full_main}~and~\ref{tab:sparse_main} report these reproduced results (marked with *).
\subsection{Sparse Observations}

Table~\ref{tab:sparse_main} presents a comparison of PRISMA against deterministic and generative baselines under the challenging {sparse observation} setting. While FunDPS (at 500 steps) achieves the highest accuracy, PRISMA demonstrates a superior balance of performance and efficiency. Our model delivers competitive accuracy, comparable to the top-performing generative methods, but at a fraction of the computational cost. Notably, PRISMA at 20 steps is over {65 times faster} than the most accurate FunDPS configuration and over {1000 times faster} than DiffusionPDE. This highlights PRISMA's significant advantage in inference speed, making it a practical and efficient choice for applications where rapid predictions are critical.

\begin{table*}[h]
    \centering
     \caption{Comparison of models on PDE problems under sparse observation (97\% pixels missing; relative $L_2$ error). For Navier--Stokes, we use next-step prediction rather than last-step prediction to enable a well-defined physical residual. \textbf{Bold} denotes best and \underline{underline} denotes second-best.}
   
    \fontsize{9pt}{9pt}\selectfont
    \setlength{\tabcolsep}{0.45em}
    \resizebox{\textwidth}{!}{%
    \begin{tabular}{l c c c c c c c c}
    \toprule
    & \multirow{2}{*}{\textbf{Steps} $(N)$} 
    & \multicolumn{2}{c}{\textbf{Darcy Flow}} 
    & \multicolumn{2}{c}{\textbf{Navier--Stokes}} 
    & \multicolumn{2}{c}{\textbf{Kolmogorov Flow}}
    & \multirow{2}{*}{\shortstack{\textbf{Avg.}\\\textbf{Rank} $\downarrow$}} \\
    \cmidrule(lr){3-4}
    \cmidrule(lr){5-6}
    \cmidrule(lr){7-8}
    & & \textbf{Forward} & \textbf{Inverse} 
      & \textbf{Forward} & \textbf{Inverse} 
      & \textbf{Forward} & \textbf{Inverse}
      & \\
    \midrule

    \textbf{FNO} & $-$ 
    & 28.20\% & 49.30\% 
    & 96.74\% & 149.48\%
    & 97.89\% & 98.85\%
    & 6.17 \\

    \textbf{PINO} & $-$ 
    & 35.20\% & 49.20\% 
    & 97.83\% & 144.38\%
    & 99.38\% & 101.23\%
    & 6.50 \\

    \textbf{DeepONet} & $-$ 
    & 38.30\% & 41.10\% 
    & -- & --
    & -- & --
    & 7.00 \\

    \textbf{PINN} & $-$ 
    & 48.80\% & 59.70\% 
    & -- & --
    & -- & --
    & 9.00 \\

    \midrule

    \textbf{DiffusionPDE} & 2000 
    & 6.07\% & 7.87\% 
    & 7.87\% & 8.00\%
    & 218.56\% & 223.56\%
    & 5.17 \\

    \textbf{FunDPS} & 200 
    & \underline{2.88\%} & 6.78\% 
    & 11.89\% & 9.16\%
    & 66.65\% & 56.57\%
    & 4.00 \\
    
    \textbf{FunDPS} & 500 
    & \textbf{2.49\%} & \textbf{5.18\%} 
    & 10.28\% & 7.91\%
    & 63.39\% & \underline{56.41\%}
    & \underline{2.33} \\

    \midrule

    \textbf{PRISMA (ours)} & 20 
    & 2.99\% & \underline{6.56\%} 
    & \underline{7.09\%} & \underline{7.90\%}
    & \underline{57.84\%} & 56.48\%
    & 2.58 \\

    \textbf{PRISMA (ours)} & 50 
    & 2.90\% & \underline{6.56\%} 
    & \textbf{6.99\%} & \textbf{7.85\%}
    & \textbf{57.51\%} & \textbf{56.37\%}
    & \textbf{1.58} \\

    \bottomrule
    \end{tabular}
    }
    \label{tab:sparse_main}
\end{table*}

\section{Physics Extrapolation for Helmholtz Wavenumber k }
\label{appendix:phys_helm_wave}

We further evaluate physics extrapolation on the Helmholtz equation by varying the wavenumber
$k$ at test time. All models are trained on the standard Helmholtz setting with $k{=}1$ and are
evaluated zero-shot on harder wave regimes with $k\in\{2,3,4\}$, without any retraining or
guidance re-tuning. Increasing $k$ changes the PDE operator and produces more oscillatory
solutions, making this a challenging test of whether the learned solver can extrapolate beyond the
training physics. Table~\ref{tab:helmholtz_wavenumber_full_sparse_noisy} reports results under
full, sparse, and noisy observation settings for both forward and inverse problems. \ourmethod\
achieves the best overall performance across most settings while using only 20 sampling steps,
showing that architectural residual guidance remains effective under shifts in the governing PDE
parameter.

\begin{table*}[!h]
    \centering
    \fontsize{9pt}{9pt}\selectfont
    \renewcommand{\arraystretch}{1.2} 
    \caption{Helmholtz physics extrapolation under varying wavenumbers ($k=2,3,4$) with Noisy,Sparse \& Full observations ($L_2$ relative error). \textbf{Best} is bolded, \underline{second-best} underlined.}
    \setlength{\tabcolsep}{0.5em}
    \resizebox{\textwidth}{!}{%
    \begin{tabular}{l c c c c c c c c}
    \toprule
      & \multirow{2}{*}{\textbf{Steps} $(N)$} & \multicolumn{2}{c}{\textbf{Noisy}}
      & \multicolumn{2}{c}{\textbf{Sparse}}
      & \multicolumn{2}{c}{\textbf{Full}}
      & \multirow{2}{*}{\shortstack{\textbf{Wavenumber}\\$\mathbf{k}$}} \\
    \cmidrule(lr){3-4}
    \cmidrule(lr){5-6}
    \cmidrule(lr){7-8}
      & & \textbf{Forward} & \textbf{Inverse} & \textbf{Forward} & \textbf{Inverse} & \textbf{Forward} & \textbf{Inverse} & \\
    \midrule

    \textbf{DiffusionPDE} & 2000 &
    \underline{39.01\%} & \underline{115.75\%} & 24.90\% & \textbf{19.42\%} & 14.73\% & \underline{15.86\%} & $2$ \\

    \textbf{FunDPS} & 200 &
    51.04\% & 624.20\% & \underline{9.43\%} & \underline{19.97\%} & \underline{9.28\%} & 18.99\% & $2$ \\

    \textbf{PRISMA (ours)} & 20 &
    \textbf{14.80\%} & \textbf{49.27\%} & \textbf{7.61\%} & 36.03\% & \textbf{7.42\%} & \textbf{11.95\%} & $2$ \\

    \midrule

    \textbf{DiffusionPDE} & 2000 &
    \underline{47.99\%} & \underline{117.41\%} & 36.23\% & \textbf{24.96\%} & 28.59\% & \underline{21.69\%} & $3$ \\

    \textbf{FunDPS} & 200 &
    52.67\% & 632.86\% & \underline{25.52\%} & \underline{27.65\%} & \underline{25.56\%} & 26.61\% & $3$ \\

    \textbf{PRISMA (ours)} & 20 &
    \textbf{19.16\%} & \textbf{51.66\%} & \textbf{14.71\%} & 37.62\% & \textbf{14.92\%} & \textbf{18.00\%} & $3$ \\

    \midrule

    \textbf{DiffusionPDE} & 2000 &
    70.33\% & \underline{114.70\%} & 64.86\% & \underline{62.35\%} & 60.32\% & \underline{55.52\%} & $4$ \\

    \textbf{FunDPS} & 200 &
    \underline{64.14\%} & 597.93\% & \underline{59.01\%} & 80.83\% & \underline{59.03\%} & 80.75\% & $4$ \\

    \textbf{PRISMA (ours)} & 20 &
    \textbf{55.22\%} & \textbf{63.20\%} & \textbf{52.05\%} & \textbf{52.34\%} & \textbf{52.04\%} & \textbf{41.13\%} & $4$ \\

    \bottomrule
    \end{tabular}%
    }
    \label{tab:helmholtz_wavenumber_full_sparse_noisy}
\end{table*}

\section{Spectral Power Analysis}
\label{app::spectral_power}

In addition to pointwise relative error, we analyze whether the predicted fields preserve the frequency content of the ground-truth solutions. This is particularly important in noisy inverse settings, where models may either over-smooth high-frequency structure or amplify noise-induced artifacts. We compute the relative error between the radially averaged power spectrum of the prediction and that of the ground truth. Table~\ref{tab:noisy_full_spectral_power_l2} reports this metric for Darcy, Poisson, and Helmholtz equations under noisy observations. Across most forward and inverse settings, \ourmethod achieves lower spectral power error than the compared diffusion-based baselines, indicating better preservation of the solution spectrum under corrupted measurements.

\begin{figure*}[!h]
    \centering
    \begin{subfigure}[t]{0.32\textwidth}
        \centering
        \includegraphics[width=\linewidth]{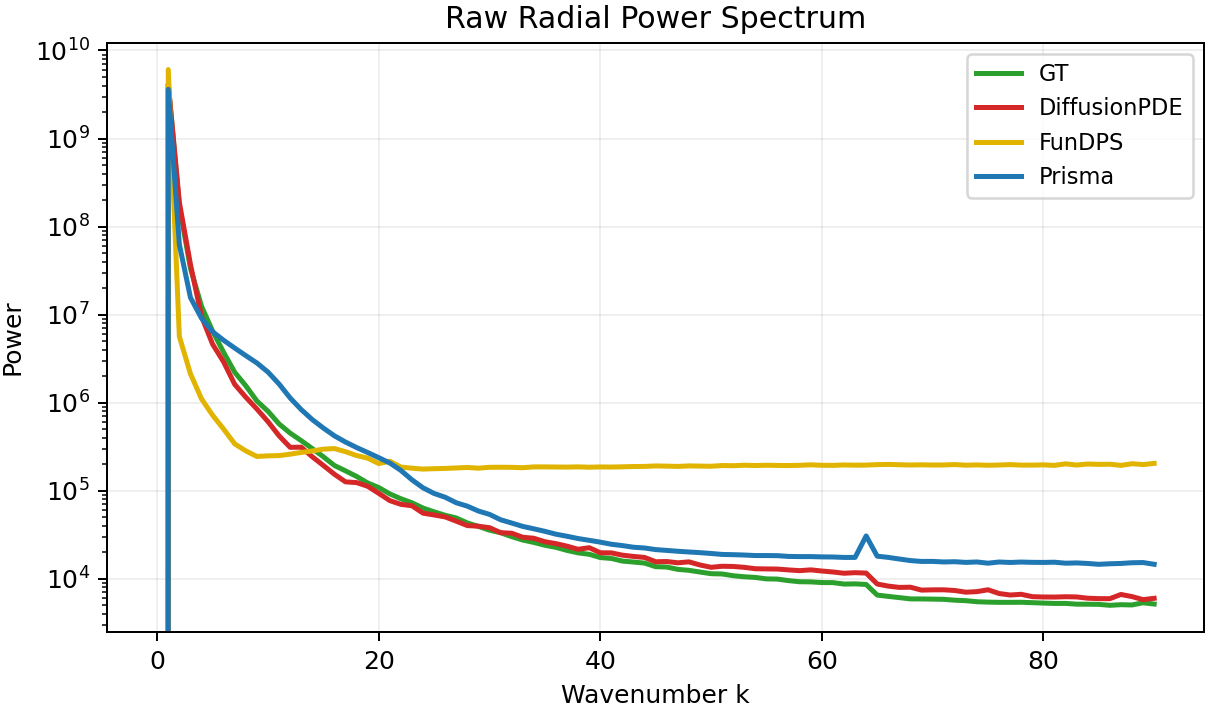}
        \caption{Darcy Inverse}
        \label{fig:rebuttal:darcy_inv}
    \end{subfigure}
    \begin{subfigure}[t]{0.32\textwidth}
        \centering
        \includegraphics[width=\linewidth]{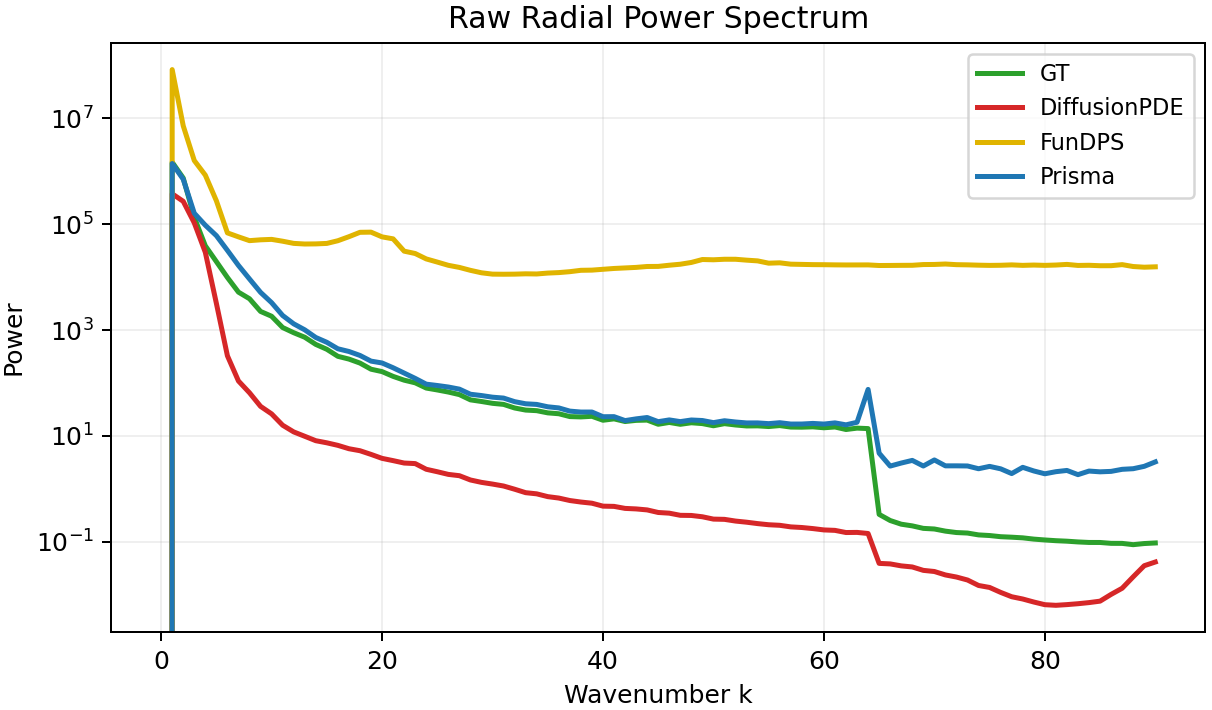}
        \caption{Helmholtz Inverse}
        \label{fig:rebuttal:helmholtz_inv}
    \end{subfigure}
    \begin{subfigure}[t]{0.32\textwidth}
        \centering
        \includegraphics[width=\linewidth]{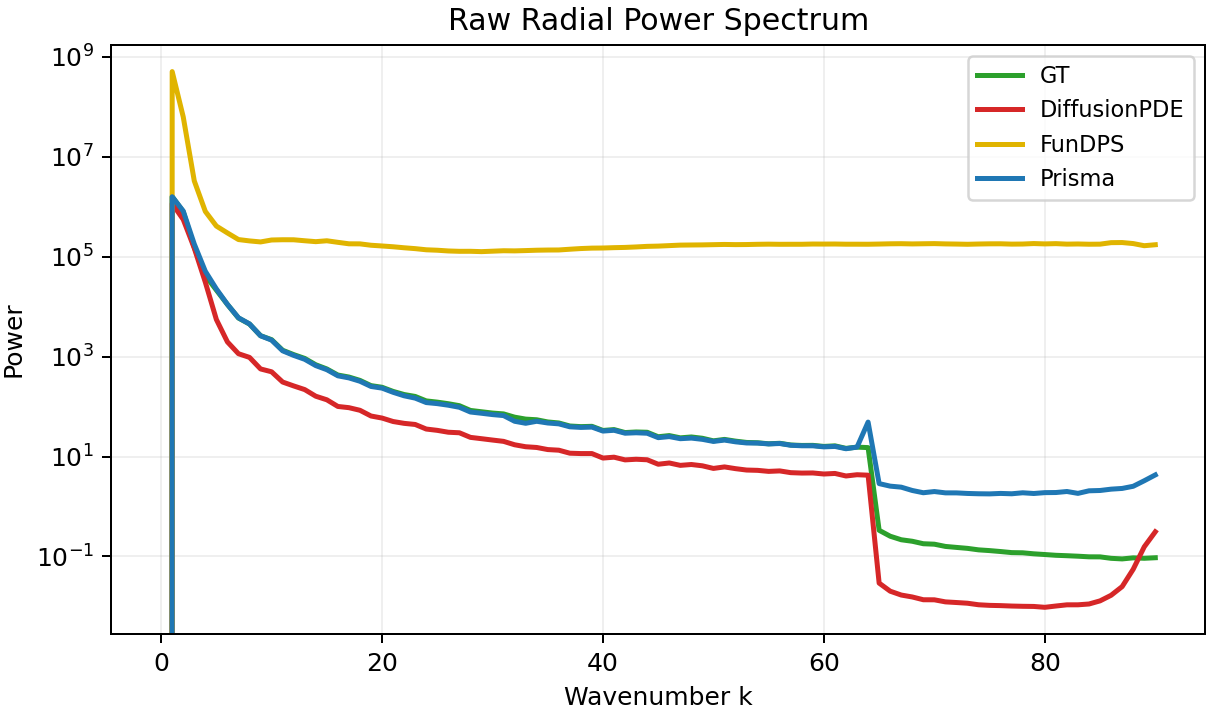}
        \caption{Poisson Inverse}
        \label{fig:rebuttal:poisson_inv}
    \end{subfigure}
    \caption{Radial power spectra under noisy inverse observations for (a) Darcy, (b) Helmholtz, and (c) Poisson. We plot the radially averaged power spectrum of the predicted fields and the ground-truth signal to analyze frequency fidelity.}
    \label{fig:rebuttal:three_plots}
\end{figure*}

Figure~\ref{fig:rebuttal:three_plots} further visualizes the radial power spectra for the noisy inverse setting. The spectra show that \ourmethod more closely follows the ground-truth power distribution across frequency modes, whereas the baselines deviate substantially, especially in higher-frequency regions. 

\begin{table}[h]
    \centering
    \caption{Relative spectral power error under noisy observations for Darcy, Poisson, and Helmholtz in both forward and inverse settings.}
    \setlength{\tabcolsep}{0.35em}
    \renewcommand{\arraystretch}{1.2}
    \resizebox{\textwidth}{!}{%
    \begin{tabular}{l c c c c c c c c}
    \toprule
      & \multirow{2}{*}{\textbf{Steps} $(N)$}
      & \multirow{2}{*}{\shortstack{\textbf{Inference}\\\textbf{Time (s)}}}
      & \multicolumn{2}{c}{\textbf{Darcy Flow}}
      & \multicolumn{2}{c}{\textbf{Poisson}}
      & \multicolumn{2}{c}{\textbf{Helmholtz}} \\
    \cmidrule(lr){4-5}
    \cmidrule(lr){6-7}
    \cmidrule(lr){8-9}
      & & & \textbf{Forward} & \textbf{Inverse} & \textbf{Forward} & \textbf{Inverse} & \textbf{Forward} & \textbf{Inverse} \\
    \midrule
    \textbf{DiffusionPDE} & 2000 & 213.0 &
    67.23\% & \underline{14.49\%} & \underline{40.52\%} & \underline{106.52\%} & \underline{45.33\%} & \underline{81.98\%} \\
    
    \textbf{FunDPS} & 200 & 4.72 &
    \underline{54.11\%} & 82.20\% & 47.56\% & 43799.14\% & 67.01\% & 6761.69\% \\
    \midrule
    
    \textbf{PRISMA (ours)} & 20 & 0.18 &
    \textbf{10.01\%} & \textbf{11.97\%} & \textbf{34.65\%} & \textbf{23.10\%} & \textbf{18.42\%} & \textbf{28.26\%} \\
    \bottomrule
    \end{tabular}%
    }
    \label{tab:noisy_full_spectral_power_l2}
\end{table}

\section{Evaluation on geometric PDEs: Eikonal signed distance fields}
\label{app:eikonal}

The signed distance function $\phi(x)$ gives the distance of each spatial location $x$ to the closest point on a shape's boundary, with the sign indicating whether $x$ lies inside or outside the shape. It satisfies the Eikonal constraint
\begin{equation}
    \|\nabla \phi(x)\| = 1,
\end{equation}
and the boundary is recovered as the zero level set $\phi(x)=0$. We use the
irregular-boundary benchmark of \citet{daw2023mitigating}.

\paragraph{Evaluation protocol.}
We evaluate only the forward full-observation setting for this benchmark. Sparse or noisy perturbations of the input geometry can change the underlying target SDF itself, making the target field ambiguous rather than simply partially observed. We report relative $L_2$ error on the predicted SDF field as the primary metric. We also visualize thresholded binary reconstructions obtained from the SDF and report mIoU as a secondary measure of boundary recovery.
\paragraph{Baseline details.}
For FunDPS and DiffusionPDE, we initialize guidance weights from the released Darcy-flow settings in their codebases, since Darcy is the closest released binary scalar-field setup. We then perform a limited local search over nearby hyperparameters for the Eikonal task, as inference is expensive for these DPS-style baselines.
\begin{table}[h]
    \centering
    \small
    \renewcommand{\arraystretch}{0.8}
    \caption{Eikonal evaluation for signed distance field prediction under full observations(in $L_2$ relative error). Only the forward setting is applicable.}
    \setlength{\tabcolsep}{1em}
    \begin{tabular}{lcc}
    \toprule
       & \textbf{Steps} & \textbf{Relative $L_2$ Error } \\
    \midrule
    \textbf{DiffusionPDE} & 2000 & 35.83\% \\
    \textbf{FunDPS} & 200 & 7.77\% \\
    \midrule
    \textbf{PRISMA (ours)} & 20 & \underline{6.97\%} \\
    \textbf{PRISMA (ours)} & 50 & \textbf{6.87\%} \\
    \bottomrule
    \end{tabular}
    \label{tab:eikonal_full_forward}
\end{table}
\vspace{-1em}

\begin{figure}[h]
    \centering
        \includegraphics[width=0.9\linewidth]{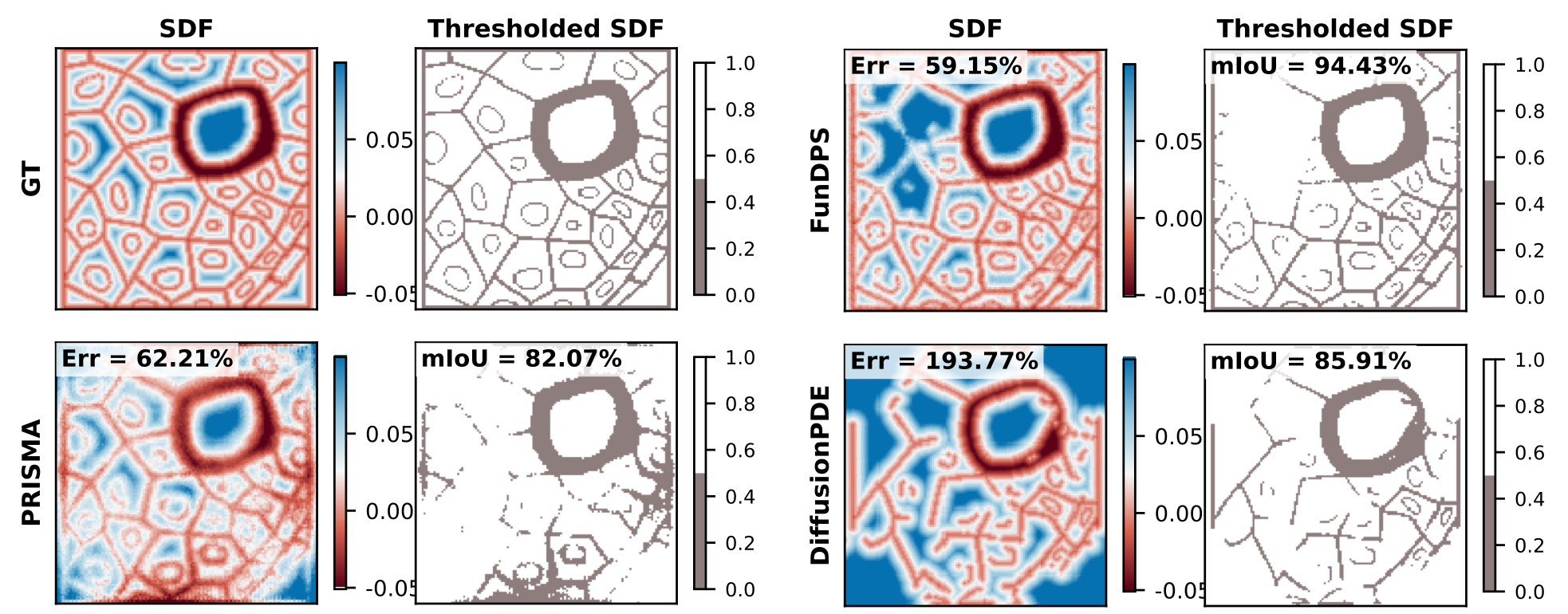}
        \caption{\small Qualitative comparison on a second Eikonal test sample (see Figure~\ref{fig:eikonal_main} for the first). The left column shows the predicted signed distance field (SDF), and the right column shows the binary reconstruction obtained by thresholding the SDF. We compare ground truth with PRISMA, FunDPS, and DiffusionPDE. For the SDF panels, we report relative $L_2$ error; for the thresholded reconstructions, we report mIoU.}
        \label{fig:eikonal_91}
\end{figure}
\vspace{-1em}
\section{Ablations}
\label{appendix:ablations}
\subsection{Ablation on the Number of Inference Steps}
\label{sec:inf_steps_ablation}

To analyze the convergence behavior of PRISMA, we evaluate its performance across a range of inference steps ($N$) from 20 to 500. This sensitivity analysis is conducted for both the challenging {noisy observation} setting and the ideal {full observation} setting (Table~\ref{tab:abla_iterations_combined}). The results clearly demonstrate that PRISMA converges remarkably quickly. In both scenarios, performance saturates early, with minimal to no improvement observed beyond 20-50 steps. This rapid convergence justifies our use of a small number of steps for inference, as it provides an optimal balance between computational efficiency and accuracy.

\begin{table*}[!htbp]
    \centering
    \caption{
    Performance of PRISMA (relative $L_2$ error \%) as the number of
    inference steps ($N$) is varied under noisy and full observations.
    }
    \label{tab:abla_iterations_combined}
    \fontsize{9pt}{9pt}\selectfont
    \setlength{\tabcolsep}{0.35em}
    \renewcommand{\arraystretch}{0.92}
    \begin{tabular}{clccccccc}
    \toprule
    \multirow{2}{*}{\textbf{Obs.}} &
    \multirow{2}{*}{\textbf{Steps} $(N)$} &
    \multicolumn{2}{c}{\textbf{Darcy Flow}} &
    \multicolumn{2}{c}{\textbf{Poisson}} &
    \multicolumn{2}{c}{\textbf{Helmholtz}} \\
    \cmidrule(lr){3-4}
    \cmidrule(lr){5-6}
    \cmidrule(lr){7-8}
    & & \textbf{Fwd} & \textbf{Inv}
      & \textbf{Fwd} & \textbf{Inv}
      & \textbf{Fwd} & \textbf{Inv} \\
    \midrule

    \multirow{5}{*}{\textbf{Noisy}}
      & 20  & 12.10 & 22.93 & 18.20 & 41.46 & 17.65 & 68.43 \\
      & 50  & 12.06 & 22.97 & 18.10 & 40.90 & 16.50 & 67.50 \\
      & 100 & 12.06 & 22.88 & 18.00 & 40.90 & 16.50 & 67.50 \\
      & 200 & 12.09 & 22.91 & 17.70 & 40.90 & 16.50 & 67.50 \\
      & 500 & 12.06 & 23.03 & 18.00 & 40.80 & 16.40 & 67.50 \\
    \midrule

    \multirow{5}{*}{\textbf{Full}}
      & 20  & 1.10 & 3.80 & 4.58 & 10.90 & 8.12 & 11.03 \\
      & 50  & 1.05 & 3.79 & 4.00 & 10.70 & 7.11 & 10.76 \\
      & 100 & 1.04 & 3.78 & 4.00 & 10.60 & 6.94 & 10.75 \\
      & 200 & 1.03 & 3.78 & 4.00 & 10.60 & 6.92 & 10.70 \\
      & 500 & 1.04 & 3.78 & 4.00 & 10.60 & 6.80 & 10.70 \\
    \bottomrule
    \end{tabular}
\end{table*}

\subsection{Residual and Accuracy During Sampling}
\label{app:convergence_tracking}
\begin{figure}[!htb]
    \centering

    \begin{subfigure}{0.95\linewidth}
        \centering
        \includegraphics[width=0.9\linewidth]
        {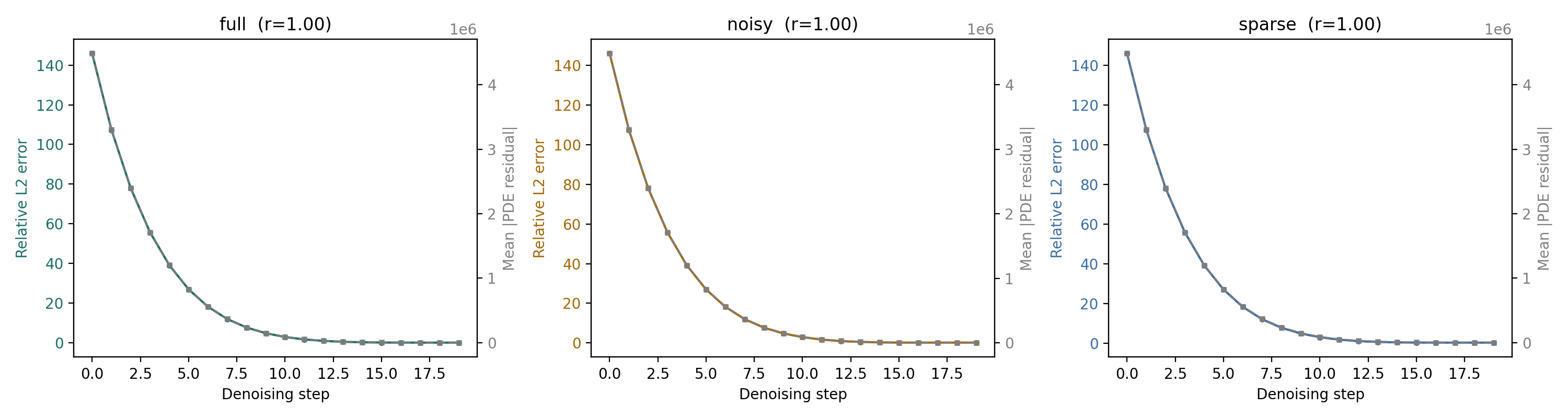}
        \caption{Helmholtz forward, 20 steps ($r>0.999$).}
        \label{fig:helmholtz_for_combined_20}
    \end{subfigure}

    \vspace{-0.1em}

    \begin{subfigure}{0.95\linewidth}
        \centering
        \includegraphics[width=0.9\linewidth]
        {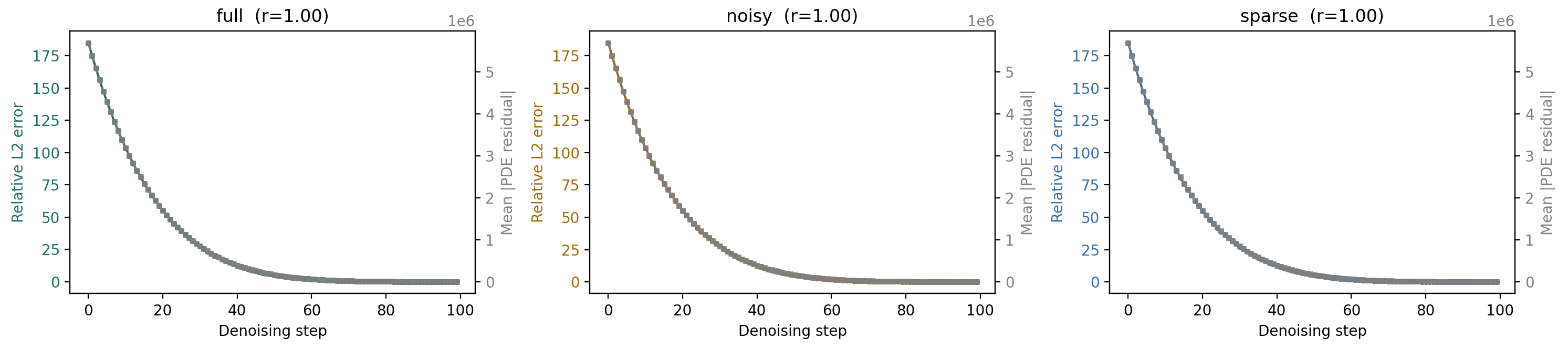}
        \caption{Helmholtz forward, 100 steps ($r>0.99$).}
        \label{fig:helmholtz_for_combined_100}
    \end{subfigure}

    \vspace{-0.1em}

    \begin{subfigure}{0.95\linewidth}
        \centering
        \includegraphics[width=0.9\linewidth]
        {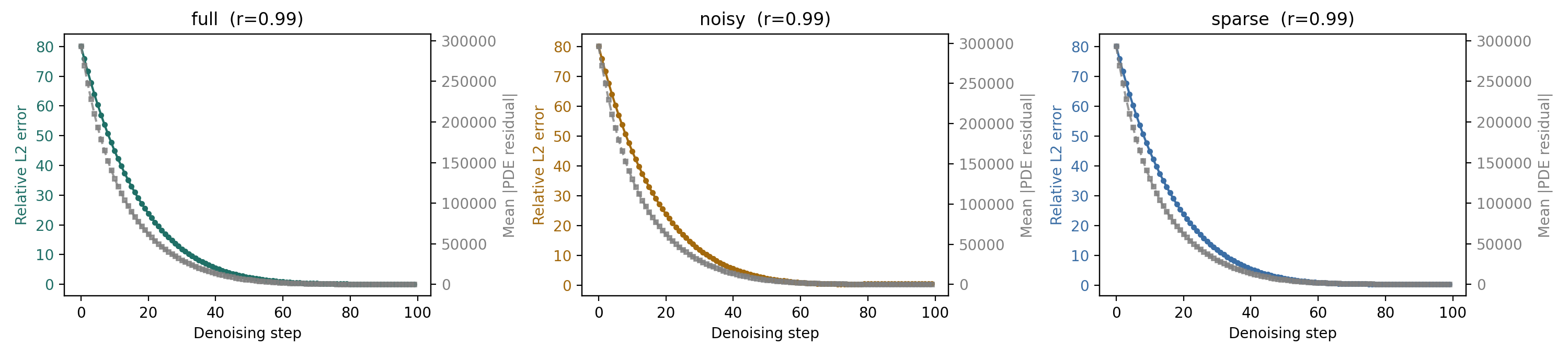}
        \caption{Kolmogorov forward, 100 steps ($r\approx0.99$).}
        \label{fig:kolmogorov_for_combined_100}
    \end{subfigure}

    \vspace{-0.5em}

    \caption{\small
    Accuracy and PDE-residual norm across the denoising trajectory.
    In all three settings, prediction accuracy is strongly correlated
    with the reduction in PDE-residual norm.
    }
    \label{fig:accuracy_residual_trajectories}
\end{figure}

Section \ref{sec:inf_steps_ablation} evaluates accuracy at a fixed set of step counts (20--500). Here (Figure~\ref{fig:accuracy_residual_trajectories}) we instead track continuous trajectories: at every step we record the relative-$L_2$ error of the intermediate prediction against ground truth and the aggregate PDE-residual norm, then compute the
Pearson correlation between the two per-step curves. We run this for Helmholtz (both a 20-step trajectory matching our default inference budget and a longer 100-step trajectory) and for Kolmogorov flow (100-step), forward direction, under full, sparse, and noisy observations (50 test samples per configuration). Reported error is the raw relative-$L_2$ fraction, not the $\times100$ percentage used in main tables.

Across all three trajectories, error and residual norm track each other closely throughout sampling (Pearson $r>0.99$ in every case), dropping sharply in the first 10--15 (of 20) or 40--50 (of 100) steps and decreasing more slowly thereafter. This is consistent with the saturation trend in
PRISMA's noise schedule (Table~\ref{tab:parameter}, $\rho=7$) is spaced nonlinearly and concentrates most steps in the low-noise regime, so most of the correction happens early, with additional steps mainly refining numerical precision rather than performing new corrections.

\subsection{Comparison at 20 steps}
To highlight the inference efficiency of our method, we conduct a direct comparison where all models are restricted to just 20 sampling steps, a regime where PRISMA excels. While the optimal performance for FunDPS and DiffusionPDE is achieved at much higher step counts (200--500 and 2000 steps, respectively), this analysis serves as a stress test to evaluate per-step convergence speed. As shown in Table~\ref{tab:abla_20steps_combined}, \ourmethod consistently achieves low error rates across all tasks in both the noisy \& full observation settings. In contrast, the baseline models produce significantly higher errors, with FunDPS often failing to converge entirely. 
\begin{table*}[!h]
    \centering
    \caption{Comparative performance (relative $L_2$ error \%) on noisy \& full observation tasks with 20 steps.}
    \setlength{\tabcolsep}{0.35em}
    \fontsize{9pt}{9pt}\selectfont
    \begin{tabular}{c l c c c c c c c}
    \toprule
    \textbf{Obs.} & \multirow{2}{*}{\textbf{Model}} & \multirow{2}{*}{\textbf{Steps} $(N)$} &
    \multicolumn{2}{c}{\textbf{Darcy Flow}} &
    \multicolumn{2}{c}{\textbf{Poisson}} &
    \multicolumn{2}{c}{\textbf{Helmholtz}} \\
    \cmidrule(lr){4-5}\cmidrule(lr){6-7}\cmidrule(lr){8-9}
    \textbf{Mode} & & & \textbf{Fwd} & \textbf{Inv} & \textbf{Fwd} & \textbf{Inv} & \textbf{Fwd} & \textbf{Inv} \\
    \midrule

    \multirow{3}{*}{Noisy}
      & \textbf{FunDPS} & 20 & 99.99 & 73.00 & 99.99 & 255.09 & 99.99 & 178.09 \\
      & \textbf{DiffusionPDE} & 20 & 37.16 & 70.85 & 125.15 & 148.97 & 123.44 & 133.35 \\
      & \textbf{PRISMA (ours)} & 20 & \textbf{12.29} & \textbf{23.15} & \textbf{18.58} & \textbf{41.75} & \textbf{17.93} & \textbf{68.87} \\
    \midrule

    \multirow{3}{*}{Full}
      & \textbf{FunDPS} & 20 & 8.88 & 17.75 & 9.755 & 39.15 & 10.08 & 39.39 \\
      & \textbf{DiffusionPDE} & 20 & 30.99 & 69.82 & 95.26 & 123.21 & 111.77 & 101.01 \\
      & \textbf{PRISMA (ours)} & 20 & \textbf{1.10} & \textbf{3.80} & \textbf{4.58} & \textbf{10.90} & \textbf{8.12} & \textbf{11.03} \\
    \bottomrule
    \end{tabular}
    \label{tab:abla_20steps_combined}
\end{table*}

\subsection{Fidelity to Physical Constraints}

To assess the physical consistency of the generated solutions, we analyze the statistical properties of the PDE residual fields. A well-trained model should produce residuals with a distribution close to a standard normal. A kurtosis value close to zero indicates that the errors are random and well-behaved, rather than systematic.

\begin{figure}[h]
  \centering
  \begin{subfigure}[t]{0.48\linewidth}
    \centering
    \includegraphics[width=0.75\linewidth]{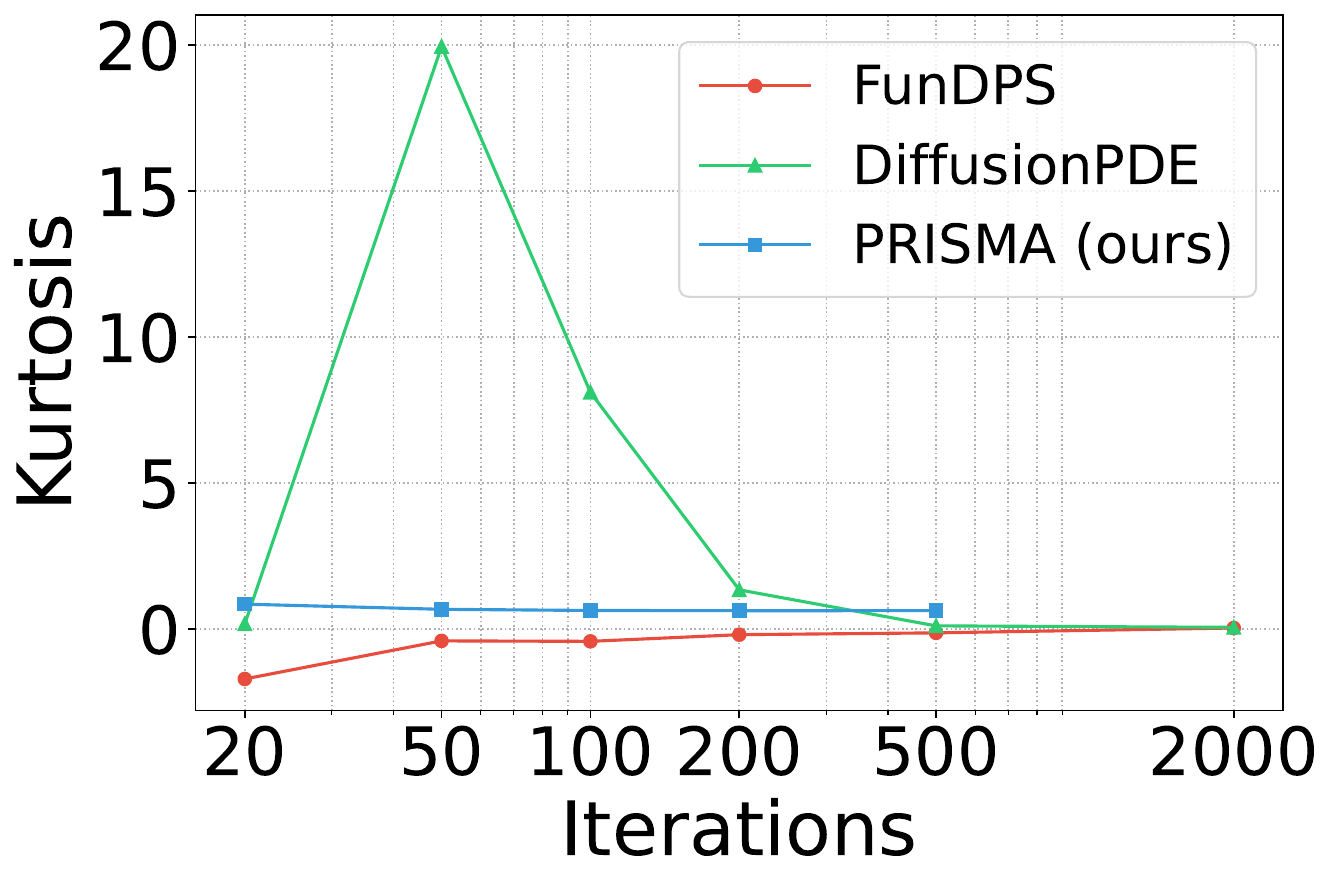}
    \caption{Forward Probem}
    \label{fig:kurt-fwd}
  \end{subfigure}\hfill
  \begin{subfigure}[t]{0.48\linewidth}
    \centering
    \includegraphics[width=0.75\linewidth]{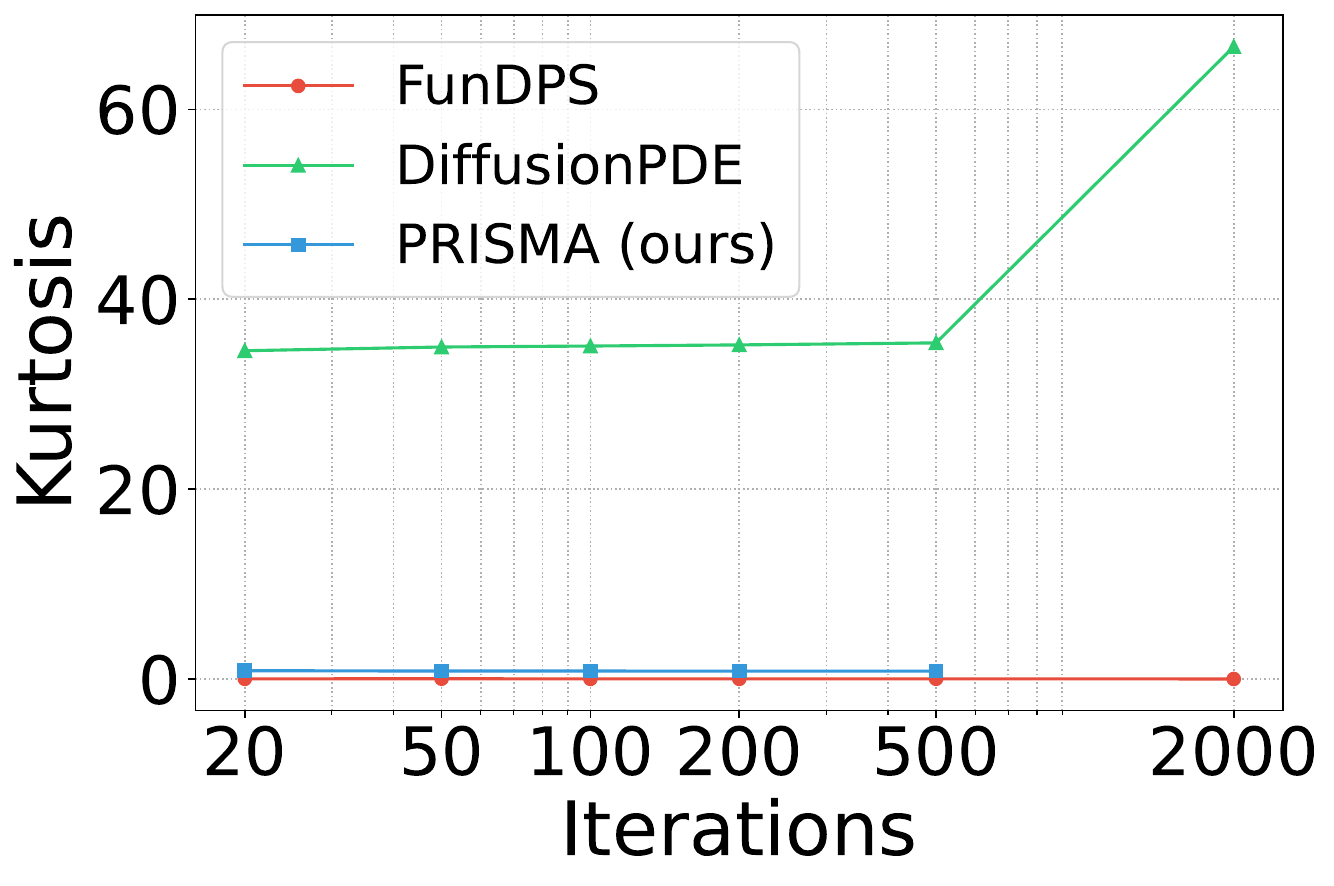}
    \caption{Inverse Problem}
    \label{fig:kurt-inv}
  \end{subfigure}

  \caption{Kurtosis of the PDE residual field plotted against the number of inference iterations for the Poisson problem. }
  \label{fig:noise_kurtosis}
\end{figure}

Figure~\ref{fig:noise_kurtosis} plots the kurtosis of the PDE residual for the Poisson problem against the number of inference iterations. The results show that {PRISMA's} residual kurtosis (blue line) is consistently stable and near zero across all iterations for both the forward and inverse problems. In contrast, DiffusionPDE (green line) exhibits highly unstable and large kurtosis values, indicating that its solutions contain significant non-Gaussian errors or physical inconsistencies. This analysis demonstrates that PRISMA's architectural guidance leads to more physically robust solutions.

\subsection{{SRA Ablations}}
\label{sec:app:sra_ablatioons}
{\textbf{SRA Ablations:} We further analyze the Spectral Residual Attention (SRA) block by running ablations 
at $64{\times}64$ with $N{=}20$ inference steps Table~\ref{tab:gating_attention_ablation}. We compare the following setups: (1) \textbf{No Gating}  where $g_{\text{res}}{=}1$ effectively; (2) \textbf{Multiplicative gate} where there is no weighted skip connection, (3) \textbf{Only magnitude} where we replace cross-attention with a per-mode weight from $\lvert\hat{r}(k)\rvert$, (4) \textbf{Attention w/o phase} which considers the magnitude only and is phase-blind ($\lvert\hat{x}(k)\rvert\!\cdot\!\lvert\hat{r}(k)\rvert$), (5) \textbf{Without guided PDE residual} where we do not guide the PDE residual as described in Appendix Section \ref{appendix:inference_algo} , and \textbf{Ours (PDE Res SRA)}. We observe that the choice of gating matters, the skip-connected scalar gate stabilizes residual injection and helps notably in noisy inverse settings. Furthermore,  we see that phase-aware spectral attention outperforms magnitude-only/phase-blind variants, indicating that per-mode phase alignment provides useful physics signal and lastly residual guidance helps SRA consistently improve over no-guidance on inverse tasks while remaining competitive on forward tasks, more so in noisy observation settings.}

\begin{table*}[!t]
    \centering
    \caption{{SRA Ablations (comparing $L_2$ error \%) of gating and attention mechanisms on 64$\times$64 resolution}}
     \fontsize{9pt}{9pt}\selectfont
    {\color{black}
    \begin{tabular}{l c c c c c}
    \toprule
      & \multirow{2}{*}{\textbf{Steps} $(N)$} 
      & \multicolumn{2}{c}{\textbf{Full Helmholtz}} 
      & \multicolumn{2}{c}{\textbf{Noisy Helmholtz}} \\
    \cmidrule(lr){3-4}
    \cmidrule(lr){5-6}
      & & \textbf{Forward} & \textbf{Inverse} 
        & \textbf{Forward} & \textbf{Inverse} \\
    \midrule

    \textbf{No Gating} & 20 &
    $3.8$ & $28.91$ &
    $33.8$ & \underline{$97.5$} \\

    \textbf{Multiplicative Gate \emph{(no skip)}} & 20 &
    \underline{$3.3$} & \underline{$12.98$} &
    $\textbf{24.7}$ & $115.07$ \\

    \textbf{Only Mag of PDE Residual \emph{(w/o attention)}} & 20 &
    $3.29$ & $13.42$ &
    \underline{$25.44$} & $115.26$ \\

    \textbf{Attention \emph{(w/o Phase)}} & 20 &
    $6.5$ & $24.77$ &
    $34$ & $102.19$ \\

    \textbf{Without Guided PDE Residual} & 20 &
    $3.52$ & $14.26$ &
    $31.26$ & $98.99$ \\
    
    \textbf{Ours (PDE Res SRA)} & 20 &
    \underline{$3.48$} & $\textbf{12.58}$ &
    $30.8$ & $\textbf{93.2}$ \\
    \bottomrule
    \end{tabular}
    }
    \label{tab:gating_attention_ablation}
\end{table*}

\section{{Spatiotemporal Field Reconstruction}}
\label{appendix:spatio-temporal}

We evaluate \ourmethod\ on time-dependent PDE settings to examine whether residual-aware
diffusion can extend beyond static 2D fields. We consider two spatiotemporal experiments:
(i) reconstruction of the full trajectory of the 1D Burgers' equation from sparse sensors, and
(ii) rollout prediction for 2D Navier--Stokes dynamics under full, sparse, and noisy observations.

\begin{figure}[htbp]
    \centering
    \captionsetup{aboveskip=2pt}
    \begin{subfigure}[t]{0.7\linewidth} 
        \centering
        \includegraphics[trim={0cm 0cm 0cm 0cm}, clip, width=\linewidth]{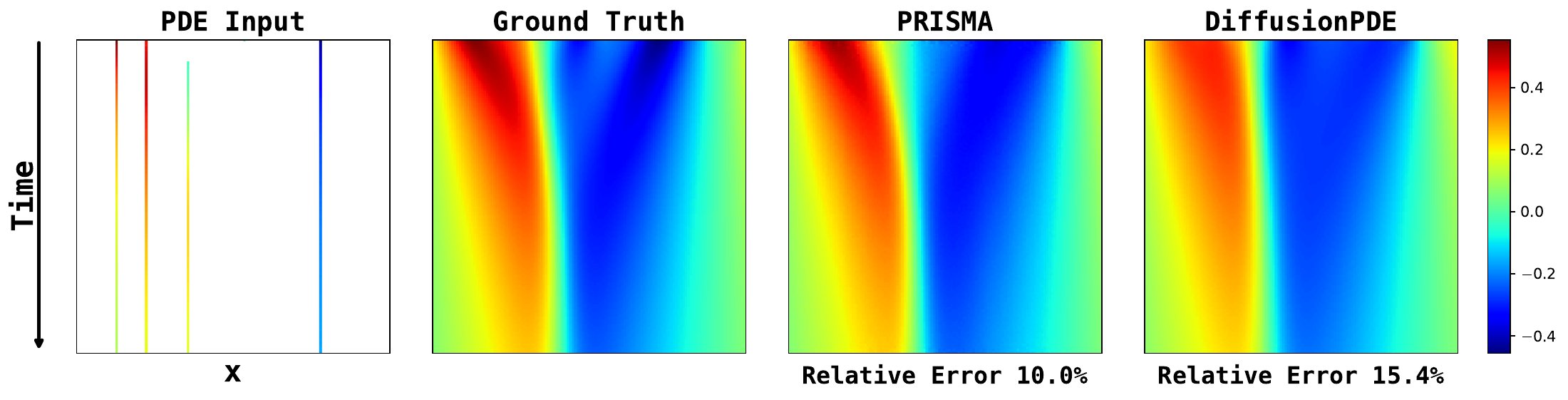}
    \end{subfigure}
    
    \vspace{0.008cm}

    \begin{subfigure}[t]{0.7\linewidth}
        \centering
        \includegraphics[trim={0.0cm 0cm 0cm 0cm}, clip, width=\linewidth]{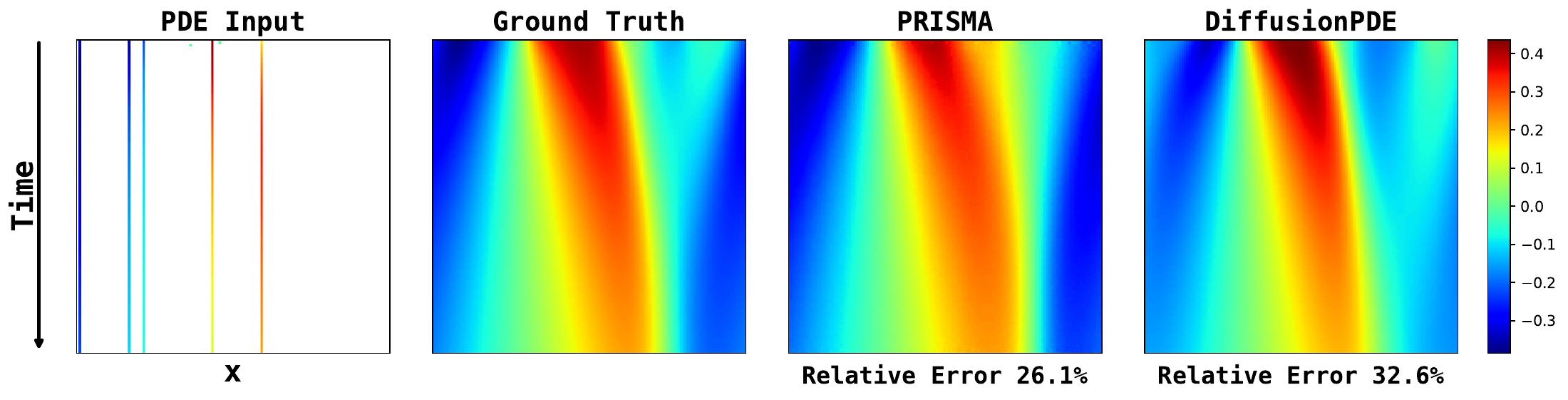}
    \end{subfigure}

    \caption{{Prediction (\textit{for two samples}) Visualization on Burgers' equation}}
    \label{fig:burger_comparison}
\end{figure}

\paragraph{Burgers' equation.}
We first evaluate \ourmethod\ on the 1D dynamic Burgers' equation, where the goal is to recover
the full solution trajectory $u_{0:T}$ over the interval $[0,T]$ from spatially sparse but temporally
continuous sensor observations. Following DiffusionPDE, we model the full spatiotemporal field
as a 2D diffusion sample. \ourmethod\ outperforms DiffusionPDE, achieving a test relative error of
$9.33\%$ compared to $10.39\%$. Figure~\ref{fig:burger_comparison} shows qualitative examples of the reconstructed Burgers' trajectories.

\paragraph{2D+time Navier--Stokes rollout.}
We evaluate a 2D+time variant of \ourmethod\ on Navier--Stokes rollout prediction. Instead of predicting only a
single next-step field, this variant jointly predicts the next three timesteps as output channels. We
compare this model against next-step DiffusionPDE, FunDPS, and \ourmethod\ models, which are
iteratively rolled out over $T=10$ timesteps. Table~\ref{tab:navierstokes_rollout_forward} reports
the final rollout error under full, sparse, and noisy observation settings. The 2D+time variant remains
competitive in the full and sparse settings, and is substantially more robust in the noisy setting,
suggesting that jointly modeling short temporal windows helps preserve Navier--Stokes dynamics
under corrupted observations. 

\begin{table}[t]
    \centering
    \fontsize{9pt}{9pt}\selectfont
    \caption{Relative $L_2$ error for Navier--Stokes rollout prediction under full, sparse, and noisy observation settings over $T=10$ timesteps. We compare DiffusionPDE, FunDPS, PRISMA, and the PRISMA 2D+Time variant. DiffusionPDE, FunDPS, and PRISMA are trained in a next-step prediction manner whereas PRISMA 2D+Time directly predicts channels across time. They are all iteratively rolled out over time to obtain the final trajectory. All values are reported on the final rollout result as percentages ($\times 100$)}
    \setlength{\tabcolsep}{0.6em}
    \begin{tabular}{l c c c c}
    \toprule
    & \multirow{2}{*}{\textbf{Steps} $(N)$} & \multicolumn{3}{c}{\textbf{Forward}} \\
    \cmidrule(lr){3-5}
    & & \textbf{Full} & \textbf{Sparse} & \textbf{Noisy} \\
    \midrule

    \textbf{DiffusionPDE} & 2000 & 94.19\% & 92.85\% & 101.81\% \\
    \textbf{FunDPS} & 200 & 5.68\% & 7.09\% & \underline{21.68\%} \\
    \midrule
    \textbf{PRISMA (\textit{w/ 2D+Time})} & 20 & \underline{2.87\%} & \underline{5.07\%} & \textbf{2.84\%} \\
    \textbf{PRISMA (ours)} & 20 & \textbf{2.27\%} & \textbf{4.07\%} & 35.37\% \\

    \bottomrule
    \end{tabular}
    \label{tab:navierstokes_rollout_forward}
\end{table}
\vspace{-1ex}
\section{{Super-resolution Analysis}}
\begin{table*}
    \centering
    \caption{{Performance metrics at different resolutions. Average batch time reported for batch size=50}}
    \setlength{\tabcolsep}{0.5em}
    \renewcommand{\arraystretch}{1.2}
    \resizebox{1.0\textwidth}{!}{%
    {\color{black}
    \begin{tabular}{c c c c}
        \toprule
        \textbf{Resolution} & \textbf{Avg. Batch Time (s)} & \textbf{Avg. Per-Sample Time (s)} & \textbf{Avg. Peak GPU (reserved) Mem (GB)} \\
        \midrule
        64x64   & $12.140 \pm 0.945$ & $0.242 \pm 0.018$ & $4.117 \pm 0.09$ \\
        128x128 & $20.036 \pm 0.326$ & $0.400 \pm 0.006$ & $15.383 \pm 0.804$ \\
        256x256 & $60.892 \pm 0.580$ & $1.217 \pm 0.011$ & $62.602 \pm 0.804$ \\
        512x512 & $242.489 \pm 0.423$ & $4.849 \pm 0.008$ & $107.114 \pm 0.399$ \\
        \bottomrule
    \end{tabular}}}
    \label{tab:super-res}
\end{table*}

\label{appendix:super-res}
{We evaluate PRISMA’s ability to generalize to higher spatial resolutions at inference using a model trained only on 64×64 resolution. For each target resolution (64, 128, 256, and 512), we upsample the 64×64 PDE input and run the same 20-step PRISMA model, without retraining the architecture, to generate the corresponding solution. Qualitative forward and inverse results are shown in {Figure \ref{fig:superres_qualitative} and the corresponding runtime and memory measurements are reported in Table \ref{tab:super-res}. Figure \ref{fig:super_res_metrics} visualizes how inference cost grows with resolution, specifically the per-sample time increases smoothly from 0.24s  $\rightarrow$  4.85s, and reserved GPU memory rises from 4 GB  $\rightarrow$  107 GB. While the 512×512 case incurs a significantly higher memory footprint, PRISMA remains stable across all resolutions tested.} Note timings here differ from Table \ref{tab:train_speed} since this model is trained at 64×64 and evaluated at upsampled resolutions with batch size 50, rather than natively trained/evaluated at each resolution.

\begin{figure}[h]
    \centering
    \includegraphics[width=0.5\textwidth]{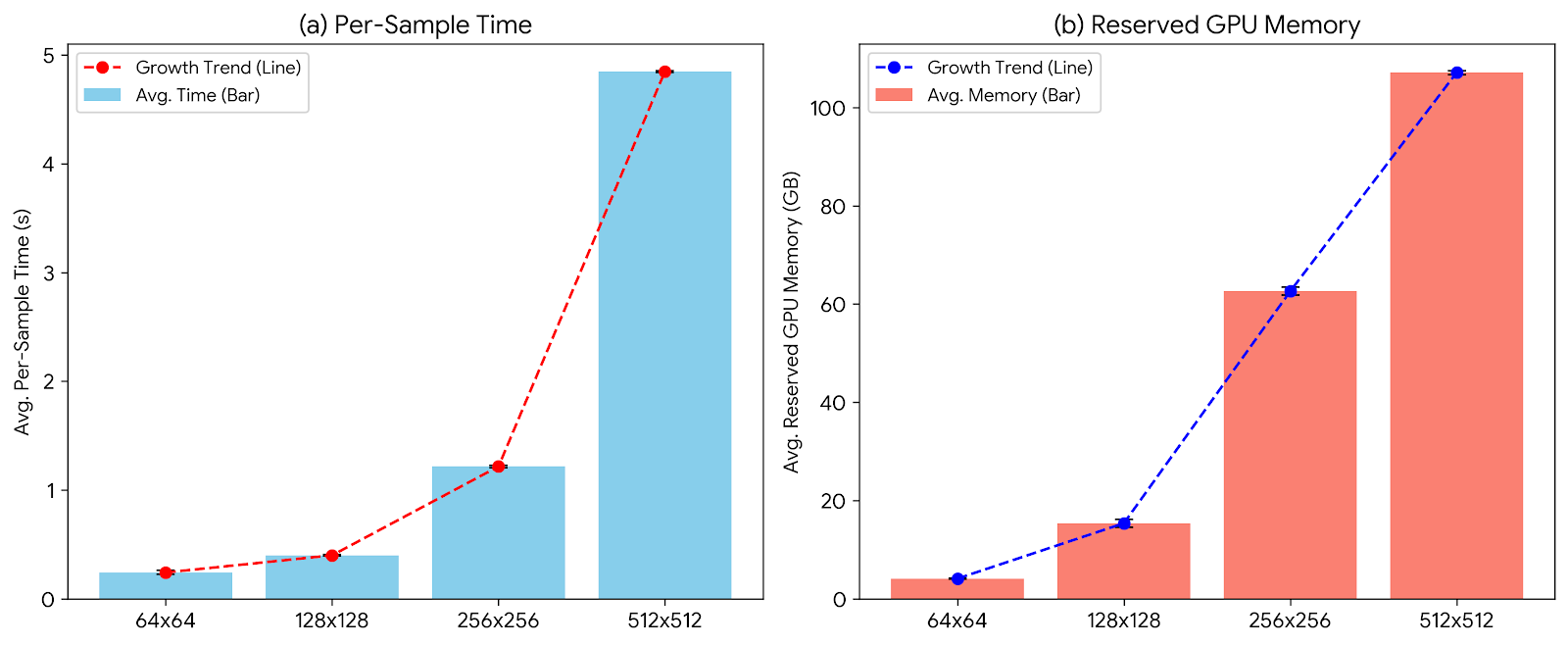}
    \caption{{Performance metrics (Per-Sample Time and Peak Reserved GPU Memory) as a function of inference resolution.}}
    \label{fig:super_res_metrics}
\end{figure}

\begin{table*}[!htb]
    \centering
    \caption{Comparison of different models on three PDE problems under varying levels of additive
    Gaussian noise corruption (10--90\% of pixels), simulating real-world measurement noise ($L_2$
    relative error for all \& error rate for Darcy inverse). \textbf{Best} is bolded, \underline{second-best}
    underlined (within each noise-level block).}
    \setlength{\tabcolsep}{0.35em}
    \color{black}
    \begin{tabular}{c l c c c c c c c}
    \toprule
    \textbf{Noise} & & \multirow{2}{*}{\textbf{Steps} $(N)$}
      & \multicolumn{2}{c}{\textbf{Darcy Flow}}
      & \multicolumn{2}{c}{\textbf{Poisson}}
      & \multicolumn{2}{c}{\textbf{Helmholtz}} \\
    \cmidrule(lr){4-5}\cmidrule(lr){6-7}\cmidrule(lr){8-9}
    \textbf{Level} & & & \textbf{Forward} & \textbf{Inverse}
        & \textbf{Forward} & \textbf{Inverse}
        & \textbf{Forward} & \textbf{Inverse} \\
    \midrule

    \multirow{5}{*}{90\%}
      & \textbf{FNO}  & -- & \underline{9.7}\% & {52.3\%} & 80.48\% & 1.44e07\% & 80.04\% & 2.92e05\% \\
      & \textbf{PINO} & -- & 70.42\% & {52.2\%} & 80.44\% & 1.12e06\% & 107.15\% & 8.74e05\% \\
      & \textbf{DiffusionPDE} & 2000 & 36.89\% & 70.04\% & 44.51\% & \underline{129.08\%} & 23.2\% & \underline{113.49\%} \\
      & \textbf{FunDPS} & 200 & {10.81\%} & \underline{48.24\%} & \underline{16.07\%} & 957.87\% & \underline{16.96\%} & 622.6\% \\
      & \textbf{PRISMA} & 20 & \textbf{9.65\%} & \textbf{22.83\%} & \textbf{15.69\%} & \textbf{40.34\%} & \textbf{16.16\%} & \textbf{66.93\%} \\
    \midrule

    \multirow{5}{*}{50\%}
      & \textbf{FNO}  & -- & \underline{9.17}\% & {49.93\%} & 55.84\% & 1.083e07\% & 78.31\% & 2.19e05\% \\
      & \textbf{PINO} & -- & 45.53\% & {49.87\%} & 61.04\% & 8.56e05\% & 62.19\% & 6.84e05\% \\
      & \textbf{DiffusionPDE} & 2000 & 36.4\% & 68.1\% & 39.61\% & \underline{127.49\%} & 22.58\% & \underline{113.17\%} \\
      & \textbf{FunDPS} & 200 & 9.18\% & \underline{46.6\%} & \underline{13.81\%} & 806.45\% & \underline{14.33\%} & 467.63\% \\
      & \textbf{PRISMA} & 20 & \textbf{8.22\%} & \textbf{20.3\%} & \textbf{12.71\%} & \textbf{35.49\%} & \textbf{12.04\%} & \textbf{59.99\%} \\
    \midrule

    \multirow{5}{*}{30\%}
      & \textbf{FNO}  & -- & \underline{8.48}\% & {49.85\%} & 14.32\% & 836.31\% & 48.2\% & 773.04\% \\
      & \textbf{PINO} & -- & 10.06\% & {49.83\%} & 11.49\% & 682.01\% & 40.23\% & 531.61\% \\
      & \textbf{DiffusionPDE} & 2000 & 35.21\% & 63.26\% & 25.78\% & \underline{117.36\%} & 22.14\% & \underline{108.04\%} \\
      & \textbf{FunDPS} & 200 & 8.53\% & \underline{42.5\%} & \underline{11.28\%} & 796.22\% & \underline{13.15\%} & 384.78\% \\
      & \textbf{PRISMA} & 20 & \textbf{6.23\%} & \textbf{17.97\%} & \textbf{11.22\%} & \textbf{32.16\%} & \textbf{9.16\%} & \textbf{54.43\%} \\
    \midrule

    \multirow{5}{*}{10\%}
      & \textbf{FNO}  & -- & 7.68\% & {49.793\%} & 9.36\% & 477.67\% & 23.56\% & 511.91\% \\
      & \textbf{PINO} & -- & \underline{5.35\%} & {49.78\%} & \textbf{6.52\%} & 461.09\% & 17.9\% & 413.07\% \\
      & \textbf{DiffusionPDE} & 2000 & 33.12\% & 53.12\% & 17.51\% & \underline{107.14\%} & 19.99\% & \underline{107.72\%} \\
      & \textbf{FunDPS} & 200 & 8.5\% & \underline{31.34\%} & 10.63\% & 598.7\% & \underline{10.84\%} & 239.65\% \\
      & \textbf{PRISMA} & 20 & \textbf{3.77\%} & \textbf{12.34\%} & \underline{7.3\%} & \textbf{27.12\%} & \textbf{8.05\%} & \textbf{43.56\%} \\
    \bottomrule
    \end{tabular}
    \label{tab:noisy_all}
\end{table*}

\section{{Noise Robustness Evaluation}}
\label{appendix:noise_percent}
{We evaluate PRISMA and all baselines under varying levels of additive measurement noise. For each sample, we corrupt a fixed percentage of pixels (90\%, 50\%, 30\%, 10\%) with unit-variance Gaussian noise, keeping the remaining pixels clean. This setup simulates real-world sensor degradation where only part of the field is corrupted. The diffusion models (including PRISMA) are not trained with any noise augmentation beyond the intrinsic diffusion noising, all methods are evaluated out-of-distribution.
Table~\ref{tab:noisy_all} report results across three PDEs in forward and inverse settings. Across all noise levels, classical operator-based models (FNO/PINO) degrade sharply, while diffusion models remain substantially more stable. FunDPS benefits from per-sample DPS optimization but becomes sensitive at high noise levels and in inverse regimes. PRISMA shows consistent performance across all noise ratios, even when a large majority of pixels are corrupted. }

\vspace{-1ex}

\section{{Sparsity Robustness Evaluation}}
\label{appendix:sparsity_robutness}

We further evaluate robustness to different random sparsity levels by varying the fraction of missing observations from 50\% to 90\%. Table~\ref{tab:varying_random_sparsity} reports results for both forward and inverse settings. \ourmethod\ remains stable as sparsity increases, with only modest degradation in the forward problem and consistently strong inverse performance. 

\begin{table*}[t]
    \centering
    \caption{Comparison under varying random sparsity levels (relative $L_2$ error). The mask is sampled randomly, and the sparsity level is varied from 50\% to 90\%.}
    \setlength{\tabcolsep}{0.5em}
    \begin{tabular}{l c c c c c c c}
    \toprule
      & \multirow{2}{*}{\textbf{Steps} $(N)$}
      & \multicolumn{2}{c}{\textbf{50\%}}
      & \multicolumn{2}{c}{\textbf{75\%}}
      & \multicolumn{2}{c}{\textbf{90\%}} \\
    \cmidrule(lr){3-4}
    \cmidrule(lr){5-6}
    \cmidrule(lr){7-8}
      & & \textbf{Forward} & \textbf{Inverse} & \textbf{Forward} & \textbf{Inverse} & \textbf{Forward} & \textbf{Inverse} \\
    \midrule
    \textbf{DiffusionPDE} & 2000 &
    4.25\% & 36.06\% & 4.84\% & 31.87\% & 6.24\% & 24.56\% \\
    \textbf{FunDPS} & 200 &
    \textbf{0.99\%} & 6.10\% & \textbf{1.13\%} & 6.10\% & \textbf{1.59\%} & 6.38\% \\
    
    \midrule
    \textbf{PRISMA (ours)} & 20 &
    1.84\% & \textbf{2.95\%} & 1.92\% & \textbf{3.34\%} & 2.24\% & \textbf{3.87\%} \\
    \textbf{PRISMA (ours)} & 50 &
    \underline{1.76\%} & \underline{2.95\%} & \underline{1.85\%} & \underline{3.35\%} & \underline{2.20\%} & \underline{3.90\%} \\

    \bottomrule
    \end{tabular}
    \label{tab:varying_random_sparsity}
\end{table*}

\begin{wrapfigure}{r}{0.45\linewidth}
    \vspace{-1.2em}
    \centering
    \includegraphics[width=0.95\linewidth]{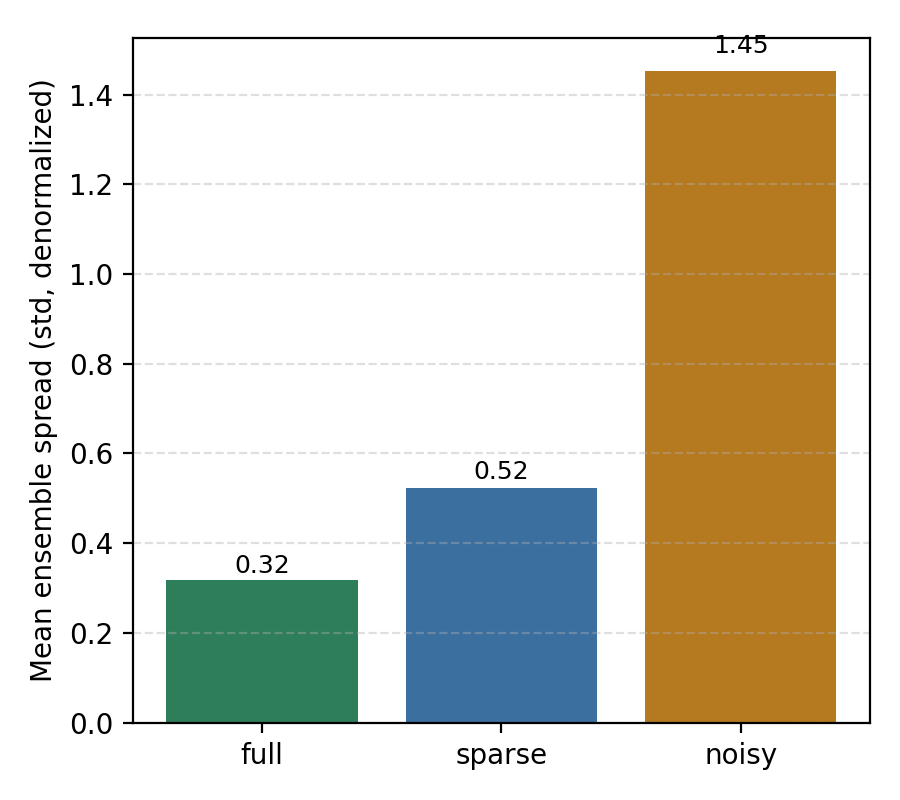}
    \caption{\small Mean ensemble spread across observation modes for Darcy Inverse.}
    \label{fig:darcy_inv_spread}
    \vspace{-1.0em}
\end{wrapfigure}
\section{Posterior Diversity}
\label{app:uncertainty}

To test whether PRISMA produces meaningful posterior samples rather than acting as a deterministic predictor wrapped in a diffusion sampler, we repeat the same conditioning (same observed field, same mask) through the sampler 16 times with independent random latents, for the Darcy inverse problem under full, sparse, and noisy observation. We measure ensemble spread (per-pixel standard deviation across the 16 members, averaged over all pixels to give one value per sample) and its correlation with per-sample prediction error, over 50 held-out test samples per mode.

The spread grows with observation difficulty -- 0.32 (full) to 0.52 (sparse) to 1.45 (noisy), nearly 4x higher under high noise, so the model is not collapsing to a single deterministic output. Spread and per-sample error are positively correlated in all three modes ($r=0.85$
full, $r=0.81$ sparse, $r=0.46$ noisy, $n=50$ samples per mode; Figure
\ref{fig:darcy_inv_scatter}), indicating that samples the model is more uncertain about do tend to be the ones it gets more wrong, not just that uncertainty and error both happen to be larger on average under harder observation regimes.
\begin{figure}[h]
\centering
\includegraphics[width=0.95\linewidth]{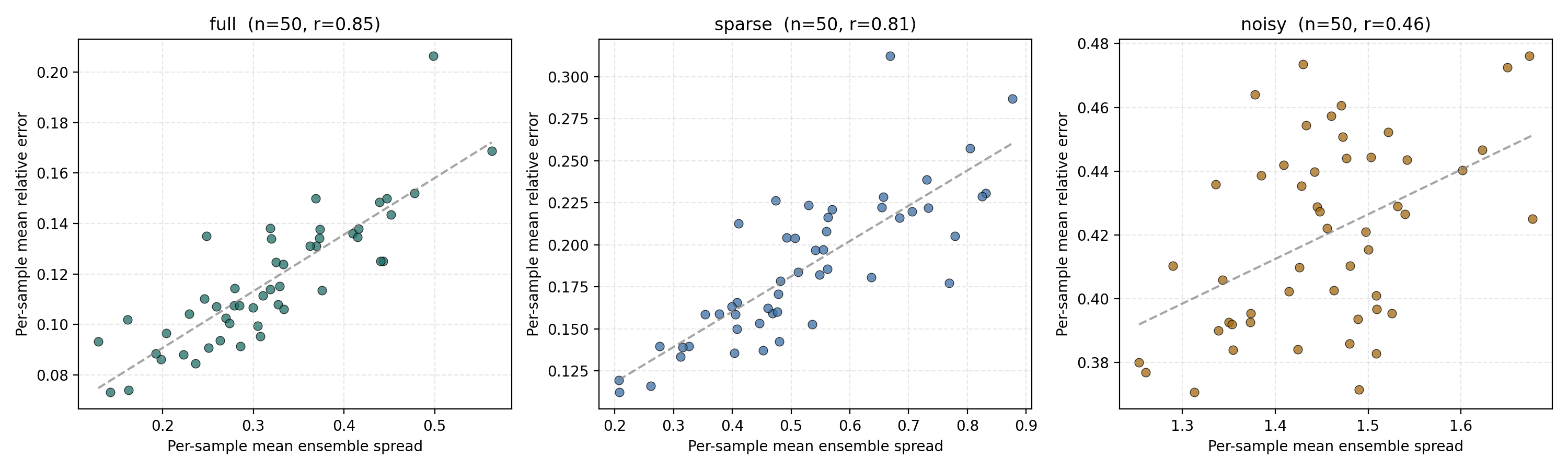}
\caption{Darcy (inverse): per-sample ensemble spread vs.\ per-sample relative error,
50 test samples per mode. The positive trend holds across the full point cloud in
each mode, not just a subset of outliers.}
\label{fig:darcy_inv_scatter}
\end{figure}

\begin{figure}[!htb]
    \centering
    \captionsetup{aboveskip=2pt}

    \begin{subfigure}{0.82\linewidth}
        \centering
        \includegraphics[width=\linewidth]
        {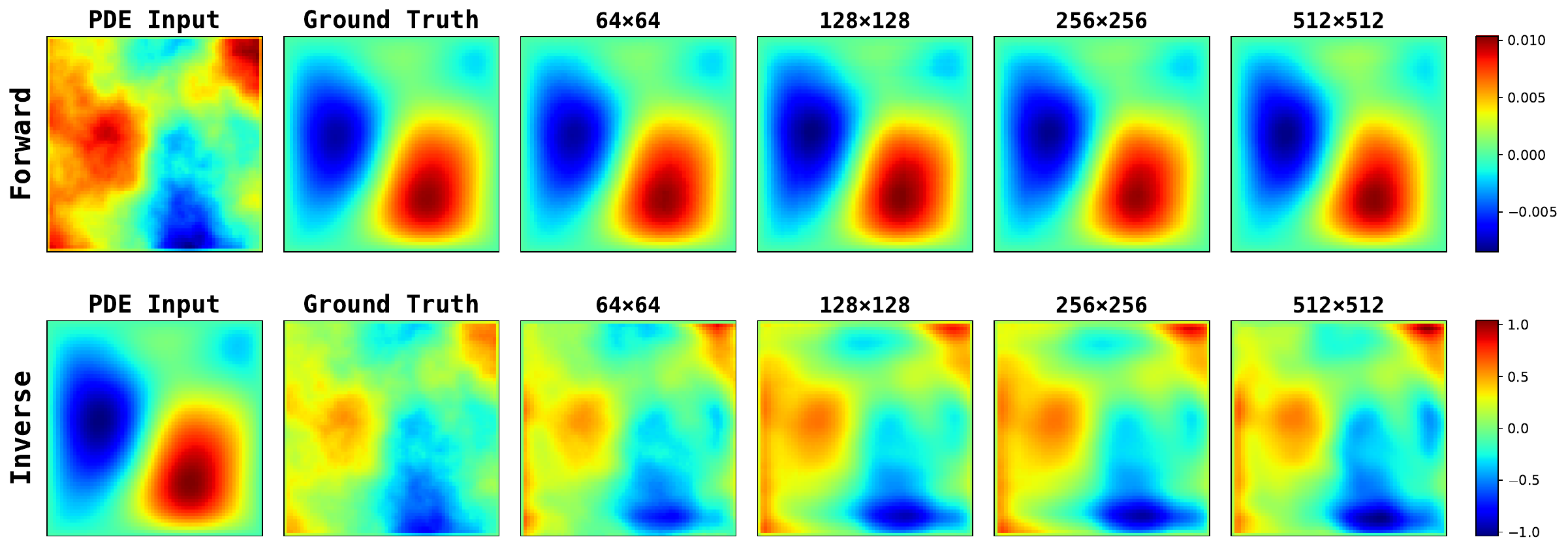}
        \label{fig:helm_71_superres_64}
    \end{subfigure}

    \vspace{-0.5em}

    \begin{subfigure}{0.82\linewidth}
        \centering
        \includegraphics[width=\linewidth]
        {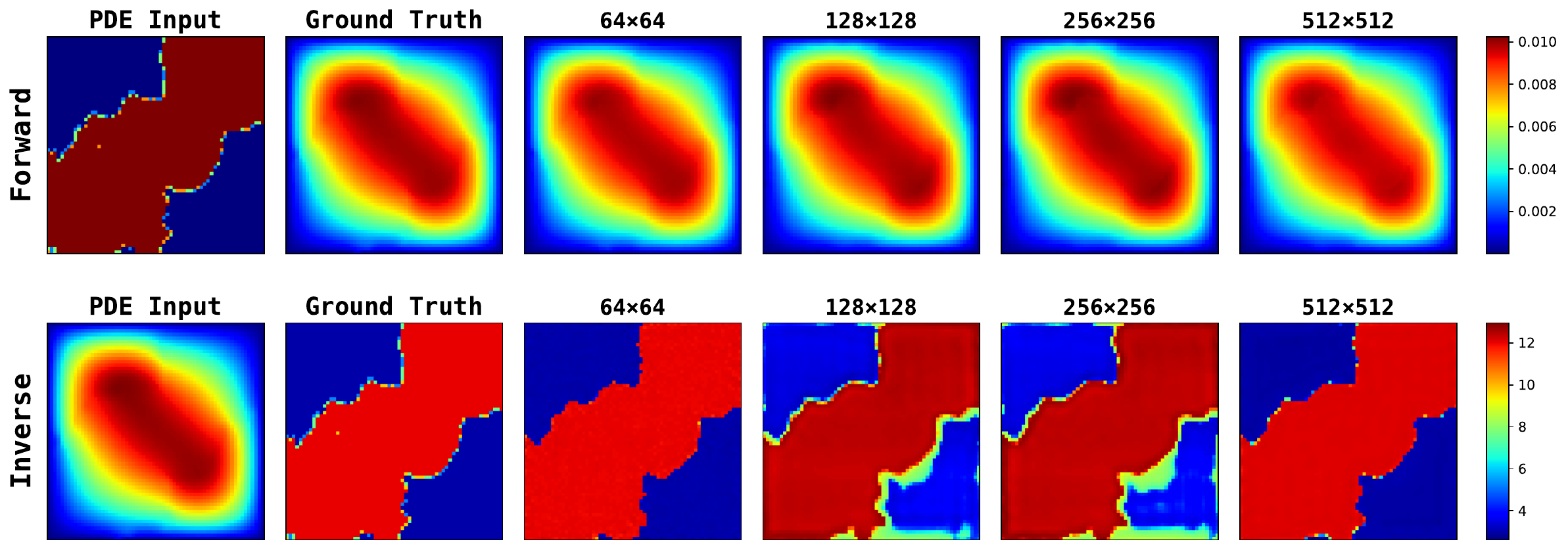}
        \label{fig:darcy_93_superres_64}
    \end{subfigure}

    \vspace{-0.5em}

    \begin{subfigure}{0.82\linewidth}
        \centering
        \includegraphics[width=\linewidth]
        {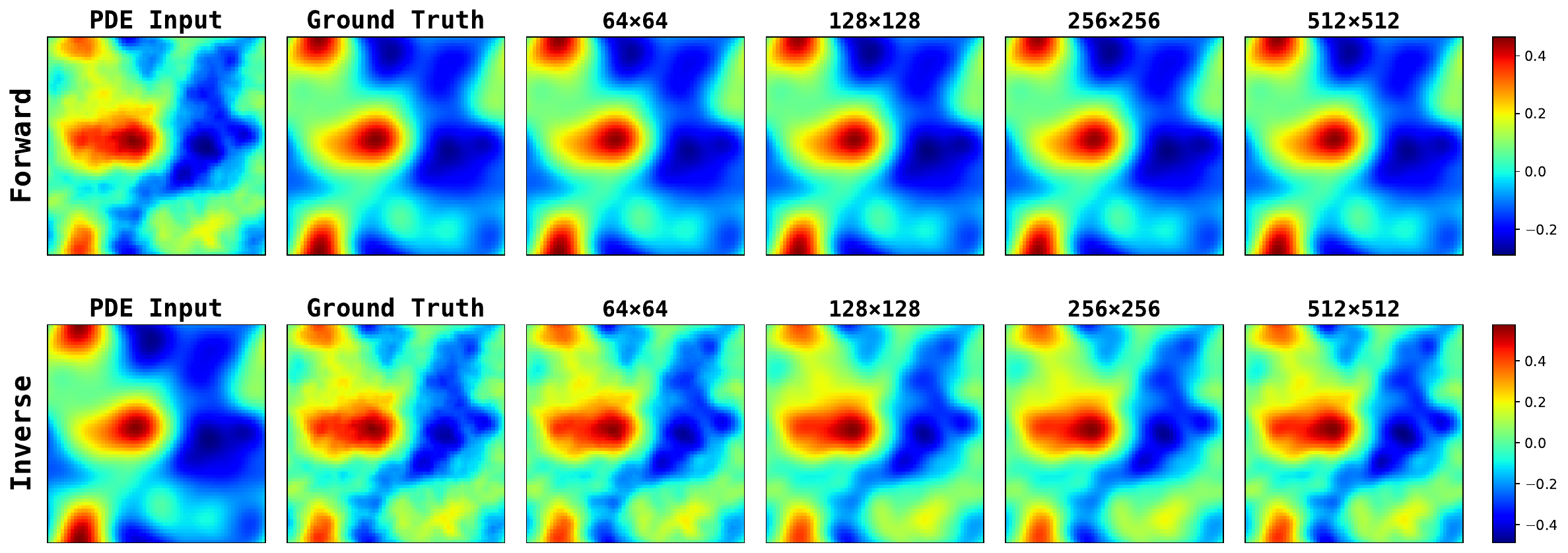}
        \label{fig:nsnb_99_superres_64}
    \end{subfigure}

    \vspace{-0.4em}

    \caption{
    Super-resolution results for Helmholtz(top), Darcy flow, and
    Navier--Stokes(bottom).
    }
    \label{fig:superres_qualitative}
\end{figure}

\vspace{-1ex}
\section{Qualitative Results}

\label{appendix:qualitative}


Figure~\ref{fig:noisy_comparison_nsb_poisson_darcy_appendix} provides a qualitative comparison of the solutions generated by PRISMA and the baseline models under the challenging {noisy observation} setting. Across both problems: Poisson and Darcy Flow, PRISMA's predictions are visually faithful to the ground truth solutions. The corresponding error maps are consistently darker, and the reported error values are significantly lower compared to the baselines. In contrast, solutions from FunDPS often appear corrupted by noise, while those from DiffusionPDE can be blurry or inaccurate, particularly in complex inverse problems. 
{Figures \ref{fig:full_comparison_darcy_appendix}, \ref{fig:full_comparison_helmholtz_appendix}, \ref{fig:full_comparison_poisson_appendix} presents qualitative comparison of the solutions generated by PRISMA and the baseline models under Full observation setting. Figure \ref{fig:sparse_comparison_nsnb_appendix} compare PRISMA predictions against the baseline models on Navier-Stokes Equation under Sparse observation settings. }

\begin{figure}[h]
    \centering
    \captionsetup{aboveskip=2pt}

    \begin{subfigure}[t]{\linewidth}
        \centering
        \includegraphics[width=\linewidth]{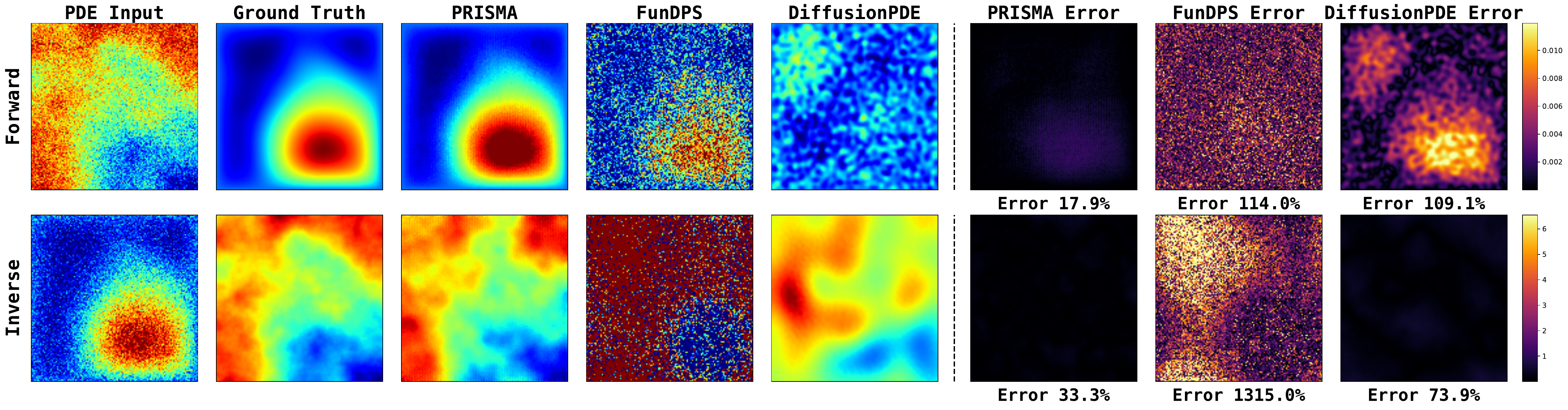}
        \label{fig:poisson_19_noisy}
    \end{subfigure}

    \begin{subfigure}[t]{\linewidth}
        \centering
        \includegraphics[width=\linewidth]{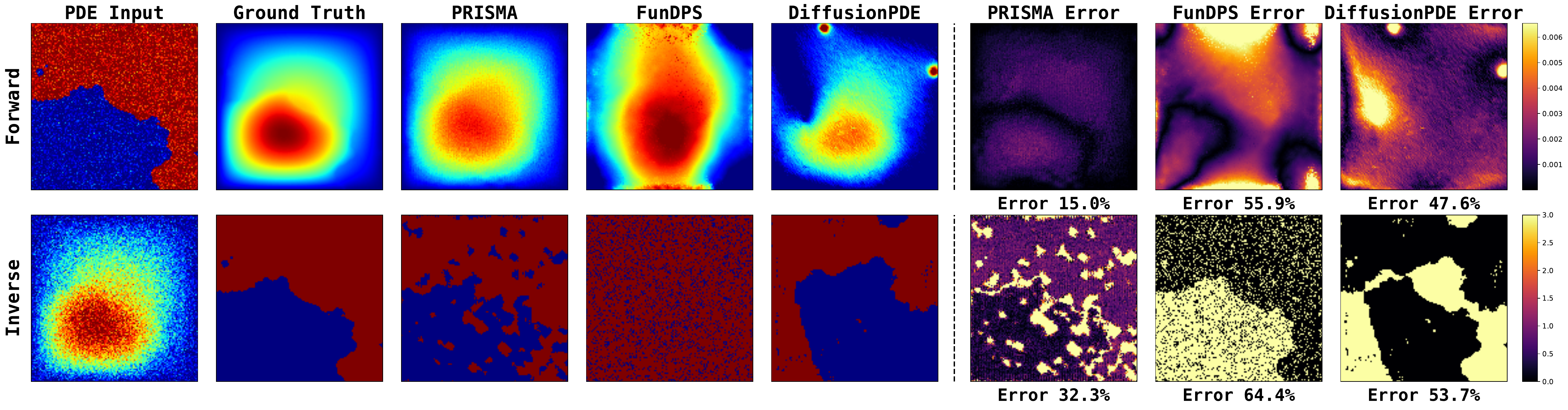}
        \label{fig:darcy_97_noisy}
    \end{subfigure}

    \caption{Poisson (\textit{top}) and Darcy (\textit{bottom}) Equations under Noisy Observation}
    \label{fig:noisy_comparison_nsb_poisson_darcy_appendix}
\end{figure}

\vspace{-2ex}

\begin{figure}[h]
    \centering
    \captionsetup{aboveskip=2pt}

    \begin{subfigure}[t]{0.485\linewidth}
        \centering
        \includegraphics[trim={0cm 0cm 2.1cm 0cm}, clip, width=\linewidth]{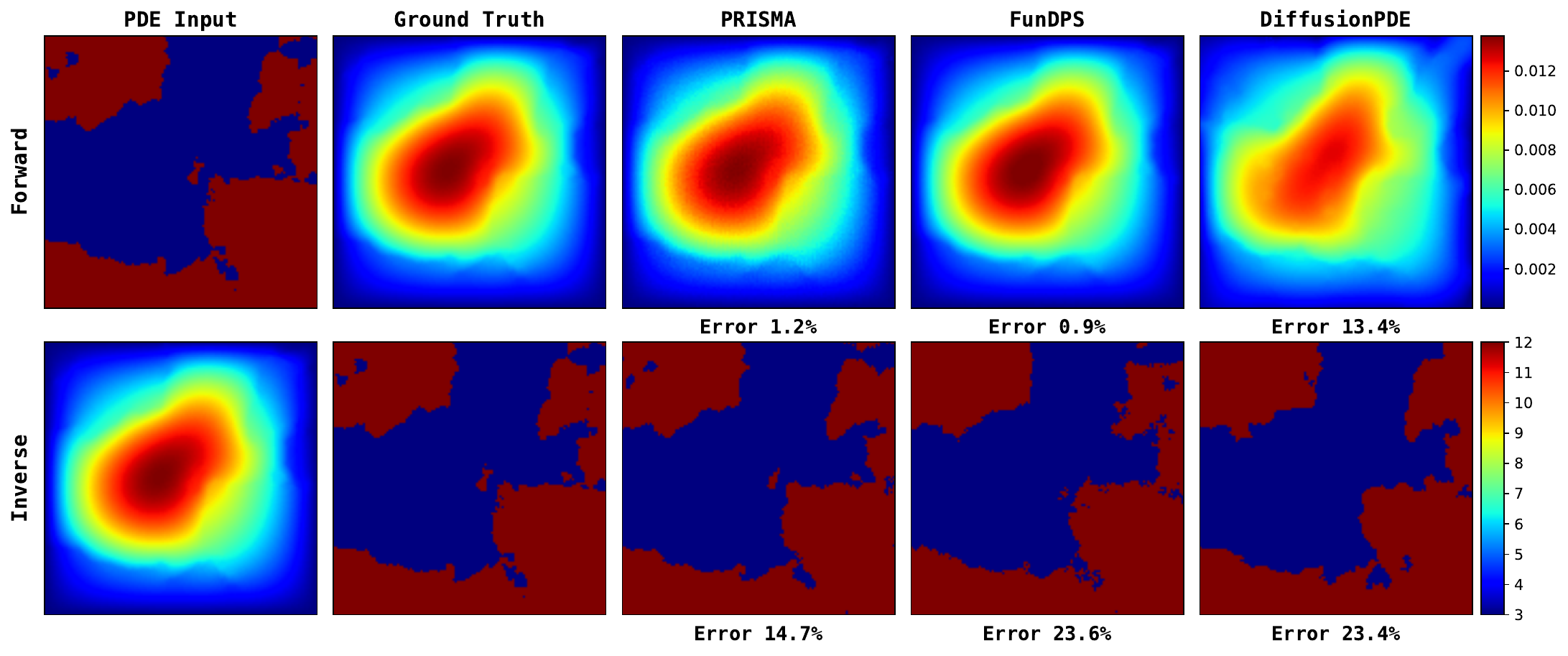}
        \label{fig:darcy_0_full}
        \vspace{-2ex}
    \end{subfigure}
    \hfill
    \begin{subfigure}[t]{0.508\linewidth}
        \centering
        \includegraphics[trim={0.7cm 0cm 0cm 0cm}, clip, width=\linewidth]{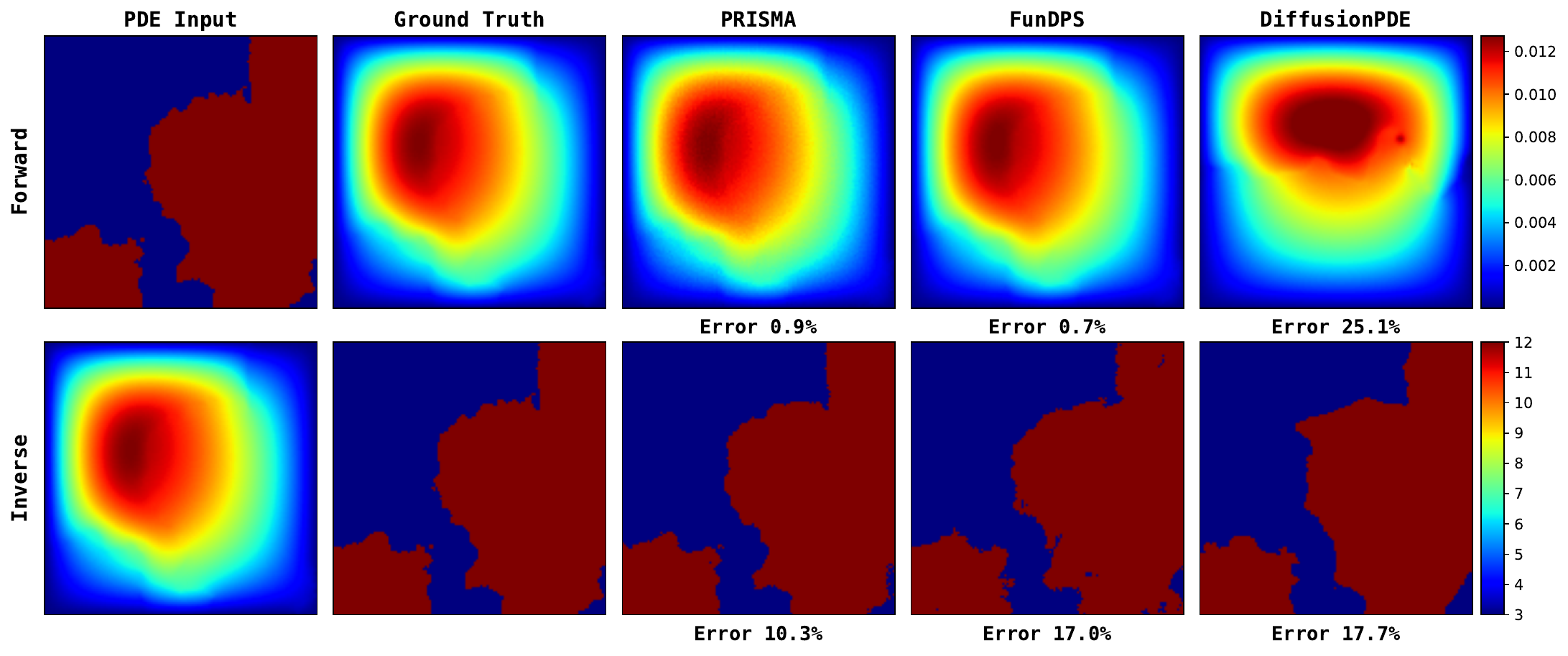}
        \label{fig:darcy_98_full}
        \vspace{-2ex}
    \end{subfigure}
    \caption{{Prediction (\textit{for two samples}) Visualization on Darcy Equation under Full Observation}}
    \vspace{-2ex}
    \label{fig:full_comparison_darcy_appendix}
\end{figure}

\begin{figure}[!htbp]
    \centering
    \captionsetup{aboveskip=2pt}

    \begin{subfigure}[t]{\linewidth}
        \centering
        \begin{subfigure}[t]{0.485\linewidth}
            \centering
            \includegraphics[
                trim={0cm 0cm 2.7cm 0cm},
                clip,
                width=\linewidth
            ]{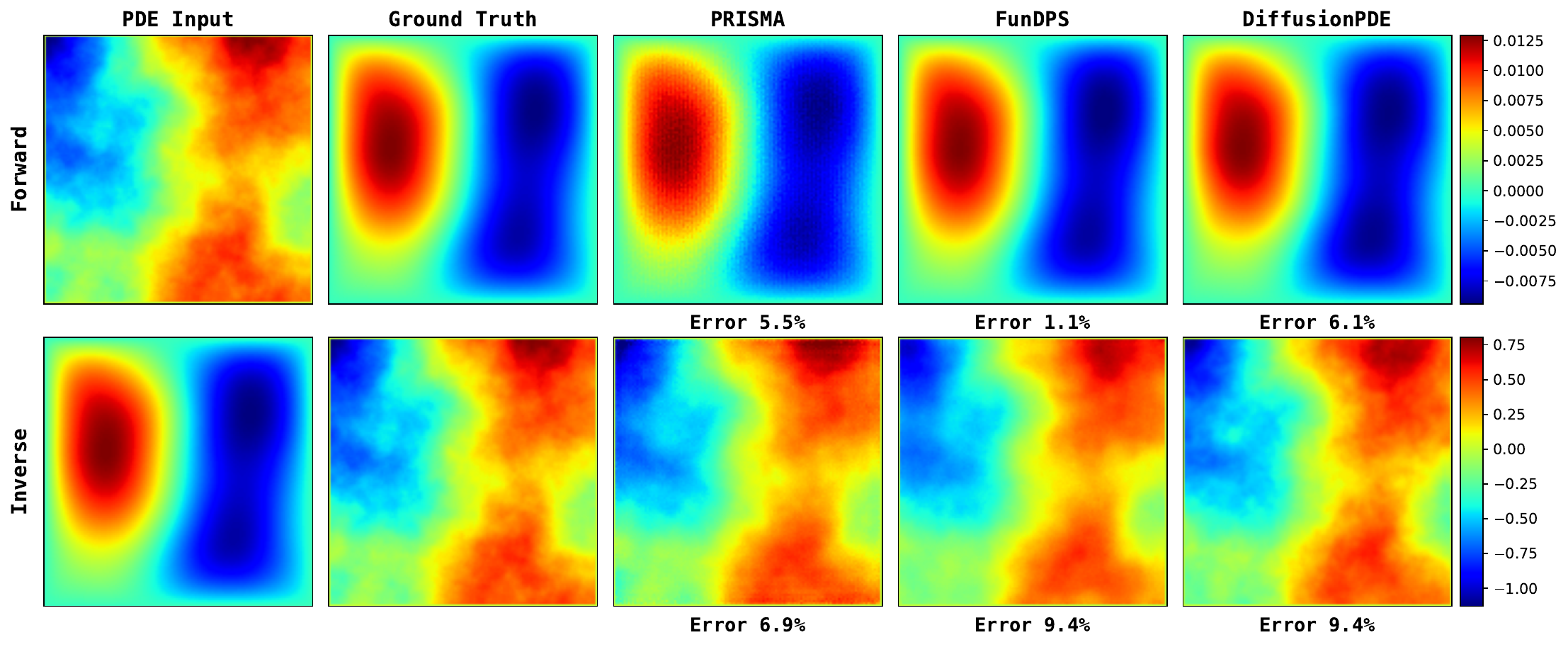}
            \label{fig:helm_98_full}
        \end{subfigure}
        \hfill
        \begin{subfigure}[t]{0.508\linewidth}
            \centering
            \includegraphics[
                trim={0.7cm 0cm 0cm 0cm},
                clip,
                width=\linewidth
            ]{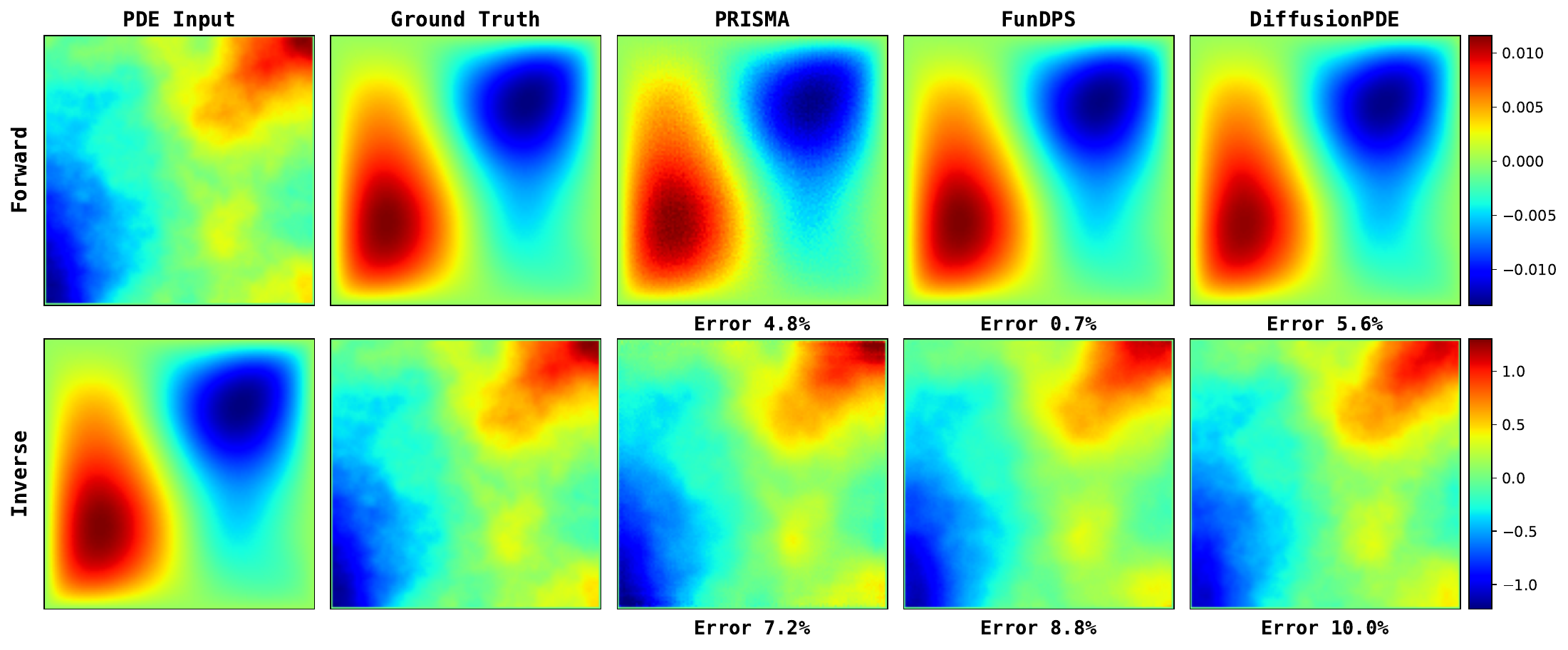}
            \label{fig:helm_97_full}
        \end{subfigure}
        \caption{Helmholtz under full observation.}
        \label{fig:full_comparison_helmholtz_appendix}
    \end{subfigure}
    \vfill
    \vspace{0.6em}

    \begin{subfigure}[t]{\linewidth}
        \centering
        \begin{subfigure}[t]{0.484\linewidth}
            \centering
            \includegraphics[
                trim={0cm 0cm 2.7cm 0cm},
                clip,
                width=\linewidth
            ]{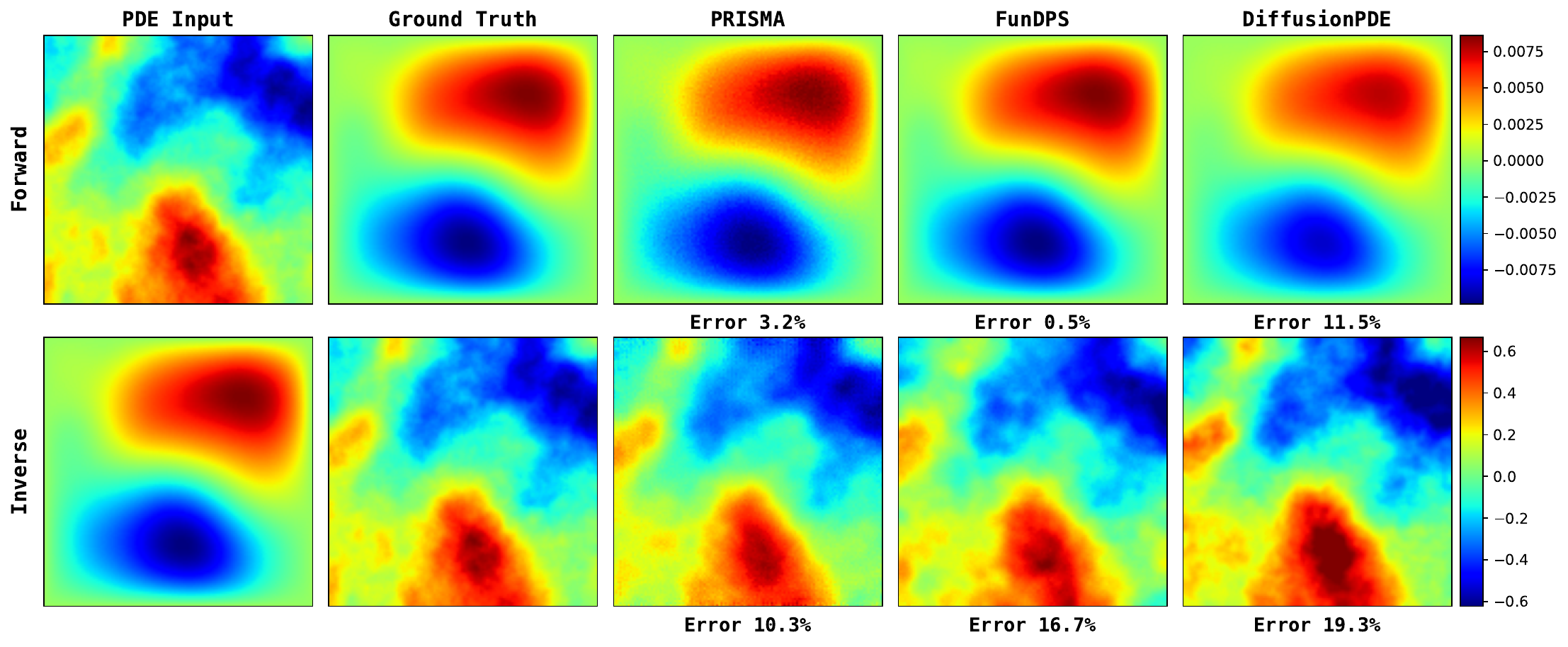}
            \label{fig:poisson_1_full}
        \end{subfigure}
        \hfill
        \begin{subfigure}[t]{0.5097\linewidth}
            \centering
            \includegraphics[
                trim={0.7cm 0cm 0cm 0cm},
                clip,
                width=\linewidth
            ]{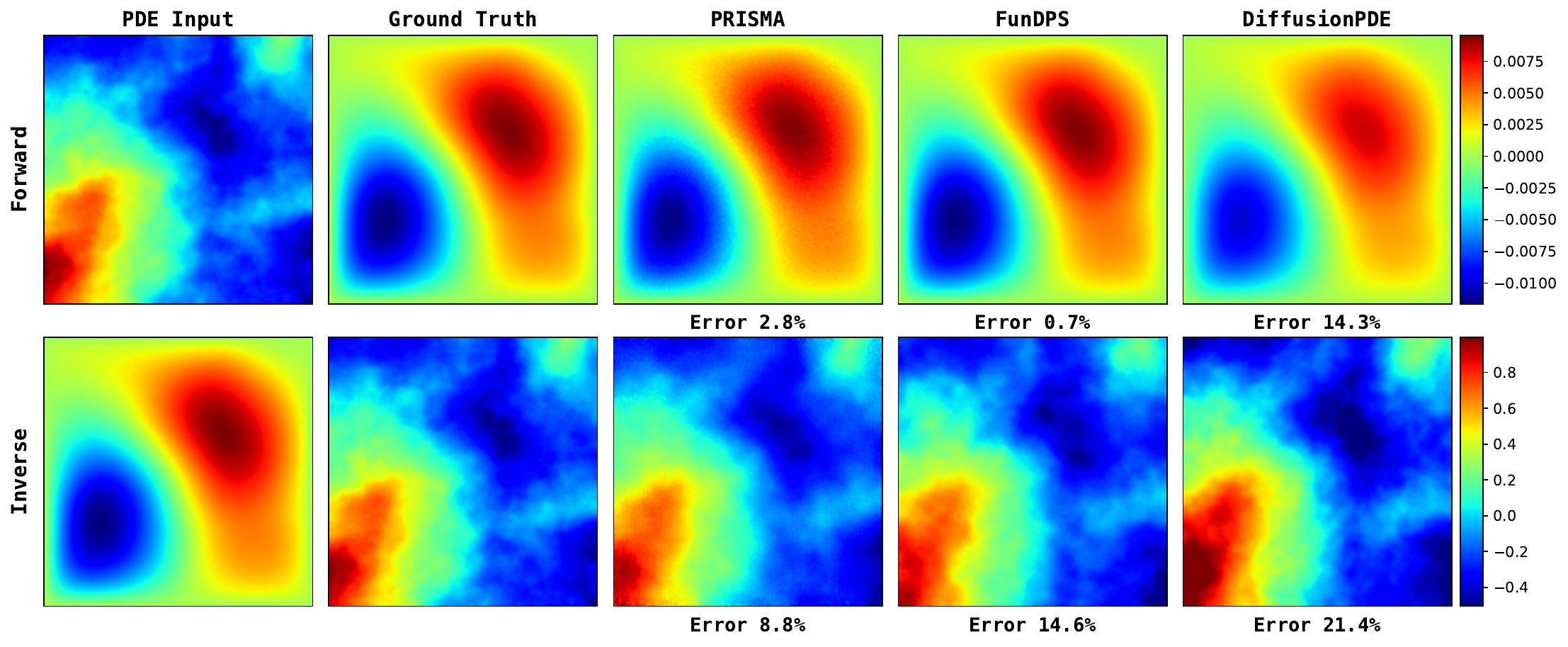}
            \label{fig:poisson_99_full}
        \end{subfigure}
        \caption{Poisson under full observation.}
        \label{fig:full_comparison_poisson_appendix}
    \end{subfigure}
     \vspace{0.6em}


    \begin{subfigure}[t]{\linewidth}
        \centering
        \begin{subfigure}[t]{0.48\linewidth}
            \centering
            \includegraphics[
                width=\linewidth
            ]{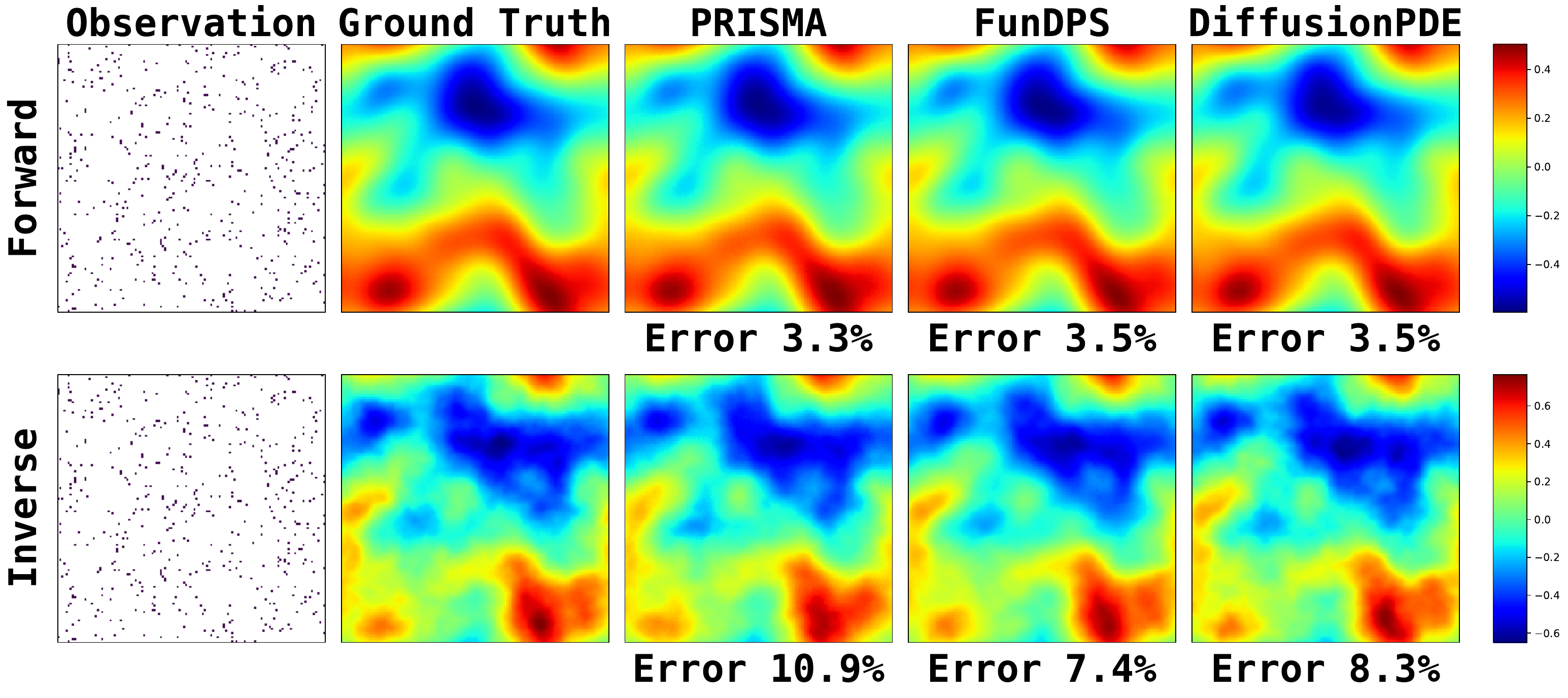}
            \label{fig:nsnb_1_sparse}
        \end{subfigure}
        \hfill
        \begin{subfigure}[t]{0.48\linewidth}
            \centering
            \includegraphics[
                width=\linewidth
            ]{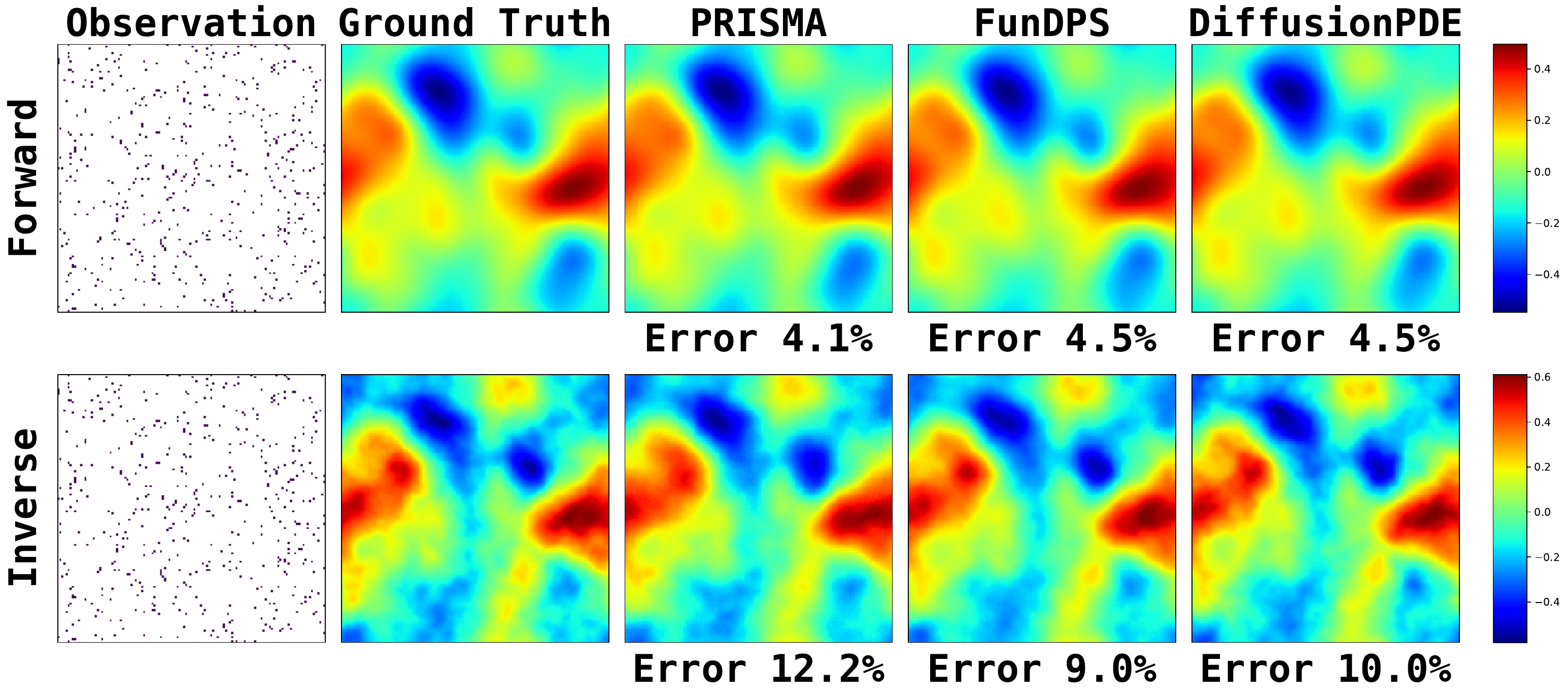}
            \label{fig:nsnb_0_sparse}
        \end{subfigure}
        \caption{Navier--Stokes (non-bounded) under sparse observation.}
        \label{fig:sparse_comparison_nsnb_appendix}
    \end{subfigure}

     \vspace{0.6em}

    \caption{
    Qualitative predictions for two representative samples on
    (a) Helmholtz under full observation,
    (b) Poisson under full observation, and
    (c) Navier--Stokes (non-bounded) under sparse observation.
    Relative $\ell_2$ error is reported below each prediction.
    }
    \label{fig:combined_qualitative_appendix}
\end{figure}

\end{document}